\documentclass{elsarticle}

\usepackage{hyperref}
\usepackage[dvipsnames]{xcolor}
\usepackage{amssymb}
\usepackage{array}
\newcolumntype{P}[1]{>{\centering\arraybackslash}p{#1}}
\newcolumntype{C}{>{\centering\arraybackslash} m{1.28cm} }
\newcolumntype{L}[1]{>{\raggedright\let\newline\\\arraybackslash\hspace{0pt}}m{#1}}
\newcolumntype{R}[1]{>{\raggedleft\let\newline\\\arraybackslash\hspace{0pt}}m{#1}}
\usepackage{mathtools, nccmath}
\usepackage{xpatch}
\usepackage{makecell}
\usepackage{pgfplots}
\usepackage{relsize}
\usepackage{multirow}
\usepackage{wrapfig}
\usepackage{lscape}
\usepackage{float}
\usepackage{blindtext}
\usepackage{rotating}
\usepackage{amsfonts}
\usepackage{pbox}
\usepackage{tcolorbox}
\usepackage[table]{colortbl}
\usepackage{tablefootnote}
\usetikzlibrary{backgrounds}
\usepackage{subfig}
\usepackage{lmodern}
\usepackage{multicol}
\usepackage{bm}
\usepackage{graphicx, rotating, caption, lscape, threeparttable}%

\usepackage{tikz}
\usetikzlibrary{shadows,fadings}
\usetikzlibrary{arrows, petri, topaths, graphs, graphs.standard}
\usepackage{tkz-graph}

\usetikzlibrary{positioning}
\usetikzlibrary{patterns}
\usetikzlibrary{arrows,shapes}
\usetikzlibrary{plotmarks}
\usepackage{varwidth}
\usetikzlibrary{fit}
\usepackage{longtable}
\usepackage{booktabs} % For professional looking tables
\usepackage{tabularx}
\usepackage{arydshln}
\usepackage{hhline}

\newcolumntype{P}[1]{>{\centering\arraybackslash}p{#1}}
\newcolumntype{C}[1]{>{\centering\arraybackslash} m{#1} }
\newcolumntype{L}[1]{>{\raggedright\let\newline\\\arraybackslash\hspace{0pt}}m{#1}}
\newcolumntype{R}[1]{>{\raggedleft\let\newline\\\arraybackslash\hspace{0pt}}m{#1}}
\usepackage{enumitem}
\usepackage{listings}
\usepackage{pifont}% http://ctan.org/pkg/pifont
\usepackage{amsthm}
\usepackage{mdframed}
\usepackage{lipsum}
\theoremstyle{definition}
\newtheorem{definition}{Definition}
\theoremstyle{plain}
\newtheorem*{hypothesis}{Hypothesis}
\usepackage{physics}
\usepackage{amsmath}
\usepackage{mathdots}
\usepackage{yhmath}
\usepackage{cancel}
\usepackage{color}
\usepackage{siunitx}
\usepackage{array}\usepackage{physics}
\usepackage{amsmath}
\usepackage{tikz}
\usepackage{mathdots}
\usepackage{yhmath}
\usepackage{cancel}
\usepackage{color}
\usepackage{siunitx}
\usepackage{multirow}
\usepackage{colortbl}
\usepackage{gensymb}
\usepackage{extarrows}
\usetikzlibrary{shadows.blur}
\usetikzlibrary{shapes.symbols}

\journal{Elsevier}

\begin{document}

\begin{frontmatter}

\title{Comparing and extending the use of defeasible argumentation with quantitative data in real-world contexts}


\author{Lucas Rizzo*}
\cortext[mycorrespondingauthor]{Corresponding author}
\ead{lucas.rizzo@tudublin.ie}
\author{Luca Longo}
\ead{luca.longo@tudublin.ie}
\address{School of Computer Science, Technological University Dublin, Dublin, Ireland}


\begin{abstract}

Dealing with uncertain, contradicting, and ambiguous information is still a central issue in Artificial Intelligence (AI). As a result, many formalisms have been proposed or adapted so as to consider non-monotonicity, with only a limited number of works and researchers performing any sort of comparison among them. A non-monotonic formalism is one that allows the retraction of previous conclusions or claims, from premises, in light of new evidence, offering some desirable flexibility when dealing with uncertainty. This research article focuses on evaluating the inferential capacity of defeasible argumentation, a formalism particularly envisioned for modelling non-monotonic reasoning. In addition to this, fuzzy reasoning and expert systems, extended for handling non-monotonicity of reasoning, are selected and employed as baselines, due to their vast and accepted use within the AI community. Computational trust was selected as the domain of application of such models. Trust is an ill-defined construct, hence, reasoning applied to the inference of trust can be seen as non-monotonic. Inference models were designed to assign trust scalars to editors of the Wikipedia project. In particular, argument-based models demonstrated more robustness than those built upon the baselines despite the knowledge bases or datasets employed. This study contributes to the body of knowledge through the exploitation of defeasible argumentation and its
comparison to similar approaches. The practical use of such approaches coupled with a modular design that facilitates similar experiments was exemplified and their respective implementations made publicly available on GitHub \cite{RizzoFramework,RizzoFuzzyImplementation}. This work adds to previous works, empirically enhancing the generalisability of defeasible argumentation as a compelling approach to reason with quantitative data and uncertain knowledge.

\end{abstract}


\end{frontmatter}

%\linenumbers

% Introduction section
\section{Introduction}

Representing and manipulating knowledge with computers is still one of the main challenges in AI. 
This knowledge includes the ability to perform common sense reasoning, which
 is often non-monotonic \cite{brewka1991nonmonotonic}. Non-monotonic reasoning allows additional information to invalidate old claims or conclusions \cite{reiter1988nonmonotonic,Pollock1974,mccarthy1980circumscription,Kowalski1991,Longo2015,brewka1991nonmonotonic}.
The classic non-monotonic reasoning example is given by `\textit{birds fly}'. It is reasonable to assume that a particular bird, Tweety, flies, \textit{unless} it is an exceptional bird: ostrich, duck, penguin, and so on \citep{reiter1988nonmonotonic}.
This type of reasoning 
can be modelled
in AI by several
non-monotonic formalisms \citep{brewka1991nonmonotonic}, 
such as inheritance networks with exception
\citep{horty1990skeptical}, semantic networks using Dempster’s rule \citep{ginsberg1984non}, non-monotonic logics \cite{mccarthy1980circumscription,moore1985semantical,reiter1980logic} and
knowledge-based systems \cite{akerkar2009knowledge}. Still, to the best of
the authors' knowledge, there is an absence in the literature of comparisons among some of these formalisms. This research 
article focuses on the comparison of knowledge-based, non-monotonic systems, which make use of rules or arguments supporting or contradicting 
certain conclusions to formalise non-monotonic reasoning.
For instance, fuzzy reasoning \citep{zadeh1965fuzzy} and expert systems 
\citep{durkin1998expert} with the addition of non-monotonic layers, and 
computational argumentation, also referred to as defeasible argumentation
 \citep{bryant2008review,prakken2001logics}. 
All these approaches have led to the development of non-monotonic reasoning models
usually based upon knowledge bases provided by human experts. Therefore, their performance depends on the amount and quality of knowledge available. 
However, they also allow such knowledge, possibly fragmented, vague, and non-algorithmic, to be represented in a natural and structured way
\citep{guida1995design}.
Intuitively, this provides a higher degree of interpretability and transparency to the reasoning process. The reason for that is because these models attempt to 
use human language and to follow the way humans reason. This attempt potentially increases their explainability, which is essential for their adoption and usage. Nonetheless, these 
advantages have not been sufficient to increase the use of defeasible argumentation technology when performing quantitative reasoning under uncertainty. In this 
case, quantitative reasoning is understood as reasoning built with domain knowledge and performed on quantitative data, thus being able to provide numerical 
inferences.
  \citeauthor{bench2007argumentation} \cite{bench2007argumentation} identified a set of challenges that need to be overcome for achieving this goal, including the lack of a strong link between 
argumentation and other formalisms and the lack of engineering solutions for the application of argumentation. Another problem arises from the early stage of 
research -- trying to add quantitative approaches to argumentation \citep{longo2016argumentation,alej2004integrating}. Often, quantitative approaches in AI are 
deemed as limited for their inability to provide justifiable conclusions \citep{Chesnevar2009empowering}.
 Hence, this study attempts to empirically evaluate the inferential capacity of defeasible argumentation 
models against other similar reasoning approaches. In addition to defeasible argumentation, non-monotonic fuzzy reasoning and expert systems are selected and 
employed as baselines, due to their vast and accepted use within the AI community. 

To perform this comparison, the problem of
representing the construct of computational trust has been chosen. Trust is a crucial human
construct investigated by several disciplines, such as psychology, sociology, and philosophy
\cite{Sabater2005Trust}.
It is an ill-defined construct, whose application lies in the domain of knowledge representation and reasoning.
In this research article, the modelling of reasoning applied to the inference of computational trust is proposed in the context of the Wikipedia project.
The goal is to design non-monotonic, knowledge-based models capable of 
assigning a trust value in the range [0, 1] $\subset\mathbb{R}$ to Wikipedia editors on a case-by-case basis.
One means complete trust should be assigned to an editor, while 0 means an absence of trust assigned to the editor.
These models are built upon domain knowledge and instantiated by quantitative data, thus can provide numerical inferences.
Moreover, the domain knowledge employed contains pieces of evidence and arguments that can be withdrawn in light of new 
information, allowing the proposed assignment of trust to be seen as a form of defeasible reasoning activity.
It is expected that the comparison of non-monotonic reasoning approaches applied in the domain of computational trust will improve the perception of defeasible argumentation in relation to other similar alternatives. Moreover, such an experiment will also add to previous works that have made similar empirical comparisons in different domains of application \cite{rizzo2020empirical,LONGO2021106514,rizzo2018investigation,RizzoML18}, potentially enhancing the generalisability of defeasible argumentation as a compelling approach to reason with quantitative data and uncertain knowledge.
In turn, this enhancement could possibly enable different applications and experiments, likely to be defeasibly modelled, to be carried out.

The remainder of this paper is organised as follows: Section 2 provides the related
work on expert systems and fuzzy
reasoning, including their options for handling non-monotonicity, followed then by defeasible argumentation and a short description of computational trust. Section 3 presents the design of the empirical experiment
proposed to allow the envisioned comparison. 
Section 4 describes the results, performs its analysis and presents
the respective discussion. Lastly, Section
5 concludes with a summary of the study, limitations, findings and recommendations of future research.

% Literature review section

\section{Literature and Related Work}\label{sec:literature_review}

 In order to enhance the understanding of
 non-monotonic reasoning approaches, in particular defeasible argumentation, this section provides the 
 reader with the main notions and properties of non-monotonicity, a brief description to the most common
 non-monotonic logics, and a precise description of the studied knowledge-based systems. Lastly, the section concludes with a short review of some works that have attempted to compare non-monotonic formalisms, followed by
 an introduction to computational trust, the chosen application domain for comparison purposes from a real-world context.

\subsection{Non-monotonic Logics}

Non-monotonic logics are fundamental for the better comprehension of the concept of non-monotonicity. Default logic, autoepistemic logic and
circumscription have commonly been referred to as the most important ones 
\cite{baroni1997full,brewka2008nonmonotonic}. Briefly, default logic \cite{reiter1980logic} 
models default reasoning, which is performed by employing default knowledge. Default knowledge is of the form that can be retracted if new information
that can falsify its preconditions becomes available. The standard example is given by
‘normally, birds fly’. If Tweety is a bird, and there is no information to assume that Tweety
does not fly, it is assumed that Tweety flies. Rules in this logic are called \textit{defaults} and are
represented by expressions in the following form:
$p(x):j_1(x)\ldots j_n(x)/c(x)$, where  $p(x)$ is a prerequisite, $j_i(x)$ are justifications and $c(x)$ is the consequent of the default. Formally in the Tweety example: $bird(Tweety):flies(Tweety)/flies(Tweety)$. In natural language: if Tweety is a bird, and based on the available information, it is possible to assume that Tweety flies, then infer that Tweety flies.

Similarly, autoepistemic logic \citep{moore1985semantical} is a formalism of modal reasoning \citep{Chagrov1997} focused on modelling reasoning about what is believed 
by known propositions. It assumes that any specific information should be known and hence it is possible to reason 
with what is known. It believes that if Tweety is a bird and 
it is not believed that Tweety does not fly, then Tweety flies.
In another way, circumscription 
\citep{mccarthy1980circumscription}, tries to represent non-monotonicity through the concept of \textit{abnormality}. A circumscription technique presumes that 
Tweety flies because there is no information to show that it has any abnormality. 

In essence, proof-theoretic formalisations are often hard to evaluate with respect to their consistency and intuitions that are supposed to be captured. 
For instance, \citeauthor{poole1989lottery} \cite{poole1989lottery} describes the problem of not committing to implicit assumptions when using default 
reasoning. 
 Suppose the standard example, ‘normally, birds fly’, and that by default reasoning Tweety flies: is it then reasonable to assume that Tweety is not an emu or a penguin? \citeauthor{poole1989lottery} proposes three solutions: not concluding that Tweety is a bird; making no commitment whether Tweet is an emu or 
 a penguin; and concluding that Tweety is neither an emu nor a penguin.
Compared to non-monotonic logics, knowledge-based systems are better suited for capturing the intuitions of a specific problem. Since rules or arguments must be predefined, only relevant non-monotonic contexts are modelled, leaving little, if any, place for confusion. The next subsections introduce such systems employed in this research article.
 
\subsection{Expert Systems}

Succinctly, expert systems are defined as systems that try to transfer a vast body of specific knowledge from a human to a computer. They attempt to
emulate such a human in a given field  \citep{durkin1998expert}
and are aimed at accomplishing tasks that require human expertise or at playing the role of an assistant \citep{jackson1990introduction}. Their 
structure is usually composed of two internal components: a \textit{knowledge base} and an 
\textit{inference engine} \citep{durkin1998expert} 
The former is provided by a human expert and generally translated 
into a set of logical rules. The latter is aimed at eliciting, firing and aggregating such rules towards a conclusive inference.
Rules are used to define what to do and what to conclude in different 
scenarios. Usually, they follow the form depicted in Fig. \ref{fig:ifthenrule}.
\\

\begin{figure}[h!]
\begin{center}
\textbf{IF} (antecedent) \textbf{THEN} (consequent) \vspace{-4mm}
\end{center}
\caption{Typical form of rules employed in rule-based expert systems.}
\label{fig:ifthenrule}
\end{figure}

These rules, in conjunction with a set of facts (for instance data) and an interpreter that decides the application of the rules, are what constitute a 
rule-based system. They can model a large range of problems, once the domain knowledge is represented as IF-THEN rules. It is important for the number 
of rules not to be too large so as to make the interpreter inefficient \citep{grosan2011rule}. 
In this case, the system can exploit users' inputs and pieces of information stored in the knowledge 
base to reason with.

\subsubsection{Non-monotonicity in Expert Systems}

The use of non-monotonic logics in expert systems has been studied for several decades \citep{gabbay1985theoretical}. Nonetheless, the general use of 
non-monotonic reasoning in industry has not been extensive [\citealp{morgenstern1997expert,morgenstern1998inheritance,puppe1993systematic}].
A few examples of expert systems that deal with non-monotonicity are proposed through the use of inheritance with exceptions in semantic 
networks \cite{morgenstern1997expert}, through the use of defeasible logic \cite{nute1990controlling}, through the use of default reasoning \cite{el2002diagnostic}, and through the use of probabilistic reasoning \cite{LEWIS20103616}.
In this research article, knowledge in expert systems is represented by rules, and the respective reasoning is performed in a single step.
In other words, data is  imported, and all rules are fired at once. Thus, to retract a rule, the notion of `contradictions' or `exceptions' are employed.
These are defined by domain experts and describe special cases in which a rule is no longer valid. 
Once a special case is triggered, a backtrack search is employed to remove affected 
rules \cite[chap. 9]{puppe1993systematic}. 
This might require excessive efforts, depending on the amount of data to be managed and the number of reasoning steps firing the 
backtrack search. However, note that even though this is a simplistic procedure, it can still be effectively implemented in a problem of single reasoning step, with a reasonable number of rules. That is the case with computational trust, the application chosen here for comparison purposes.
Because reasoning applied to model computational trust can be performed in a single step with a reasonable number of rules, expert systems designed in this study do not follow a usual multi-step reasoning process. Further information on the domain of application and design methodologies will be detailed afterwards.

\subsection{Fuzzy Reasoning}\label{reivewfuzzy} 

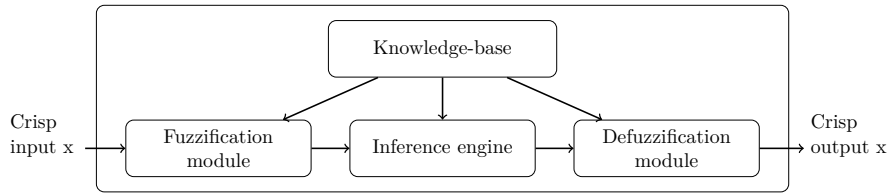
\begin{figure}[h!]
\begin{center}
\normalsize
\resizebox{1.0\linewidth}{!}{
\begin{tikzpicture}[
    scale=0.44,
    %start chain=1 going below, 
    %start chain=2 going right,
    node distance=6mm,
    desc/.style={
        %scale=0.75,
        %on chain=5,
        rectangle,
        rounded corners,
        draw=black, 
         thin,
        text centered,
        text width=10cm,
        minimum height=10mm,
        %fill=blue!30
        }
]

% Descriptions

\newcommand{\boxwidth}{3.8cm}

\node [desc, fill=none, text width=12cm, minimum height=33mm] (N) at (0, -1)  {};

\node [desc, text width=\boxwidth] (A) at (0, 1)  {{Knowledge-base}};

\node [desc, text width=0.8*\boxwidth] (B) at (-9,-3)  {{Fuzzification module}};
\node [desc, text width=0.8*\boxwidth] (C) at (0,-3)  {{Inference engine}};
\node [desc, text width=0.8*\boxwidth] (D) at (9,-3)  {{Defuzzification\\module}};

\draw [->, thick] (A) -- (C);
\draw [->, thick] (A) -- (B);
\draw [->, thick] (A) -- (D);

\draw [->, thick] (B) -- (C);
\draw [->, thick] (C) -- (D);

\node [text width=1.2cm, minimum height=5mm] (Z) at (-16, -2.0)  {Crisp};
\node [text width=1.2cm, minimum height=5mm] (E) at (-16, -3.0)  {input x};

\node [text width=1.5cm, minimum height=5mm] (Z) at (16.5, -2.0)  {Crisp};
\node [text width=1.5cm, minimum height=5mm] (F) at (16.5, -3.0)  {output x};

\draw [->, thick] (E) -- (B);
\draw [->, thick] (D) -- (F);
\end{tikzpicture}
}
\end{center}

\caption{General structure of a fuzzy control system. Adapted from \cite{cordon2011historical}.}
\label{fig:fuzzycontrol}
\end{figure}

Fuzzy reasoning models are the product of knowledge-based systems that incorporate fuzzy logic and/or fuzzy sets \citep{zadeh1965fuzzy} into their reasoning and 
knowledge representation techniques \citep{kandel1991fuzzy}.
Fuzzy logic calculus coupled with the notion of fuzzy membership functions allows for an effective knowledge representation of imprecise and uncertain 
pieces of information. Such notions can be exploited by IF-THEN rules of similar structure to those depicted in Fig. \ref{fig:ifthenrule}.
In this case,
a fuzzy rule-based system arises, which is most useful when modelling systems that make use of linguistic variables as their antecedents and consequents. 
In turn, rules can be employed for the construction of a fuzzy control system. Usually, such a system is composed of a set of crisp inputs, a knowledge base, a 
fuzzification module, an inference engine and a defuzzification module
\citep{passino1998fuzzy}, as depicted by the diagram in Fig. \ref{fig:fuzzycontrol}.

This structure begins by the fuzzification module assessing the membership grades of crisp inputs associated with fuzzy sets. 
Subsequently, a fuzzy inference method must be applied, in order to produce an inference.

Many of such inference methods are available \cite{takagi1993fuzzy,tsukamoto1979,mamdani1974application}. This research article employs the `Mamdani' fuzzy inference method \citep{mamdani1974application}, which is often used in practice \citep{ross1995fuzzy}. 
Finally, 
defuzzification methods \citep{hellendoorn1993defuzzification}, such as cetroid or mean max membership, must be applied to convert a fuzzy set into a crisp output in different fashions, contrarily to fuzzification methods that convert crisp inputs to fuzzy sets.

\subsubsection{Non-monotonicity in Fuzzy Reasoning} \label{sec:nmfuzzy}

Some supplementary extensions of fuzzy inference systems have been suggested, in order to tackle the use of non-monotonic rules. Unfortunately, these extensions 
are few and not well established.
For example, in \citep{castro1998non}, conflicting rules have their conclusions aggregated by a possible averaging function. However, an inference from a non-monotonic fuzzy rule cannot be propagated since the theory does not allow circularity.
Another type of non-monotonicity in fuzzy systems is investigated in \citep{gegov2014rule}. In this instance, non-monotonicity arises when identical consequents 
are inferred by distinct permutations of variables in the antecedents. 
The goal of this approach is to remove 
redundant rules, while preserving the crisp values returned by the defuzzification module of a Mamdani fuzzy system. 
A third approach is given by \citeauthor{siler2005fuzzy} [\citealp{siler2005fuzzy}, chap. 8],
whereby possibility theory \citep{dubois1998possibility} is included into the fuzzy reasoning system to tackle conflicting instructions.
\citeauthor{siler2005fuzzy} [\citealp{siler2005fuzzy}] make the assumption that `truth values represent \textit{necessity}, the extent to which the data support a proposition'. At the same 
time, they also treat `truth values that represent 
\textit{possibility}, the extent to which a truth value represents the extent to which the data fails to refute a proposition'.
The possibility of proposition $A$ is denoted $Pos(A)$, while its necessity is denoted $Nec(A)$. Both are values between [0, 1] $\subset 
\mathbb{R}$. Necessity is also assumed to represent the traditional truth values reviewed on the previous subsections, while truth values that represent 
possibility need to be added to the system.
\textcolor{black}{Moreover, it is assumed that adding supporting evidence can affect the necessity but not the possibility of a proposition, and adding 
contradicting evidence can never increase possibilities. In other words, $Nec(A) \le Pos(A)$, for any proposition $A$.
In this case it is guaranteed that propositions are defeasible. For example, if $Nec(a) = 1$ and $Pos(a) = 0$, it 
would not be possible to refute $a$}.
Under these circumstances, the effect on the necessity of a proposition $a$ by a set of 
propositions $\{Q_1,\ldots,Q_k\}$ contradicting $a$, and a set of propositions $\{P_1,\ldots,P_j\}$ supporting $a$, is derivable as:

\begin{equation}\label{eq:possibility}
 Nec(a) = (Nec(a) \bigcup_{j} Nec(P_j)) \bigcap _k (\neg Nec(Q_k))
\end{equation}
where $\neg Nec(Q_i) = 1 -  Nec(Q_i)$, the union is implemented by the `max' operator and the intersection by the `min' operator.
A few axioms used to develop conventional possibility theory are not considered in this approach, due to their incompatibility with other fuzzy logics. However, 
according to [\citealp{siler2005fuzzy}]
the advantage provided is a functional theory, when incorporated into fuzzy reasoning with rule-based systems.
For instance, suppose a proposition $a$ whose necessity is 0.3 and possibility is 1.0; if $P$ (necessity 0.4) supports $a$ and $Q$ (necessity 0.2) refutes $a$, 
then $Nec(a) = min(max(0.3, 0.4), 1 - 0.2)$. Thus, the new extent to which $a$ support its truth is 0.4, because of the support 
of $P$ and the failed attempt of refutation from $Q$, due to its low necessity.
Unlike other reviewed implementations of non-monotonicity, note that this approach does not 
restrict 
the type of membership functions, methods of fuzzy inference, methods of defuzzification or propagation of inferences generated by non-monotonic rules. However, 
it does require that possibility values be defined. Simple approaches might be to assume possibility 1 for propositions that can be refuted by any other piece 
of information, and possibility 0 for propositions that cannot be refuted by any other piece of information.

As for applications of non-monotonic fuzzy systems, to the best of our knowledge, no paper has adopted these reviewed approaches in real-world contexts.
Instead, they have been evaluated under simulated environments or by hypothetical examples. In particular, \citeauthor{siler2005fuzzy} \cite{siler2005fuzzy}
exemplify its proposal of adding possibility theory into a fuzzy reasoning system with a simplified application in the medical field for demonstration 
purposes. The goal was to determine the anatomical significance of regions in an echocardiogram composed by ultrasound images of a beating heart. Rules were 
automatically created from a classification database. Contradictions arose from noisy images inferring mutually exclusive conclusions and were resolved with the aid of Equation (\ref{eq:possibility}).

\subsection{Defeasible Argumentation}\label{AT}

Argumentation deals with the study of assertion and definition of arguments usually emerged from divergent opinions. 
Its 
use is rooted in the common tradition, from Aristotle to the present, of employing reasons to decide how to act. 
In AI, unlike expert systems and fuzzy reasoning, computational argumentation theory was introduced as a formalism for modelling non-monotonic reasoning. Thus, it does 
not require an implementation of non-monotonicity as described in the previous reasoning approaches. Non-monotonicity is naturally present through distinct 
techniques for the evaluation of the dialectical status of arguments. This evaluation usually determines which arguments should be ultimately accepted or 
rejected in an interconnected network of arguments. Nowadays, AI is regarded as one of the main areas of application of argumentation theory. In this field, 
argumentation aimed at developing computational models of arguments might also be referred to as \textit{defeasible argumentation} \citep{prakken2001logics} 
---
a paradigm that has become increasingly significant \citep{bench2007argumentation} and widely employed for modelling non-monotonic reasoning 
\citep{chesnevar2000logical}. 

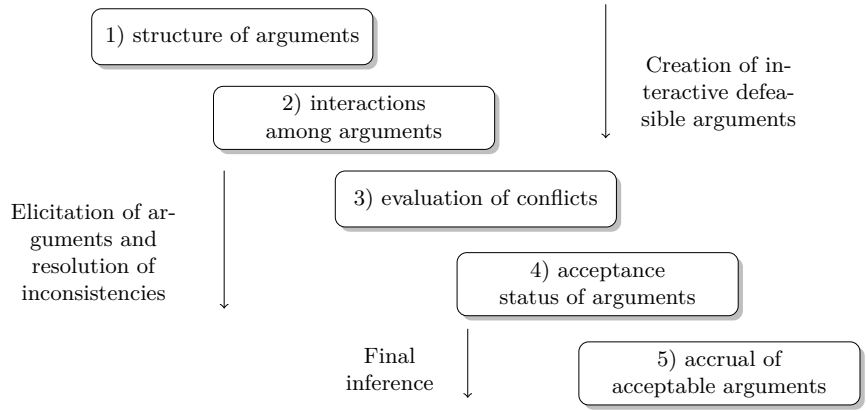
\begin{figure}[h!]
\begin{center}
\small
\resizebox{1\linewidth}{!}{
\begin{tikzpicture}[
    scale=0.58,
    %start chain=1 going below, 
    %start chain=2 going right,
    node distance=6mm,
    desc/.style={
        %scale=0.75,
        %on chain=5,
        rectangle,
        rounded corners,
        draw=black, 
         thin,
        text centered,
        text width=8cm,
        minimum height=8mm,
        %fill=blue!30
        }
]

% Descriptions

\newcommand{\boxwidth}{3.8cm}

\node [desc, fill=gray!0, drop shadow, text width=\boxwidth] at (0.0,-3)  {1)  structure of arguments};
\node [desc, fill=gray!0, drop shadow, text width=\boxwidth] at (3.0,-5) {2) interactions among arguments};
\node [desc, fill=gray!0, drop shadow, text width=\boxwidth]  at (6.0,-7)  {3) evaluation of conflicts};
\node [desc, fill=gray!0, drop shadow, text width=\boxwidth]   at (9.0,-9.1)  {4) acceptance \\status of arguments};
\node [desc, fill=gray!0, drop shadow, text width=\boxwidth]  at (12.0,-11.3)  {5) accrual of \\acceptable arguments};

\draw[ ->,black, text width=3.0cm, text centered] (9.2,-2.2) -- (9.2,-5.5 ) node [  above right] {Creation of interactive defeasible 
arguments};

\draw[ ->,black, text width=3.5cm, text centered] (-0.2,-6.3) -- (-0.2,-9.7 ) node [  above left] {Elicitation of arguments and \\ resolution of 
\\inconsistencies};

\draw[ ->,black, text width=2.0cm, text centered] (5.8,-10.2) -- (5.8,-11.9 ) node [  above left] {Final inference};
\end{tikzpicture}
}
\end{center}

\caption{Five layers structure \citep{longo2016argumentation} for the creation of argument-based models of inference.}
\label{fig:multilayerstructure}

\end{figure}

Computational argumentation systems are usually structured around layers specialising in the definition of internal structure of arguments, the definition of 
arguments interactions, the resolution of conflicts between arguments and the possible resolution strategies for reaching a justifiable 
conclusion \citep{prakken2001logics} .
However, as the boundaries of such layers might not be precisely defined, a few layered structures have been proposed \cite{Prakken2002,atkinson2017Towards}  for the development of computational 
models of argument. Another example of multi-layered structure can be found in \citep{longo2016argumentation} and is depicted in Fig. \ref{fig:multilayerstructure}.

 The present 
research article adopts this structure, due to the nature of the experiment and application selected for comparison purposes (computational trust),
which relies on the the knowledge of a single expert, in order to achieve numerical 
inferences. In the literature of defeasible argumentation, many works are focused on each of these layers. 

For instance, in Layer 1, \citeauthor{Toulmin1958} \citep{Toulmin1958} was one of the first to introduce a conceptual model of argument with better structured arguments. \citeauthor{walton1996argumentation} \citep{walton1996argumentation} proposes a different approach with several different argumentation schemes in which a subject can build his point of view, such as argument from consequence, appeal to expert opinion, and argument from analogy. Other structures are also possible  \cite{bentahar2010taxonomy} and they help to clarify possible ways in which arguments can be represented. Nonetheless, argumentation systems might still be constructed with simpler arguments, for instance when these are represented by a pair of premises and a conclusion.

In Layer 2, several works attempt to model the relationship between arguments and the management of their interactions, which 
means the actual arguing process.  The classification 
proposed by \citeauthor{prakken2010abstract} \cite{prakken2010abstract} exemplifies three different classes: \textit{undercutting} and
\textit{rebutting} attack, first formalised in \citep{pollock1987defeasible}, and \textit{undermining} attack, introduced in \citep{vreeswijk1993studies}.
An undermining attack refers to an argument being attacked on one of its premises, thus it is the only possible class of attack for deductive inferences. In 
contrast, the classes of undercutting and rebutting attack target respectively the inference link and the conclusion of an argument -- structures which can 
be denied only in a defeasible argument. Another type of interaction is given by supporting relations between arguments. An argument system that employs relations of attack and support is said to be 
making use of the concept of \textit{bipolarity} \citep{cayrol2009bipolar}. This concept is not employed in this research article, but there is a broad 
amount of research on the topic of bipolar argumentation which the interested reader is referred 
to \citep{villata2012modelling,cayrol2010coalitions,amgoud2008bipolarity,Nouioua2010Bipolar}.

Subsequently in Layer 3, is possible to find works focused on characterising the success of an attack (or often referred to as 
\textit{defeat}), since defeat relations can often be influenced by reliability of tests and expertise. 
Commonly, attacks have a form of binary relation. However, in order to determine a defeat, two other trends have been observed in the literature of 
argumentation: the preferentiality/strength of arguments and the preferentiality/strength of attacks. Strength of arguments is recognised as a valid source of information for those deciding on a collection of acceptable arguments \citep{dunne2011weighted}. These 
can be employed in a variety of ways, such as using priorities among rules to solve their rebuttals \cite{prakken1997argument}, allowing only arguments of greater strength to attack arguments of lesser strength \cite{pollock1995cognitive}, or even inverting the direction of attacks \cite{amgoud2014Rich}.
\citeauthor{dunne2011weighted} \cite{dunne2011weighted} proposed, instead, the use of strength of attack relations. Their approach is justified by the fact that it is not only the strength of 
arguments that is important, but also the strength of the attack that one argument makes on another.

Once attacks have been evaluated the acceptance status of arguments can be defined through acceptability semantics in Layer 4.
Conflicts by themselves do not demonstrate which arguments should be ultimately accepted. To do so, it is necessary to evaluate the overall interaction 
of arguments across the conflicting set. Most frequently, this evaluation relies on the abstract argumentation theory proposed by \citeauthor{Dung1995} \cite{Dung1995} and later extended \citep{wu2010labelling,caminada2009logical,caminada2006issue}. 
Hence, acceptability semantics return \textit{extensions}, or subsets of arguments
that can be mutually acceptable, according to a specific rationale of the semantics. Therefore, a notion of 
\textit{scepticism} is usually employed in the informal discussion behind the behaviour of semantics \citep{baroni2009semantics}. For example, the grounded semantics is considered more \textit{sceptical} for taking fewer committed choices and always providing a single \textit{extension}. By contrast, the preferred semantics is seen as a more \textit{credulous} approach for being more audacious when accepting arguments and has consequently been able to provide more than one extension. Still, Dung’s semantics and its variations are not the only class of semantics employed in abstract argumentation theory. Another well-known class of semantic is the ranking-based one \citep{amgoud2013ranking,amgoud2016ranking,cayrol2005graduality,matt2008game,Pu2014argument,Dondio2018Ranking}. In ranking-based semantics, the goal is not to 
provide an extension, but to rank arguments to define the most important one(s). It is a more flexible class of semantics, in the sense that arguments are not 
strictly rejected or accepted, but instead a graded assessment of arguments is provided, based on the topology of the argumentation framework.

Finally, in Layer 5, the assessment of the statements supported by acceptable arguments is performed. When performing defeasible inferences, for practical purposes, usually a single decision needs to be made or a single action performed. However, multiple acceptable arguments may be computed in the previous steps. Usually, these coincide with possible consistent points of view that can be simultaneously considered for describing the knowledge being modelled. In the case of extension-based semantics, extensions might contain multiple arguments, multiple extensions might be computed, or both. In the case of ranking-based semantics, multiple arguments might be ranked at the top -- a situation that can easily occur when multiple arguments are not attacked. Despite the suggestion in \citep{konieczny2015supported} of ranking the arguments by the number of extensions they 
belong to, this is not a problem usually addressed in the literature. Thus, in the present research, premises and conclusions of employed 
arguments are linked to categorical or numerical datasets. This allows for a simplified quantification of their values and aggregation in different fashions, 
namely average, sum or median. In turn, this aggregation produces a final inference (number), which is only possible due to the use of 
arguments built upon associated numerical datasets.

Some works make use of all the reviewed multi-layer structures (Fig. 
\ref{fig:multilayerstructure}) in their systems \citep{chang2009mixed,hunter2010argumentation,craven2012efficient}, whereas others do not
\citep{patkar2006evidence,glasspool2006argumentation,grando2013argumentation}.

Table \ref{table:workslayers} lists
several argument-based systems and their approaches for each layer in some prominent areas of argumentation. Some of 
these approaches may not have been previously reviewed, since they are beyond the scope of this study.

\begin{table}[!h]
%%\vspace{-3mm}
\centering
\footnotesize
\setlength{\tabcolsep}{2.5pt}
\renewcommand*{\arraystretch}{1.0}
   \caption{Examples of argument-based systems and their internal configurations across layers.}
\setlength{\leftmargini}{0mm}
   \label{table:workslayers}
    \begin{tabular}{C{0.6cm}C{2.0cm}C{1.7cm}C{1.7cm}C{1.7cm}C{1.5cm}C{1.7cm}} % Column formatting, @{} suppresses leading/trailing space
    \hline
    \textbf{Ref.} &  \textbf{Application} & \textbf{Layer 1} & \textbf{Layer 2} & \textbf{Layer 3} & \textbf{Layer 4} & \textbf{Layer 5}\\
    \hline
    
    \citep{hunter2010argumentation} & Treatment outcomes & Inference rule & Rebuttals & Preference list & Extens. based & Utility theory \\ \arrayrulecolor{gray!40}\hline
    
    \citep{craven2012efficient} & Healthcare & Natural language & Binary relation & Preference list & Preferred with pref. & User preference \\ \arrayrulecolor{gray!40}\hline
    
    %\citep{karacapilidis2001computer} & Collaborative decision-making & Natural language & Bipolarity & Preference relation & First-order logic & n/a\\ \arrayrulecolor{gray!40}\hline
    
    %\citep{ashley1991modeling} & Law & Factors + issues & n/a & n/a & Analogy to past cases & n/a \\ \arrayrulecolor{gray!40}\hline
    
    \citep{glasspool2006argumentation} & Care planning & Toulmin & n/a & n/a & n/a & n/a \\ \arrayrulecolor{gray!40}\hline

    %\citep{longo2013breast} & Cancer recurrence & Inference rule + fuzzy logic & Undercutting + rebutting & n/a & Extens. based & Extens. cardinality + 
%fuzzy \\ \arrayrulecolor{gray!40}\hline
    
    \citep{longo2014formalising} & Mental workload & Inference rule + fuzzy logic & Undercutting + rebutting & Strength of att. and arg. & 
Grounded + Preferred & Extens. cardinality + fuzzy logic\\ \arrayrulecolor{gray!40}\hline

    \citep{RizzoML18} & Healthcare & Standard logic & Undercutting + rebutting & Strength of arg. & 
Grounded, Preferred, Categoriser & Extens. cardinality + average\\ \arrayrulecolor{gray!40}\hline
    %\citep{atkinson2005dialogue} & Dialogue-game protocol & Argument schemes & Ad hoc list & n/a & Persuasion & n/a \\ \arrayrulecolor{gray!40}\hline
    
    %\citep{karunatillake2006argument} & Argument-based negotiation & Argument schemes & Undercutting + rebutting &  n/a & Negotiation & n/a \\ \arrayrulecolor{gray!40}\hline
    
    \citep{dondio2007presumptive} & Computational trust & Argument schemes & n/a & n/a & n/a & Sum aggregation function \\
\arrayrulecolor{black}\hline
\end{tabular}
\end{table}

\subsection{Comparison of Non-monotonic Formalisms}

A comprehensive comparison of different approaches to non-monotonic reasoning is a subject not sufficiently developed in the literature.
In particular, to the best of the authors’ knowledge, there is a very limited number of empirical works that incorporates the diversity of reasoning methodologies as done in this research article. Still, some works have proposed different comparisons among non-monotonic formalisms and are important to mention. For instance,
\citeauthor{brewka1997nonmonotonic} \cite{brewka1997nonmonotonic} provide a good overview in order to categorise non-monotonic logics by modal-preference logics, fixed-point logics and abductive methods. The recent work in \cite{hlobil2018} presents guidelines for selection of non-monotonic logics 
resulting in 17 different types of logics. \citeauthor{KONOLIGE1988343} \cite{KONOLIGE1988343} also studies the relation between non-monotonic logics, specifically between default and autoepistemic logic.

\citeauthor{delladio2006comparison} \cite{delladio2006comparison} investigate the relations between a normal default logic and a variant of defeasible logic programming.
The former is a special 
case in which justifications and consequents of default rules are the same, or in terms of default logic: $p(x):c(x)/c(x)$. The latter is a formalism that 
combines logic programming and defeasible argumentation, to allow the representation of defeasible and non-defeasible rules. 
The authors show an equivalence between the consequents from the normal default logic and
different answers given by the defeasible logic programming. Still, it investigates a theoretical relationship, limited by certain cases of the 
selected formalisms.

\citeauthor{dutilh2017reasoning} \cite{dutilh2017reasoning} make an empirical test on the accuracy of two formal non-monotonic reasoning models: preferential logic, a non-monotonic logic extended from a monotonic one; and screened belief revision, a particular version of belief revision theories \citep{gardenfors1992}. 
The experiment attempts to 
demonstrate which of the two formalisms can better predict belief bias, a form of human reasoning.
This examination is different from the investigation proposed in this research article, which instead assesses the inferential capacity of non-monotonic, knowledge-based reasoning approaches for quantitative inferences in real-world contexts.

More recently, \citeauthor{arieli2021logic} \cite{arieli2021logic} perform a comparative study among logic-based approaches to formal argumentation and a theoretical discussion about the relations of these and other non-monotonic reasoning formalisms. For instance, it describes how certain extensions of autoepistemic logics  \citep{moore1985semantical} and default logic  \cite{reiter1980logic} could be translated to an extension of assumption-based argumentation \cite{bondarenko1997abstract}, another formalism designed to capture and generalise non-monotonic reasoning. 

\citeauthor{yang2004comparison} \cite{yang2004comparison} compare first-order predicate logic, fuzzy logic and non-monotonic logic implemented through negation as failure. The methods were 
contrasted using a simulation approach in which experiment facts were considered as random numbers. In turn, a set of algorithms is provided for their investigated
transformations and modifications. 
This work possesses a very similar 
motivation to the present research article when comparing different reasoning formalisms, even if some are monotonic.
It attempts to evaluate the 
capacity of inference and the complexity of these distinct approaches. Yet, despite proposing an interesting mechanism for experimentation, the study does not 
evaluate the subject in real-world domains. Instead, the simulation approach seems to elucidate the performance of such methods only in a 
computationally-controlled environment, for instance when different types of transformation must be applied to numerical datasets.

\subsection{Computational Trust Modelling}\label{computationaltrust}

Computational trust modelling, a knowledge-representation and reasoning problem, has been selected in order to allow the comparison of defeasible argumentation with other similar reasoning techniques.
It has been investigated by several disciplines from different perspectives, such as psychology, sociology, and philosophy \citep{Sabater2005Trust}. 
Many definitions of trust can be found in the literature \citep{parsons2010reasoning}. Briefly, it can be described as a prediction that a trusted entity will bring to completion the expectations of a trustier in some specific context.
The first computational model of trust was proposed in \citep{marsh1994trust}. Its goal was to enable artificial agents to make trust-based decisions in the 
domain of Distributed Artificial Intelligence.
The modelling of computational trust has also several applications in digital systems, for instance: reputation management 
\citep{yashkina2019expressing,melnikov2018towards}; 
social search and collective intelligence \citep{Longo2009information,longo2010enhancing}; user behaviour modelling \citep{Longo09Toward}; and self-adaptive 
recommendations \citep{luca2009enabling,dondio2008translation}.

Several works have examined the relation between argumentation and computational trust.
For instance, \citeauthor{Toni2010} \cite{Toni2010} review how argumentation can help agents make decisions. It also discusses how arguments can improve the assessment of the trustworthiness of certain agents by supporting predictions on these agents' future behaviours. In turn,
\citeauthor{parsons2014trust} \cite{parsons2014trust} present a set of argument schemes for reasoning about trust. It is aimed at providing a computational mechanism for establishing arguments about trustworthiness. These schemes are also followed by a set of critical questions that can rule out their use.
Other works have also focused on defining argument-based approaches for reasoning about trust \cite{amgoud2014argumentation,Tang2012}. 

In this research article, the context under evaluation comes from the Wikipedia project. 
Collaborative, user-generated content is of
essential importance in the web. Hence, sites such as Wikipedia, TripAdvisor and Flickr
leverage the interest and contribution of people all over the world. The drawback comes from
the discrepant origin and quality of such contributions, leading to the complication of visitors and content moderators assessing their reliability.
Wikipedia itself is under continual change from different sources, ranging from domain experts and to casual contributors, to vandals and committed editors.
Therefore, many works have investigated the problem of computing the trust of Wikipedia editors or articles.
For instance, \citeauthor{Adler2007trust} \cite{Adler2007trust} present a 
content-driven reputation system for Wikipedia editors, assuming that the reputation of editors can be used as a rough guide to the trust assigned to articles 
edited by them. In turn, reputation is assigned according to the longevity of the text inserted, and the longevity of the text edited by each editor. In a 
subsequent work, \citeauthor{Adler2008assigning} \cite{Adler2008assigning} compute the trust of a word in a Wikipedia article according to the reputation of the original editor of the word, as 
well as the reputation of editors who edited content in the vicinity of the word. The study demonstrates that text labelled as high trust has a significantly 
lower chance of being edited in the future. Similarly, \citeauthor{zeng2006computing} \cite{zeng2006computing} explore the revision history of an article to assess the trustworthiness of the 
article through a dynamic Bayesian network. 
In short, other works evaluate the trust of Wikipedia’s contributors through 
a multi-agent trust model \cite{Krupa2009trust} and the Wikipedia editor reputation through the stability of content inserted
\cite{Javanmardi2010reputation}.

To conclude, let us point out that the proposed use of defeasible argumentation and other reasoning approaches in this research article is \textit{not} aimed at enhancing the assessment of computational trust. Hence, the performed experiments are not compared with the aforementioned works. 
Nonetheless, to the best of the author's knowledge, the use of non-monotonic reasoning, instantiated by quantitative information, for the inference of trust of Wikipedia editors has only been attempted in previous works \cite{RizzoDL20, rizzo2020thesis}. These employed different sets of data and/or reasoning approaches. Thus, the investigation proposed in this research article is a more comprehensive one, extending these and other previous works \cite{rizzo2020empirical,rizzo2018investigation,RizzoML18} that have compared knowledge-based, non-monotonic reasoning approaches applied in different domains of application.
Therefore, the main goal of this research article is to enhance the generalisability of defeasible argumentation as an effective approach to reason with quantitative, uncertain and conflicting information in real-world contexts. The next section provides a precise description of the research problem, the formulated hypothesis and the methods applied to test it.

% Design section

\section{Design and Methodology}\label{designmethodology}

The research problem being addressed is how defeasible argumentation compares to similar reasoning approaches, such as expert systems and fuzzy reasoning,
when used for the formalisation of non-monotonic reasoning models of inference. Moreover, this paper focuses on the case in which such models can be instantiated by quantitative data from real-world domains. Most recently, there seems to be an increase in the use of defeasible argumentation as the basis of current
models employed in practice. Therefore, the assumption is that this approach could also be more
suitable for modelling non-monotonic reasoning and producing non-monotonic reasoning models of inference in the domain of computational trust.
The confirmation of this assumption would reinforce the applicability and generalisability of defeasible argumentation 
with quantitative data in real-world contexts. Consequently, it could also aid other scholars to adopt the proposed approach in other domains of applications. To investigate this, an \textit{inductive type of research} is proposed – that
is, one which attempts to propose broader generalisations from specific observations. An
observation comes from the recent increased use of argument-based models in fields such
as health care, knowledge-representation and reasoning, and multi-agent systems.
This
leads to the following hypothesis: 

\begin{hypothesis} If computational trust is modelled with defeasible argumentation, 
then the inferential capacity of its models will be superior than that achieved by non-monotonic fuzzy reasoning and expert systems models
according to a predefined set of 
evaluation metrics from the domain of application. 
\end{hypothesis}

\begin{figure}[!h]
%\begin{wrapfigure}{l}{0.6\textwidth}
\centering
\small
 \scalebox{1}{
\begin{tikzpicture}[
	node distance=10mm,
	desc/.style={
		rectangle,
		rounded corners,
		draw=black, 
		 thin,
		text centered,
		minimum height=2mm,
		fill=blue!		},
    model/.style={
		circle,
		%rounded corners,
		draw=black, 
		 thin,
		text centered,
		fill=blue!10
		}
	every node/.style={font=\sffamily}
]

\node [desc, fill=none, dotted, text width=11.7cm, minimum height=15mm] (Z1) at (-2, 0)  {};
\node [desc, text width=5.0cm, fill=white] (A) at (-5.0,0) {Knowledge-base 1 (\ref{app:trust})};
\node [desc, text width=5.0cm, fill=white] (A2) at (1.0,0) {Knowledge-base 2 (\ref{app:trust})};

\node [desc, fill=none, dotted, text width=11.7cm, minimum height=40mm] (Z2) at (-2, -4)  {};
\node [desc, fill=gray!30, text width=3.1cm, minimum height=5mm] (es) at (-6,-4.0) {\textbf{Expert system \\ models}\\
1. IF-THEN rules\\ 2. Inference engine\\ 3. Rules' aggregation};

\node [desc, fill=gray!30, text width=3cm, minimum height=5mm] (fr) at (-2.5,-4.0) {\textbf{Fuzzy reasoning \\ models}\\
1. Fuzzification\\ 2. Inference engine\\ 3. Defuzzification};

\node [desc, fill=gray!30, text width=4cm, minimum height=5mm] (at) at (1.5,-4.0) {\textbf{Argument-based models}\\
1. Structure of arguments\\ 2. Conflicts of arguments\\ 3. Evaluation of 
conflicts\\ 4. Acceptance status\\ 5. Accrual of  arguments};

%\node [desc, text width=2.3cm, fill=white, minimum height=7.5mm] (B) at (8,-4) {Dataset};
%\node [desc,fill=blue!0, draw=none, text width=1.6cm, font=\footnotesize] (G) at (4,-5)  {Instantiation of models};

\node [desc, fill=none, dotted, text width=11.0cm, minimum height=15mm] (Z3) at (-2, -8.5)  {};
\node [desc, fill=gray!10, text width=3cm, minimum height=5mm] (D1) at (-5.5,-8.5) {Expert system models' inferences};
\node [desc, fill=gray!10, text width=3cm, minimum height=5mm] (D2) at (-2.0,-8.5) {Fuzzy reasoning models' inferences};
\node [desc, fill=gray!10, text width=3cm, minimum height=5mm] (D3) at (1.5,-8.5) {Argument-based models' inferences};

%\node [desc, fill=gray!10, text width=3cm, minimum height=5mm] (D2) at (8,-8.0) {Inferences through average and weighted average of features};

\node [desc,fill=blue!0, draw=none, text width=4.2cm, font=\footnotesize] (G) at (-2.0,-11.0)  {\textbf{Comparison of inferences}};

\node [rectangle, fill=cyan!0, text width=0.0cm, font=\scriptsize] (K1) at (-5.5,-11)  {};
\node [rectangle, text width=0.0cm, font=\scriptsize] (K3) at (-2,-10.8)  {};
\node [rectangle, text width=0.0cm, font=\scriptsize] (K4) at (1.5,-11)  {};
\node [rectangle, fill=cyan!0, text width=0.0cm, font=\scriptsize] (K2) at (8,-11)  {};

\path
(D1) [-, dashed] edge node {}  (K1)
(D2) [-, dashed] edge node {}  (K3)
(D3) [-, dashed] edge node {}  (K4)
(K1) [-, dashed] edge node {}  (G)
(K4) [-, dashed] edge node {}  (G)

%(D2) [-, dashed] edge node {}  (K2)
%(K2) [-, dashed] edge node {}  (G)
;

\path 
%(A) [->] edge node {}  (at)
%(A2) [->] edge node {}  (at)
(Z1) [->] edge node {}  (Z2)
(Z2) [->] edge node {}  (Z3)
% (B) [->] edge node {}  (at)
% (B) [->] edge node {}  (D2)
%(at) [->] edge node {}  (D)
;

\node [circle, draw=black,text width=0.0cm, font=\scriptsize] (K3) at (-2.0,-1.3)  {};
\node [circle, text width=4.5cm, font=\scriptsize] (K5) at (2.0,-1.3)  {Design of non-monotonic reasoning \\ models from domain knowledge};

\node [circle, draw=black,text width=0.0cm, font=\scriptsize] (K6) at (-2.0,-7.0)  {};
\node [circle, text width=5cm, font=\scriptsize] (K7) at (1.4,-7.0)  {Instantiation of models with \\ quantitative, real-world datasets \\ from the domain of computational trust};

\path
(K3) [->, dashed] edge node {}  (K5)
(K6) [->, dashed] edge node {}  (K7)
;

\end{tikzpicture}
}
%\caption{Design and evaluation strategy schema.}
\captionsetup{type=figure,labelsep=space,font=small,labelfont=bf}
\captionof{figure}{Design of a comparative empirical research study and its evaluation schema aimed at evaluating the inferential capacity of defeasible argumentation.}
\label{fig:design}

\end{figure}
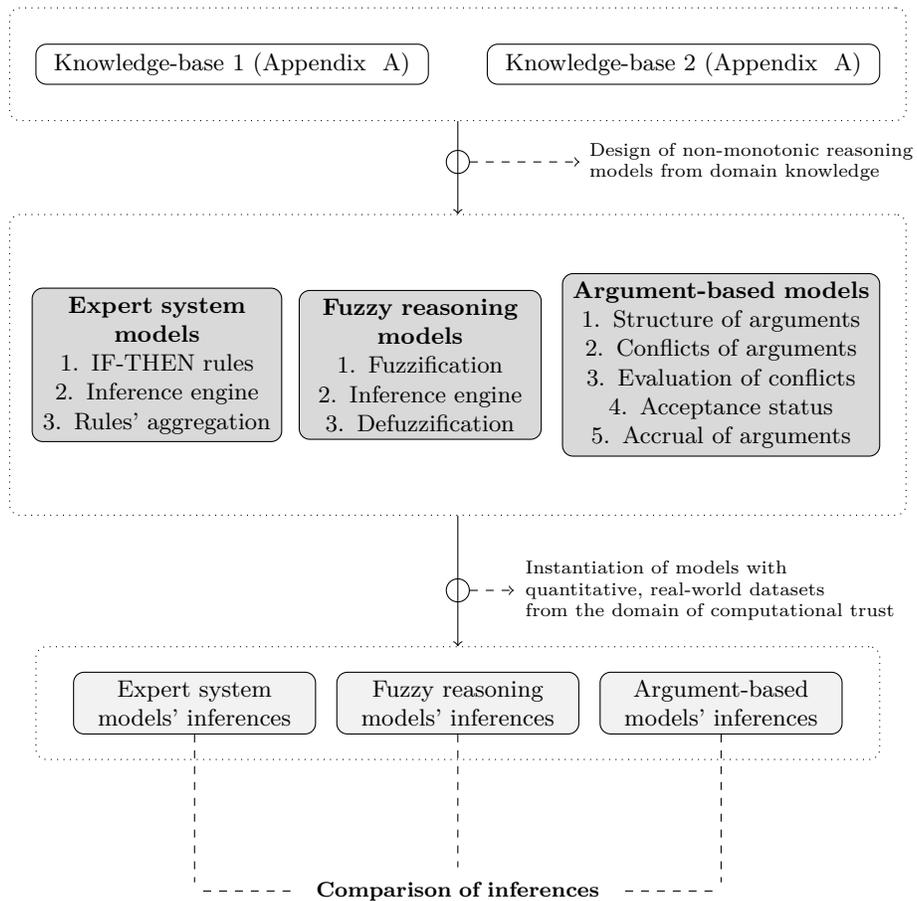

To test this hypothesis, non-monotonic reasoning models of inference are designed and built with the pre-existing theoretical 
knowledge of the investigated reasoning approaches, as reviewed in Section \ref{sec:literature_review}. 
The goal is to design non-monotonic reasoning models capable of 
assigning a trust value in the range [0, 1] $\subset\mathbb{R}$ to Wikipedia editors. One means complete trust should be assigned to an editor, while 0 means an absence of trust assigned to the editor.
In turn, such models are instantiated with real-world data, 
allowing a statistical comparison of produced inferences.
Fig. \ref{fig:design} depicts the designed 
experiment with the evaluation phases incorporated into the flow.
First, knowledge bases structured around natural language terms are employed by the 
non-monotonic reasoning approaches for the design of inferential models. These same models are later instantiated by real-world 
datasets from the domain of computational trust, producing three sets of inferences, one for each reasoning approach. These sets are subsequently analysed for the comparison of the 
inferential capacity of the reasoning approaches. 
This comparison is done by assessing the values assigned to Wikipedia Barnstar editors. A Barnstar\footnote{\url{https://en.wikipedia.org/wiki/Wikipedia:Barnstars}} represents an award used by Wikipedia to recognise valuable editors.
It is a non-automatic award bestowed from a Wikipedia editor to another Wikipedia editor. Therefore, it is not a ground truth for trust. Instead, it is used as a proxy measure to identify trustworthy editors and to allow the selection of evaluation metrics later detailed in this section.

\subsection{Datasets Employed and Knowledge Bases Design}

Wikipedia makes all its data available for download through HTML or XML dumps\footnote{\url{https://dumps.wikimedia.org/}}, including articles, articles’ history and 
complete text data. Hundreds of different language editions are available for download. Since no natural language information is 
analysed, but only quantitative data related to editors, the XML dump of the Portuguese-language\footnote{File \texttt{ptwiki-20200820-stub-meta-history.xml}} edition and the XML dump of the Italian-language\footnote{File \texttt{itwiki-20200801-stub-meta-history.xml}} edition of the Wikipedia were selected for examination and downloaded on 8 January 2020. These were selected mainly due their respective sizes that were more appropriate to the computational resources available. Moreover, it was expected that two sets of data would be able to reinforce findings and confirm possible observed differences in the inferences produced by designed non-monotonic reasoning models.
According to the 
Wikimedia Foundation’s Analytics,\footnote{\url{https://stats.wikimedia.org}}, the Portuguese file contained $999,696$ pages created (excluding pages being redirects), $1,947,023$ editors, and $133$ Barnstar editors, while the Italian file contained $1,576,621$ articles, $2,804,142$ editors, and $106$ Barnstar editors, both up to January 2020. 
Each dumped Wikipedia page is identified by its title and it has a number of associated revisions containing:
i) its own ID;
ii) a time stamp;
iii) a contributor (editor) identified by a user name or IP address if anonymous;
iv) an optional commentary left by the editor;
v) the current number of bytes of the page on 
current revision; 
vi) an optional tag  indicating whether the revision is minor or  major and should be reviewed by other editors.
Fig. \ref{fig:xmlstructure} (p. \pageref{fig:xmlstructure}) depicts an example of the XML structure.
This data was extracted for the definition of features listed in Table 
\ref{table:wikifeatures} for each editor, resulting in two datasets\footnote{\url{https://doi.org/10.6084/m9.figshare.14939319.v1}}. Temporal factors such as presence, regularity and frequency factor were first proposed in \cite{longo2007temporal}. A time window of 30 days was selected for evaluation of the frequency and regularity factors, similarly to the statistical 
analyses performed by the Wikimedia Foundation.

\begin{table}[!h]
\footnotesize
 \setlength{\tabcolsep}{4pt}
     \renewcommand*{\arraystretch}{1}
   \caption{Summary of features employed by a human reasoner for trust assessment.}
\label{table:wikifeatures}   
\centering
\begin{tabular}{lL{9.4cm}}
\hline
 \textbf{Feature} & \textbf{Description}
\\
\arrayrulecolor{black}\hline
Pages & Number of unique pages ($\mathbb{N}_{\geq 0}$) edited by the user.\\
\arrayrulecolor{gray!40}\hline
Activity  & Number of edits ($\mathbb{N}_{\geq 0}$) performed by the user.\\
\arrayrulecolor{gray!40}\hline
Anonymity & Categorical value (Yes [1], No [0]) indicating whether the user is anonymous or not. Anonymous users are identified by their IP.\\
\arrayrulecolor{gray!40}\hline
Not Minor & Ratio [0, 1] $\subset\mathbb{R}$ of edits flagged by the own editor for revision. 1 (0) means all (no) edits of the editor flagged by him or herself as not minor. \\
\arrayrulecolor{gray!40}\hline
Comments & Ratio [0, 1] $\subset\mathbb{R}$ of edits in which a comment was included. One comment allowed per edition. \\
\arrayrulecolor{gray!40}\hline
Presence & Ratio [0, 1] $\subset\mathbb{R}$ between the registration date of the user and the date of the beginning of the system (January 2001).\\
\arrayrulecolor{gray!40}\hline
Frequency & Frequency ratio [0, 1] $\subset\mathbb{R}$ of edits per time window of 30 days in the editor's life cycle. Maximum value limited at 1. \\
\arrayrulecolor{gray!40}\hline
Regularity & Regularity ratio [0, 1] $\subset\mathbb{R}$ per time window of 30 days. 1 means at least one interaction every 30 days in the editor's life cycle. \\
\arrayrulecolor{gray!40}\hline
Bytes & Overall integer number of bytes edited by the user. Insertions/ deletions respectively increase/decrease the amount of bytes. \\
\arrayrulecolor{black}\hline
\end{tabular}
\end{table}

The set of extracted features was employed for the construction of two knowledge bases with the author’s knowledge and intuition in this domain. 
\ref{app:trust} (p. 
\pageref{app:trust}) 
lists the information contained in them.
The set of IF-THEN rules constructed was the same for both knowledge bases. They were defined intuitively and with the aid of external sources. For instance, 
the numerical ranges associated with natural language terms employed to describe \texttt{activity factor} and \texttt{bytes} were defined with the aid of the 
Wikimedia Foundation’s Analytics. According to the reports of this foundation, an editor is considered a 
contributor if he/she has made more than 10 editions in his/her life cycle. Hence an \texttt{activity factor} $\ge 10$ was used to infer \textit{medium high} 
trust, while 
\texttt{activity 
factor} $\ge 20$ was used to infer \textit{high} trust. Similarly, the last report on the mean size of articles in the Portuguese Wikipedia showed a mean of 
2,388 bytes per article, while 90\% of articles had more than 512 bytes. This information was used to intuitively infer \textit{high} (\textit{medium high})  
if an editor had contributed at least 2,388 bytes (512 bytes) throughout all his/her editions. Other features were normalised in the range [0, 1] 
$\subset\mathbb{R}$, in order to provide a more standard reasoning process. For instance, the feature \texttt{not minor} was divided by the \texttt{activity 
factor}, 
hence providing the percentage of editions flagged as major for   all editions of the editor. Features normalised in the range [0, 1] $\subset\mathbb{R}$ were then 
described by four natural language terms: \textit{low}, \textit{medium low}, \textit{medium high} and \textit{high}. Table \ref{table:featurestransf} (p. 
\pageref{table:featurestransf}) lists the feature transformations employed, the associated natural language terms and their respective numerical ranges. It is 
important to highlight that the constructed IF-THEN rules are limited by the type of data employed. While Wikipedia provides the text history of all its 
articles’ editions, no natural language data is exploited in this study. A knowledge base also formed by features that take advantage of natural language data 
would likely contain stronger information for the inference of the editors' trust.

Following this, a set of contradictions among IF-THEN rules was defined. This process was made in two different ways, resulting in two different knowledge bases.
The first was made by means of an intuitive manner, trying to establish evident relationships, such as low \texttt{frequency factor} contradicting 
the use of high \texttt{presence factor} to infer high trust. In other words, high \texttt{presence factor} was interpreted as an indication of high trust, unless the same editor also had a low \texttt{frequency factor}.
A set of premises was also considered for the definition of agents whose trust should be low, such as a vandal or a bot.
For example, an editor who is 
anonymous, 
has a low number of \texttt{comments}, a very low number of \texttt{not minor} edits, a high number of \texttt{pages} edited, and a high number of \texttt{bytes} inserted 
was considered a bot. In other words, this set of characteristics was considered sufficient to assume that an editor was a bot. In turn, this set of 
premises was used to
contradict several IF-THEN rules inferring high trust. This full knowledge base is reported in \ref{app:trust}, with the resulting graphical 
representation depicted in Fig. \ref{figure:kb6}.

\begin{figure}[!h]
%\begin{wrapfigure}{l}{0.6\textwidth}
\centering
 \scalebox{1}{
\tikzset{every picture/.style={line width=0.75pt}} %set default line width to 0.75pt        

\begin{tikzpicture}[x=0.75pt,y=0.75pt,yscale=-1,xscale=1]
%uncomment if require: \path (0,315); %set diagram left start at 0, and has height of 315

%Shape: Rectangle [id:dp46456375474633527] 
\draw   (71,85) -- (344,85) -- (344,124.43) -- (71,124.43) -- cycle ;
%Straight Lines [id:da40402405109166195] 
\draw    (103,86.33) -- (103,124.43) ;
%Straight Lines [id:da6956119526613649] 
\draw    (133,86.33) -- (133,124.43) ;
%Straight Lines [id:da8934016409972956] 
\draw    (162,85.43) -- (162,124.43) ;
%Straight Lines [id:da03965160511155985] 
\draw    (192,85.43) -- (192,124.43) ;
%Straight Lines [id:da17819716195816793] 
\draw    (222,85.43) -- (222,124.43) ;
%Straight Lines [id:da37483508428115364] 
\draw    (251,85.43) -- (251,124.43) ;
%Straight Lines [id:da458892327156879] 
\draw    (281,85.43) -- (281,124.43) ;
%Straight Lines [id:da2842125940800413] 
\draw    (312,85.21) -- (312,124.21) ;
%Shape: Rectangle [id:dp7005771159069016] 
\draw   (70,186) -- (343,186) -- (343,225.43) -- (70,225.43) -- cycle ;
%Straight Lines [id:da34793054001886303] 
\draw    (102,187.33) -- (102,225.43) ;
%Straight Lines [id:da28420753895143647] 
\draw    (132,187.33) -- (132,225.43) ;
%Straight Lines [id:da37507384355732065] 
\draw    (161,186.43) -- (161,225.43) ;
%Straight Lines [id:da4682704653341956] 
\draw    (191,186.43) -- (191,225.43) ;
%Straight Lines [id:da330463798478442] 
\draw    (221,186.43) -- (221,225.43) ;
%Straight Lines [id:da5244485909270711] 
\draw    (250,186.43) -- (250,225.43) ;
%Straight Lines [id:da2683531746327479] 
\draw    (280,186.43) -- (280,225.43) ;
%Straight Lines [id:da7028819848801213] 
\draw    (311,186.21) -- (311,225.21) ;
%Straight Lines [id:da8592249387831141] 
\draw    (144.33,104) -- (111.95,204.43) ;
\draw [shift={(111.33,206.33)}, rotate = 287.87] [color={rgb, 255:red, 0; green, 0; blue, 0 }  ][line width=0.75]    (10.93,-3.29) .. controls (6.95,-1.4) and (3.31,-0.3) .. (0,0) .. controls (3.31,0.3) and (6.95,1.4) .. (10.93,3.29)   ;
%Straight Lines [id:da9212685688100939] 
\draw    (204.67,105.67) -- (118.95,207.8) ;
\draw [shift={(117.67,209.33)}, rotate = 310] [color={rgb, 255:red, 0; green, 0; blue, 0 }  ][line width=0.75]    (10.93,-3.29) .. controls (6.95,-1.4) and (3.31,-0.3) .. (0,0) .. controls (3.31,0.3) and (6.95,1.4) .. (10.93,3.29)   ;
%Straight Lines [id:da27081809820661085] 
\draw    (262.33,105.67) -- (125.22,217.07) ;
\draw [shift={(123.67,218.33)}, rotate = 320.90999999999997] [color={rgb, 255:red, 0; green, 0; blue, 0 }  ][line width=0.75]    (10.93,-3.29) .. controls (6.95,-1.4) and (3.31,-0.3) .. (0,0) .. controls (3.31,0.3) and (6.95,1.4) .. (10.93,3.29)   ;
%Straight Lines [id:da1635928215251672] 
\draw    (175.67,105.33) -- (116.02,205.61) ;
\draw [shift={(115,207.33)}, rotate = 300.74] [color={rgb, 255:red, 0; green, 0; blue, 0 }  ][line width=0.75]    (10.93,-3.29) .. controls (6.95,-1.4) and (3.31,-0.3) .. (0,0) .. controls (3.31,0.3) and (6.95,1.4) .. (10.93,3.29)   ;
%Straight Lines [id:da13205441527856698] 
\draw    (294,103.67) -- (122.7,209.95) ;
\draw [shift={(121,211)}, rotate = 328.18] [color={rgb, 255:red, 0; green, 0; blue, 0 }  ][line width=0.75]    (10.93,-3.29) .. controls (6.95,-1.4) and (3.31,-0.3) .. (0,0) .. controls (3.31,0.3) and (6.95,1.4) .. (10.93,3.29)   ;
%Straight Lines [id:da8279914211256933] 
\draw    (233.33,105.67) -- (123.76,212.93) ;
\draw [shift={(122.33,214.33)}, rotate = 315.61] [color={rgb, 255:red, 0; green, 0; blue, 0 }  ][line width=0.75]    (10.93,-3.29) .. controls (6.95,-1.4) and (3.31,-0.3) .. (0,0) .. controls (3.31,0.3) and (6.95,1.4) .. (10.93,3.29)   ;
%Straight Lines [id:da877864590437397] 
\draw    (87.5,107.5) -- (107.92,205.04) ;
\draw [shift={(108.33,207)}, rotate = 258.17] [color={rgb, 255:red, 0; green, 0; blue, 0 }  ][line width=0.75]    (10.93,-3.29) .. controls (6.95,-1.4) and (3.31,-0.3) .. (0,0) .. controls (3.31,0.3) and (6.95,1.4) .. (10.93,3.29)   ;
%Straight Lines [id:da7933306502696342] 
\draw    (148,106.33) -- (325.93,207.35) ;
\draw [shift={(327.67,208.33)}, rotate = 209.57999999999998] [color={rgb, 255:red, 0; green, 0; blue, 0 }  ][line width=0.75]    (10.93,-3.29) .. controls (6.95,-1.4) and (3.31,-0.3) .. (0,0) .. controls (3.31,0.3) and (6.95,1.4) .. (10.93,3.29)   ;
%Straight Lines [id:da9156924666919506] 
\draw    (148,106.33) -- (206.36,211.58) ;
\draw [shift={(207.33,213.33)}, rotate = 240.99] [color={rgb, 255:red, 0; green, 0; blue, 0 }  ][line width=0.75]    (10.93,-3.29) .. controls (6.95,-1.4) and (3.31,-0.3) .. (0,0) .. controls (3.31,0.3) and (6.95,1.4) .. (10.93,3.29)   ;
%Straight Lines [id:da6071204398123784] 
\draw    (148,104.67) -- (264.84,208.67) ;
\draw [shift={(266.33,210)}, rotate = 221.67000000000002] [color={rgb, 255:red, 0; green, 0; blue, 0 }  ][line width=0.75]    (10.93,-3.29) .. controls (6.95,-1.4) and (3.31,-0.3) .. (0,0) .. controls (3.31,0.3) and (6.95,1.4) .. (10.93,3.29)   ;
%Straight Lines [id:da08959095002413942] 
\draw    (144.33,104) -- (174.14,213.4) ;
\draw [shift={(174.67,215.33)}, rotate = 254.76] [color={rgb, 255:red, 0; green, 0; blue, 0 }  ][line width=0.75]    (10.93,-3.29) .. controls (6.95,-1.4) and (3.31,-0.3) .. (0,0) .. controls (3.31,0.3) and (6.95,1.4) .. (10.93,3.29)   ;
%Straight Lines [id:da5735136471286413] 
\draw    (148,104.67) -- (291.4,212.8) ;
\draw [shift={(293,214)}, rotate = 217.02] [color={rgb, 255:red, 0; green, 0; blue, 0 }  ][line width=0.75]    (10.93,-3.29) .. controls (6.95,-1.4) and (3.31,-0.3) .. (0,0) .. controls (3.31,0.3) and (6.95,1.4) .. (10.93,3.29)   ;
%Straight Lines [id:da9201751411199939] 
\draw    (148,104.67) -- (236.07,212.45) ;
\draw [shift={(237.33,214)}, rotate = 230.75] [color={rgb, 255:red, 0; green, 0; blue, 0 }  ][line width=0.75]    (10.93,-3.29) .. controls (6.95,-1.4) and (3.31,-0.3) .. (0,0) .. controls (3.31,0.3) and (6.95,1.4) .. (10.93,3.29)   ;
%Straight Lines [id:da20768505943266558] 
\draw    (148,106.33) -- (89,208.77) ;
\draw [shift={(88,210.5)}, rotate = 299.94] [color={rgb, 255:red, 0; green, 0; blue, 0 }  ][line width=0.75]    (10.93,-3.29) .. controls (6.95,-1.4) and (3.31,-0.3) .. (0,0) .. controls (3.31,0.3) and (6.95,1.4) .. (10.93,3.29)   ;
%Straight Lines [id:da6055364160193479] 
\draw    (117.67,105) -- (323.22,210.75) ;
\draw [shift={(325,211.67)}, rotate = 207.22] [color={rgb, 255:red, 0; green, 0; blue, 0 }  ][line width=0.75]    (10.93,-3.29) .. controls (6.95,-1.4) and (3.31,-0.3) .. (0,0) .. controls (3.31,0.3) and (6.95,1.4) .. (10.93,3.29)   ;
%Straight Lines [id:da8576341895187507] 
\draw    (324.33,107.33) -- (126.73,220.67) ;
\draw [shift={(125,221.67)}, rotate = 330.15999999999997] [color={rgb, 255:red, 0; green, 0; blue, 0 }  ][line width=0.75]    (10.93,-3.29) .. controls (6.95,-1.4) and (3.31,-0.3) .. (0,0) .. controls (3.31,0.3) and (6.95,1.4) .. (10.93,3.29)   ;
%Straight Lines [id:da3769642839278835] 
\draw    (178,106.33) -- (328.91,202.72) ;
\draw [shift={(330.6,203.8)}, rotate = 212.57] [color={rgb, 255:red, 0; green, 0; blue, 0 }  ][line width=0.75]    (10.93,-3.29) .. controls (6.95,-1.4) and (3.31,-0.3) .. (0,0) .. controls (3.31,0.3) and (6.95,1.4) .. (10.93,3.29)   ;
%Straight Lines [id:da808772445484562] 
\draw    (295.33,105.33) -- (210.26,210.78) ;
\draw [shift={(209,212.33)}, rotate = 308.9] [color={rgb, 255:red, 0; green, 0; blue, 0 }  ][line width=0.75]    (10.93,-3.29) .. controls (6.95,-1.4) and (3.31,-0.3) .. (0,0) .. controls (3.31,0.3) and (6.95,1.4) .. (10.93,3.29)   ;
%Straight Lines [id:da6524296696296763] 
\draw    (114.67,105.33) -- (263.71,212.5) ;
\draw [shift={(265.33,213.67)}, rotate = 215.72] [color={rgb, 255:red, 0; green, 0; blue, 0 }  ][line width=0.75]    (10.93,-3.29) .. controls (6.95,-1.4) and (3.31,-0.3) .. (0,0) .. controls (3.31,0.3) and (6.95,1.4) .. (10.93,3.29)   ;
%Straight Lines [id:da500443342662618] 
\draw    (266.33,106.67) -- (299.98,209.5) ;
\draw [shift={(300.6,211.4)}, rotate = 251.88] [color={rgb, 255:red, 0; green, 0; blue, 0 }  ][line width=0.75]    (10.93,-3.29) .. controls (6.95,-1.4) and (3.31,-0.3) .. (0,0) .. controls (3.31,0.3) and (6.95,1.4) .. (10.93,3.29)   ;
%Straight Lines [id:da8721733416673234] 
\draw    (299.67,107.67) -- (271.22,206.08) ;
\draw [shift={(270.67,208)}, rotate = 286.12] [color={rgb, 255:red, 0; green, 0; blue, 0 }  ][line width=0.75]    (10.93,-3.29) .. controls (6.95,-1.4) and (3.31,-0.3) .. (0,0) .. controls (3.31,0.3) and (6.95,1.4) .. (10.93,3.29)   ;
%Straight Lines [id:da4248666357380253] 
\draw    (263.67,107) -- (239.13,210.72) ;
\draw [shift={(238.67,212.67)}, rotate = 283.31] [color={rgb, 255:red, 0; green, 0; blue, 0 }  ][line width=0.75]    (10.93,-3.29) .. controls (6.95,-1.4) and (3.31,-0.3) .. (0,0) .. controls (3.31,0.3) and (6.95,1.4) .. (10.93,3.29)   ;
%Straight Lines [id:da30698339539415165] 
\draw    (235.67,105.67) -- (268.05,206.1) ;
\draw [shift={(268.67,208)}, rotate = 252.13] [color={rgb, 255:red, 0; green, 0; blue, 0 }  ][line width=0.75]    (10.93,-3.29) .. controls (6.95,-1.4) and (3.31,-0.3) .. (0,0) .. controls (3.31,0.3) and (6.95,1.4) .. (10.93,3.29)   ;
%Straight Lines [id:da8882809275580503] 
\draw    (207.5,104.71) -- (176.88,213.41) ;
\draw [shift={(176.33,215.33)}, rotate = 285.74] [color={rgb, 255:red, 0; green, 0; blue, 0 }  ][line width=0.75]    (10.93,-3.29) .. controls (6.95,-1.4) and (3.31,-0.3) .. (0,0) .. controls (3.31,0.3) and (6.95,1.4) .. (10.93,3.29)   ;
%Straight Lines [id:da9662850205658027] 
\draw    (297.67,106.67) -- (240.62,212.57) ;
\draw [shift={(239.67,214.33)}, rotate = 298.31] [color={rgb, 255:red, 0; green, 0; blue, 0 }  ][line width=0.75]    (10.93,-3.29) .. controls (6.95,-1.4) and (3.31,-0.3) .. (0,0) .. controls (3.31,0.3) and (6.95,1.4) .. (10.93,3.29)   ;
%Straight Lines [id:da6479020959712256] 
\draw    (239,106.67) -- (294.06,210.57) ;
\draw [shift={(295,212.33)}, rotate = 242.07999999999998] [color={rgb, 255:red, 0; green, 0; blue, 0 }  ][line width=0.75]    (10.93,-3.29) .. controls (6.95,-1.4) and (3.31,-0.3) .. (0,0) .. controls (3.31,0.3) and (6.95,1.4) .. (10.93,3.29)   ;
%Straight Lines [id:da33682557441926675] 
\draw    (114,108) -- (233.9,214.07) ;
\draw [shift={(235.4,215.4)}, rotate = 221.5] [color={rgb, 255:red, 0; green, 0; blue, 0 }  ][line width=0.75]    (10.93,-3.29) .. controls (6.95,-1.4) and (3.31,-0.3) .. (0,0) .. controls (3.31,0.3) and (6.95,1.4) .. (10.93,3.29)   ;

% Text Node
\draw (71.82,73.58) node [anchor=north west][inner sep=0.75pt]  [rotate=-312.61,xslant=0] [align=left] {Pages};
% Text Node
\draw (102.82,73.58) node [anchor=north west][inner sep=0.75pt]  [rotate=-312.61,xslant=0] [align=left] {Activities};
% Text Node
\draw (133.82,71.58) node [anchor=north west][inner sep=0.75pt]  [rotate=-312.61,xslant=0] [align=left] {Anonymity};
% Text Node
\draw (163.82,72.58) node [anchor=north west][inner sep=0.75pt]  [rotate=-312.61,xslant=0] [align=left] {Not minor};
% Text Node
\draw (194.82,74.58) node [anchor=north west][inner sep=0.75pt]  [rotate=-312.61,xslant=0] [align=left] {Comments};
% Text Node
\draw (223.82,74.58) node [anchor=north west][inner sep=0.75pt]  [rotate=-312.61,xslant=0] [align=left] {Presence};
% Text Node
\draw (252.82,74.58) node [anchor=north west][inner sep=0.75pt]  [rotate=-312.61,xslant=0] [align=left] {Frequency};
% Text Node
\draw (285.82,74.58) node [anchor=north west][inner sep=0.75pt]  [rotate=-312.61,xslant=0] [align=left] {Regularity};
% Text Node
\draw (317.82,73.58) node [anchor=north west][inner sep=0.75pt]  [rotate=-312.61,xslant=0] [align=left] {Bytes};
% Text Node
\draw (50.82,257.58) node [anchor=north west][inner sep=0.75pt]  [rotate=-312.61,xslant=0] [align=left] {Pages};
% Text Node
\draw (71.82,267.58) node [anchor=north west][inner sep=0.75pt]  [rotate=-312.61,xslant=0] [align=left] {Activities};
% Text Node
\draw (94.82,271.58) node [anchor=north west][inner sep=0.75pt]  [rotate=-312.61,xslant=0] [align=left] {Anonymity};
% Text Node
\draw (124.82,272.58) node [anchor=north west][inner sep=0.75pt]  [rotate=-312.61,xslant=0] [align=left] {Not minor};
% Text Node
\draw (147.82,275.58) node [anchor=north west][inner sep=0.75pt]  [rotate=-312.61,xslant=0] [align=left] {Comments};
% Text Node
\draw (183.82,269.58) node [anchor=north west][inner sep=0.75pt]  [rotate=-312.61,xslant=0] [align=left] {Presence};
% Text Node
\draw (208.82,276.58) node [anchor=north west][inner sep=0.75pt]  [rotate=-312.61,xslant=0] [align=left] {Frequency};
% Text Node
\draw (241.82,273.58) node [anchor=north west][inner sep=0.75pt]  [rotate=-312.61,xslant=0] [align=left] {Regularity};
% Text Node
\draw (292.82,250.58) node [anchor=north west][inner sep=0.75pt]  [rotate=-312.61,xslant=0] [align=left] {Bytes};

\end{tikzpicture}
}
\captionsetup{type=figure,labelsep=space,font=small,labelfont=bf}
\captionof{figure}{Visual aid in the form of a relationship matrix employed by the knowledge base designer. Each 
arrow represents a possible conflict that could exist intuitively when employing two features to infer some level of computational trust.}
\label{figure:matrix}
\end{figure}
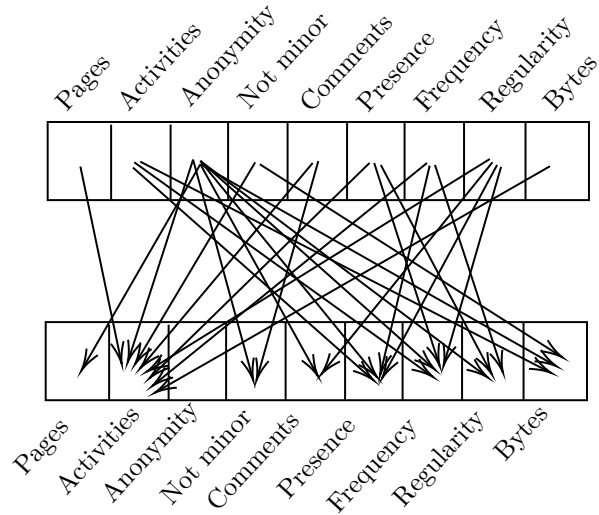

In the second knowledge base designed, the contradictions among IF-THEN rules were identified with the visual aid of a relationship matrix, Fig. \ref{figure:matrix}. It
 depicted the name of the features in two 
equal rows in order to help the author to identify pairs of features that could possibly have conflicts according to his/her knowledge.
Many more contradictions were identified in this case, since it was easier to visualise all the possible pairs of features that could have some 
conflicting set of beliefs. 
For example, the arrow from \texttt{pages} to \texttt{activity factor} was drawn as a reminder that while a high \texttt{activity factor} could be an indicator of high 
computational trust, this should not be the case if the editor has only modified a low number of pages. The idea is that a trustworthy editor would, based on the author's belief and intuition, collaborate on a high number of pages when performing a high number of edits.
In addition, IF-THEN rules that did not contradict each other, but which inferred different trust levels, were considered this time 
to contradict each other, resulting in a much larger number of contradictions. The rationale for this was to assume that only one trust level should be 
accepted, demanding more from the conflict resolution strategy of each non-monotonic reasoning approach and ideally performing fewer calculations for the 
aggregation of rules/arguments before a final inference.
Fig. \ref{figure:kb7} (p. \pageref{figure:kb7}) depicts the resulting graphical representation of this knowledge base.

Fuzzy membership functions (FMF) were also designed by the author to model natural language terms such as high, low, very low, and others (Fig. \ref{fig:fuzzyTrust}, p. \pageref{fig:fuzzyTrust}). These are necessary for the implementation of fuzzy reasoning models. Theoretically, such functions can provide higher precision for the modelling of natural language terms, or ‘fuzzy’ concepts. Natural language terms related to the same feature, for instance, low/medium low \texttt{frequency factor}, were designed in such a way that some intersection was possible between the terms. 
However, note that defining FMF is also a fuzzy process. Hence, two types of FMF were attempted, a linear and a gaussian one, which are often employed by fuzzy systems. The reason for having different types was to investigate their impact on the inferential capacity of fuzzy reasoning models. Other types could have been defined and further research can be made in this aspect.

In conclusion, note that these knowledge bases were built by a single agent, without the collaboration of other domain experts. 
Still, it is important to highlight that, despite such knowledge bases
not being subject to a formal process of validation, for instance by being inspected by other
experts, it is reasonable to assume that they are credible. This assumption comes from the
fact that the author has more than 10 years of heavy internet usage, competent
qualifications in computer science, and is experienced in a multitude of digital collaborative
environments. 
A different approach would be to collaborate with a larger group of experts,
producing greater and more sophisticated knowledge bases.
Nonetheless, this approach could also lead to other issues, for example due to an
expert’s difficulty in 
understanding the knowledge used by computational models, an expert’s capacity for verbalising knowledge, or an expert’s capacity for understanding the amount 
of detail required \citep{milton2007knowledge}. 
In summary, the process of knowledge acquisition is a familiar problem and a frequent bottleneck 
of knowledge-driven approaches in general. It is not the goal of this research article to propose a solution for such issues. Rather, this research creates 
trustworthy, credible knowledge bases. In turn, it employs the same knowledge bases to perform the envisioned comparison among 
non-monotonic reasoning approaches.

\subsection{Design of Non-Monotonic Reasoning Models Employing Expert Systems} \label{design:expert}

\begin{figure}[h]
\centering
\includegraphics[width=\textwidth]{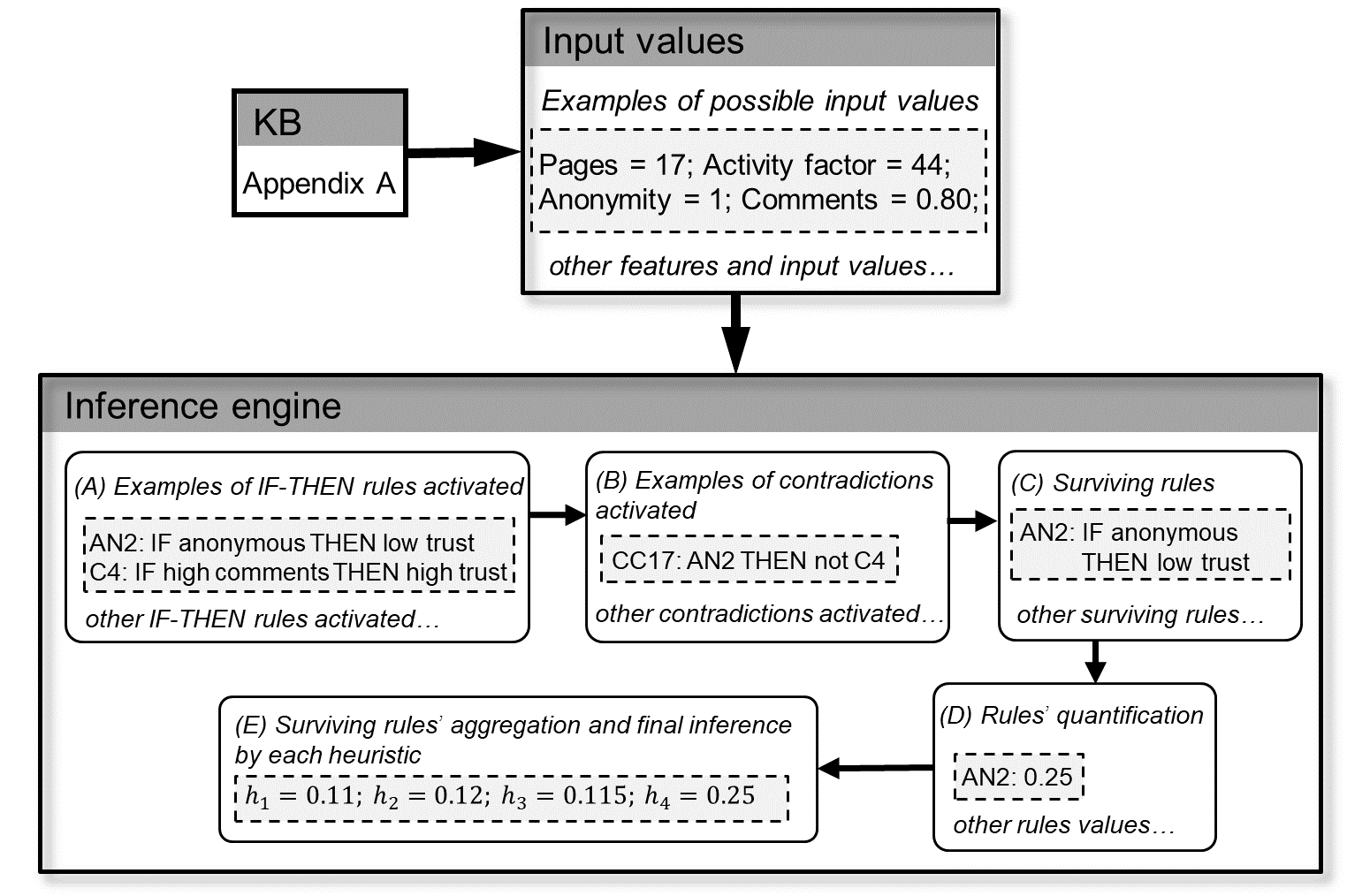}
\caption{An illustration of a reasoning process modelled by an expert system for a single Wikipedia editor.}
\label{figure:expert_system}
\end{figure}

This section provides a step-by-step description of the inferential process of possible expert system models when applied for modelling a non-monotonic 
reasoning process in the domain of computational trust. A running example is 
depicted in Fig. \ref{figure:expert_system}.
This example is referred to throughout this section and is aimed at providing a complete overview of the designed expert system inferential process.

\subsubsection{IF-THEN Rules}\label{sec:ifthenrules}

The first step of a rule-based expert system is to model a knowledge base usually gathered from an expert with rules in 
the form
\textit{``IF (antecedent) THEN (consequent)''}. 
In this research article, the antecedent is a set of premises associated with several quantitative features that are believed by the expert to influence the 
consequent being inferred (computational trust). The consequent might have different levels and is assumed to be derived from the premises by a domain expert. Therefore, no single deductive system is applied. The same premise/s might be used by different domain experts, but leading to different conclusions.
Each level of a premise in the antecedent, as well as each level of the consequent, is mapped to a numerical range also by the domain expert.
In this way, features associated to a certain level, such as low and high, can be evaluated true also according to continuous values. 
To formalise the generic 
case, IF-THEN rules are precisely defined.
This definition follows the logic structure in which 
antecedents can have multiple premises joined with AND/OR boolean operators, while the consequent is a single statement 
[\citealp{durkin1998expert}, chap. 3].

\begin{definition}[Generic IF-THEN rule]\label{def:genericrule}
A generic IF-THEN rule is defined, without loss of generalisability for OR and AND operators, as:

\begin{center} \textbf{IF} (\texttt{$i_1$} $\in [l_{1}, u_{1}]$ \textsc{AND} \texttt{$i_2$} $\in [l_{2}, 
u_{2}]$ )
\textsc{OR} (\texttt{$i_{3}$} $\in [l_{3}, u_{3}]$ \textsc{AND} \texttt{$i_{4}$} $\in 
[l_{4}, u_{4}]$) \\
 \textbf{THEN} \textit{consequent} $\in [l_{c}, u_{c}]$ 
\end{center}

Where \textbf{\texttt{$i_n \in \mathbb{R}$}} is the input value of the feature $n$ with numerical range $[l_{n}  \in \mathbb{R}$, $u_{n}  \in \mathbb{R}]$;
the range [$l_{c}  \in \mathbb{R}$, $u_{c}  \in \mathbb{R}$] is the numerical range for the consequent level being inferred with; and 
\textsc{AND} and 
\textsc{OR} are boolean logical operators.
\end{definition}

\subsubsection{Inference Engine and Non-monotonic Extension}

The inference engine starts with the activation of IF-THEN rules by input data. This data is used to evaluate antecedents of rules, activating a subset whose evaluation returns true. This evaluation is based on the numerical ranges provided by the knowledge base designer. For 
instance, in rule C4, \textit{high} \texttt{comments} means that a \texttt{comments} 
input value between [0.75, 1] $\subset \mathbb{R}$ provided by a user will activate the rule.
A rule can also be 
contradicted by other rules that intend to bring forward and support contradictory information.
Formally, these can also be seen as \textit{meta-rules}
[\citealp{durkin1998expert}, chap. 3], or rules that describe how other rules should be used, as in the contradictions in the running example. These are activated in the same fashion as IF-THEN rules.
The main difference comes from the fact that the consequent of 
these meta-rules, or contradictions, might impact other meta-rules or other IF-THEN rules, while the consequent of IF-THEN rules is being employed only for the inference of computational trust.
If both an IF-THEN rule and at least one contradiction challenging the 
rule have been activated, then the inference engine discards the rule. This mechanism will eventually form a set of surviving rules.
The running example depicts some input values, two activated rules and one surviving rule.

In this research article, the rules in the set of surviving rules will always be inferring the same consequent, but most likely at different levels. 
Since the goal is to aggregate them and to extract a unique scalar, with the most representative of the 
consequents being inferred, an aggregation strategy is needed. In this situation, a usual expert system would have a typical set of choices for
selection of rules \citep{mcdermott1978production,durkin1998expert}, for example, deciding a priority for each rule, returning multiple outcomes or choosing the first rule activated. However, none of these 
strategies are applicable in this research study. The constructed knowledge bases do not give explicit preferences among 
rules, order of activation or possibility to compute more than one output. Because of this, IF-THEN rules must be quantified and aggregated to infer a single scalar.
In the present study, this value is defined according to the 
numerical  range of the consequent of the rule, the numerical range of its premises and the input values provided for the rule activation. In the basic scenario of an 
IF-THEN rule with only one premise, it is quantified as the minimum (respective maximum) value of the numerical range of its consequent if its premise is 
activated with its minimum (respective maximum) value. For instance, consider rule C4 rewritten with illustrative numerical ranges:

\begin{itemize}
 \item [-] \textcolor{black}{C4 rewritten}: \textbf{IF} \texttt{comments} $\subset$ [0.75, 1] \textbf{THEN} trust $\subset$ [0.75, 1]
\end{itemize}

In this case, if the input value for \texttt{comments} is 0.75, then C4 value will be 0.75. Analogously, if the input value for \texttt{comments} is 1, 
then C4 value will be 1. Activation values greater than 0.75 
and less than 1 are evaluated according to a linear relationship. This is defined by a function $f$ as proposed in \cite{rizzo2020thesis}:

\begin{definition}[Generic rule value]\label{def:rulevalue}
The value of a generic IF-THEN rule $r$ is given by the function:

\begin{center}
    $f(r) = \frac{|u_{c} - l_{c}|}{R_{max} - R_{min}} \times (v - R_{max}) + u_{c}$, where 
 \\ \vspace{3mm}
 $v = min[max(i_1, i_2), max(i_{3}, i_{4})]$,
 \\
 $R_{max} = min[max(u_{1}, u_{2}), max(u_{3}, u_{4})]$,
 \\
 $R_{min} = min[max(l_{1}, l_{2}), max(l_{3}, l_{4})]$
 \end{center}

\end{definition}

The value of a rule will always lie between the numerical range $[l_{c}$, $u_{c}]$ of its consequent. Moreover, the boundaries in this range will define the type of relationship between premises and consequent:
\begin{itemize}
 \item If $l_{c} < u_{c}$, then Definition \ref{def:rulevalue} will model a linear relationship. The higher the value of the premise/s, the higher the value of the conclusion.
 \item If $l_{c} > u_{c}$, then Definition \ref{def:rulevalue} will model a contrary linear relationship. The higher the value of the premise/s, the lower the value of the conclusion.
 \item If $l_{c} = u_{c}$, then Definition \ref{def:rulevalue} will model a constant function whose every input results in the same output ($u_{c}$). This 
might be useful to model consequents with categorical levels.
\end{itemize}

Briefly, Definition \ref{def:rulevalue} 
provides an evaluation formula for rules that employ logical operators \textsc{AND}/\textsc{OR}, replacing them for \textit{max} and 
\textit{min} 
operators. Different 
operators could have been employed if required by the domain of application or human 
reasoner.  Moreover, if there is no reason to select one operator over another, they could also be a parameter when designing expert system models. However, 
having in mind the rules contained in the knowledge bases employed in this study, the adoption of other 
operators would likely not have a significant impact on the results. The antecedents of these rules are often formed by a single premise, thus, their 
evaluation would follow 
a simple linear relationship regardless of the aggregation strategy adopted for multiple premises.

Finally, four heuristics defined in previous works \cite{Rizzo2016} are employed for the aggregation of values assigned to surviving IF-THEN rules. The strategies are defined, 
in order to extract 
different points of view from remaining rules and to accommodate the use of rule weights when weights are provided.
Weights among rules are provided in the employed knowledge bases. They are the result of a
pairwise comparison procedure between the 9 employed features (Table \ref{table:featurestransf}, p. \pageref{table:featurestransf}) performed by the knowledge base designer. Hence, they will be numbers in the range [0, 8] $\subset \mathbb{N}$, being 0 if a feature is considered less important than any other feature for the inference of computational trust, and 8 if it is considered more important than any other feature.
The weight of a feature will also represent the weight of the rule employing this feature.
The aim is to investigate the impact of adding this 
extra information on the inferential capacity of the expert systems models. Thus, the heuristics are:

\begin{itemize}\label{list:heuristics}
\item[-] \boldsymbol{$h_1$}: definition of the sets of surviving rules grouped by their consequent level. Extraction of the largest set. Average of the values 
of the rules in the largest set. In case two or more of the largest sets exist, the above process is repeated for each, and their average is returned. The idea 
is to give importance to the largest set of surviving rules supporting the same consequent level.
\item[-] \boldsymbol{$h_2$}: same as $h_1$ but applying the weighted average instead of the average. The goal here is to
allow the possibility of defining weights to specific rules.
\item[-] \boldsymbol{$h_3$}: average value of all surviving IF-THEN rules. This is to give equal importance to all surviving IF-THEN rules, regardless 
of which level of the consequent they were supporting.
\item[-] \boldsymbol{$h_4$}: same as $h_3$ but applying the weighted average instead of the average. Again, the goal is to allow the use of weights attributed 
to specific rules.
\end{itemize}
The running example depicts an illustrative output for the four heuristics in the last step of the inference engine.

\subsection{Design of Non-monotonic Reasoning Models Employing Fuzzy Reasoning} \label{design:fuzzy}
\begin{figure}[h]
\centering
\includegraphics[width=\textwidth]{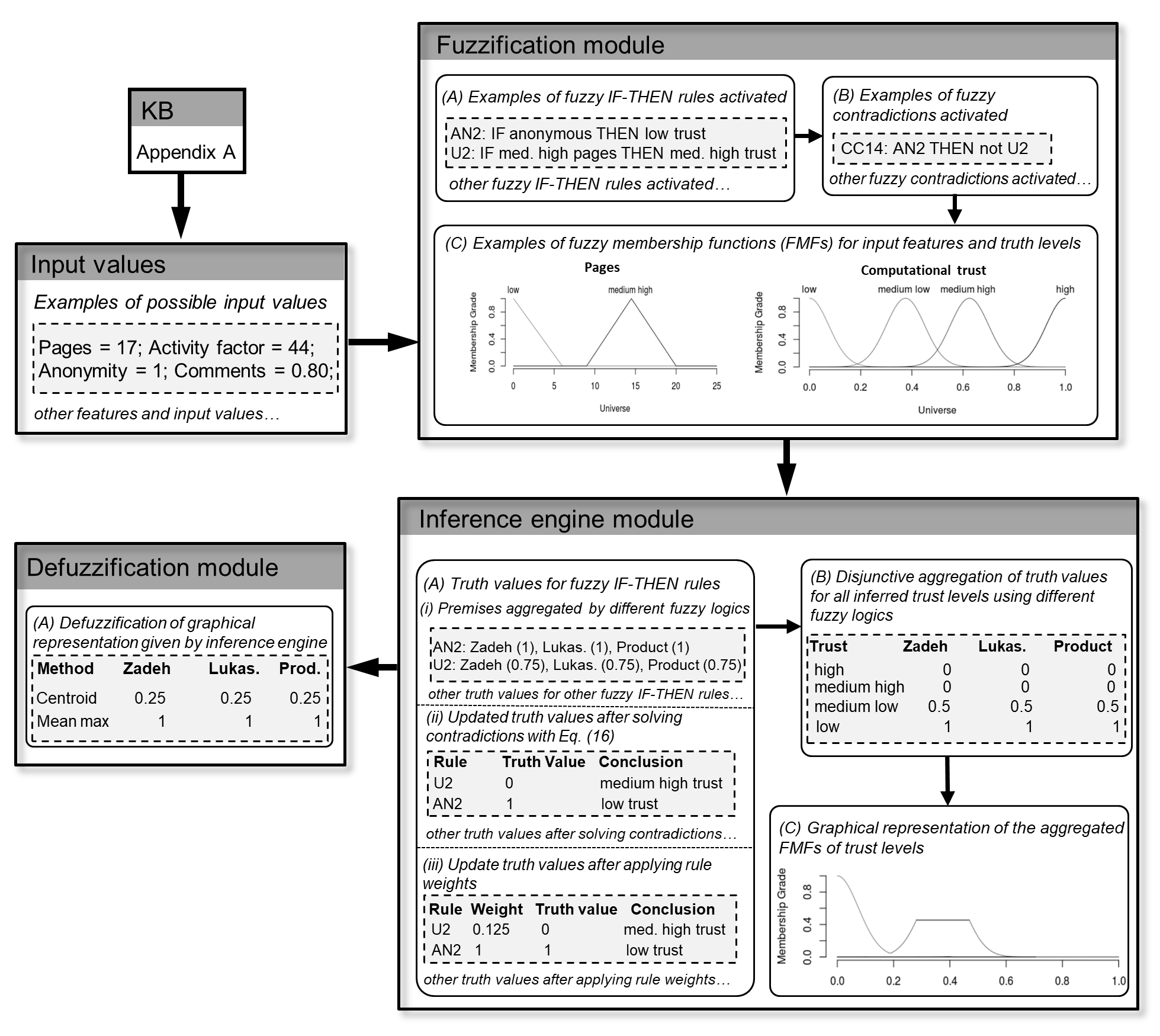}
\caption{An illustration of a reasoning process modelled by fuzzy reasoning for a single Wikipedia editor.}
\label{fig:fuzzy_reasoning}
\end{figure}

Fuzzy reasoning provides a robust representation of linguistic 
information by using fuzzy membership functions. In this research article, the structure of a Mamdani fuzzy control system and the use of possibility theory as defined in [\citealp{siler2005fuzzy}, chap. 8], and reviewed in Section \ref{reivewfuzzy}, are employed for the definition of fuzzy 
reasoning models of inference. As with expert systems, a running example of a single inference is depicted in Fig. \ref{fig:fuzzy_reasoning} and referred throughout this subsection.

\subsubsection{Fuzzification Module}

The first step, the fuzzification module, starts with the definition of \textit{fuzzy} IF-THEN 
rules and \textit{fuzzy} contradictions. Their structure is the same as that presented for expert systems, but they are computed in a different fashion. 
Each linguistic term associated with a feature level or consequent level, such as \textit{high} or \textit{low}, is then described by 
a FMF, which is also provided by the knowledge base designer. In the running example, some FMF employed are depicted in the fuzzification module.

\subsubsection{Inference Engine and a Non-monotonic Extension}

Once the fuzzification step has been completed and the knowledge base of the expert translated into fuzzy IF-THEN rules and fuzzy contradictions, the next step 
is to evaluate the initial truth values of the fuzzy IF-THEN rules. To do so, each membership grade on the antecedent of these rules needs to be evaluated 
according to the input data. For instance, consider rule U2 in the running example. If \texttt{pages} = 17, then the membership grade of the linguist term \textit{medium high} is 
0.75, according to the FMF for \texttt{pages}. In this case, the initial truth value of U2, before solving contradictions, is also 0.75. U2 is a simplified example, but more than one feature can be contained in each rule’s antecedents.
Hence, a t-conorm and a t-norm are necessary to implement the notions
of union and 
intersection. 
In the literature of fuzzy logic, 
t-norms (fuzzy intersection) and t-conorms (fuzzy union) \citep{klement2002triangular}, are a function $F: [0, 1]^2 \rightarrow [0, 1]$ such that the axioms 
of commutativity, associativity, monotonicity, and boundary 
condition are satisfied.
Most commonly, the Zadeh, the Product and the 
\L{}ukasiewicz operators are employed and are those selected for investigation in this research study, as listed in Table \ref{table:fuzzyoperators}. Other operators can be seen in \citep{klement2002triangular}.

\renewcommand\theadalign{c}
\begin{table}[!h]
%\scriptsize
\footnotesize
   \caption{T-norms and t-conorms usually employed by fuzzy systems. $x, y \in [0, 1]$.}
\label{table:fuzzyoperators}   
\begin{tabular}{P{2.5cm}P{4.2cm}P{4.2cm}} 
\hline
\multirow{2}{*}{\textbf{Fuzzy operator}} & \multirow{2}{*}{\textbf{T-Norm}} & \multirow{2}{*}{\textbf{T-Conorm}}\\
\\
Zadeh & $T_m(x, y)$ = min($x$,$y$) & $C_m(x, y)$ = max($x$,$y$) \\
\L{}ukasiewicz & $T_{L}(x, y)$ =  max($x$ + $y$ - 1, 0)  & $C_L(x, y)$ = min($x$ + $y$, 1) \\
Product & $T_p(x, y)$ = $x \cdot y$  & $C_p(x, y)$ = $x$ + $y$ - $x \cdot y$ \\
% Drastic & $T^*(x,y) = \bigg\{$\hspace{-6mm}\thead{$y$ if $x = 1$ \\ $x$ if $y = 1$ \\ 0 otherwise} & $C^*(x, y) = \bigg\{$\hspace{-6mm} \thead{$y$ if $x = 0$ 
% \\ $x$ if $y = 0$ \\ 1 otherwise} \\
\hline
\end{tabular}
\end{table}

Following the calculation of the fuzzy IF-THEN rules’ initial truth values, possibility theory is adopted for the resolution of contradictions. According to the 
approach proposed in \citep{siler2005fuzzy}, truth values are used to represent \textit{possibility} (Pos) and 
\textit{necessity} (Nec) as defined in Section 
\ref{sec:nmfuzzy}.
In this study, necessity is represented by the membership grade of a proposition and possibility 
is always 1 for all propositions.
This means that all propositions (or rules) are open to be retracted. Since there is no addition of supporting information, but only attempts 
to contradict or refute information, it is possible to employ Equation 
(\ref{eq:possibility}) (p. \pageref{eq:possibility}) to deal with the contradictions in the knowledge bases of this study.
For instance, the new necessity of fuzzy rule U2, if it is contradicted only by the fuzzy contradiction CC14, is given by:
\begin{itemize}
\vspace{-2mm}
\item [-] Nec(U2) = min (Nec(\textit{medium high} \texttt{pages}), 1 - Nec(\texttt{anonymous})) 
%\vspace{-6mm}
\end{itemize}

Nec(\textit{medium high} \texttt{pages}) is the membership grade of the linguistic variable \textit{medium high} \texttt{pages}.
Three situations might arise in this case:
\begin{itemize}
 \item If Nec(\texttt{anonymous}) = 0, then CC14 has no impact on the necessity of U2.
 \item If Nec(\texttt{anonymous}) = 1, then U2 is refuted and assumes new necessity 0.
 \item If $0 <$ Nec(\texttt{anonymous}) $< 1$, then U2 can only maintain the same necessity or have it decreased to a value 
greater than 0, indicating a partial refutation.
\end{itemize}

The new necessity of the fuzzy rule U2 represents the truth value of \textit{medium high} \texttt{trust} in this rule. However, it is important to highlight that the approach 
developed in \citep{siler2005fuzzy} has been inspired by a multi-step forward-chaining reasoning system. In this research study, reasoning is done in a single 
step, in the sense that data is imported, and all rules are fired at once. Nonetheless, it is possible to define a precedence order of fuzzy contradictions. 
More precisely, it is possible to define a tree structure in which the consequent of a fuzzy contradiction is the antecedent of the next fuzzy contradiction. In 
this way, Equation (\ref{eq:possibility}) can be applied from the root or roots to the leaves. 
In case of cyclic contradictions, or contradictions whose consequents impact each other's premises, they are solved simultaneously.
For that, the 
truth value of all fuzzy rules is stored before solving any cyclic fuzzy contradictions. In turn, the final truth value of fuzzy rules is
calculated according to these stored values.

Having the fuzzy contradictions solved by the proposed mechanism, rule weights (if defined) 
can be applied to the current truth values of fuzzy IF-THEN rules. In this study,  the approach proposed in \citep{ishibuchi2001effect} is selected. In this case, rule weights are normalised in the range $[0, 1] \subset \mathbb{R}$
and multiplied by the current truth value of each rule.
In this example, it is also assumed that the weight of a feature represents the weight of the rule that contains this feature, as implemented by expert system models.

Eventually, a disjunctive approach is employed for computing the truth values of the consequent levels. Hence, each consequent level is given by the maximum 
value of the truth values of the fuzzy IF-THEN rules that infer the same consequent level.
If a conjunctive approach was selected (using the minimum value instead), the set of rules would be jointly satisfied, representing a stricter proposal. Here,
the disjunctive approach is selected for being a more flexible proposal that guarantees that at least one rule is satisfied.
The membership grade of updated fuzzy IF-THEN rules will then propagate to their consequents, producing a set of truncated membership functions associated with their 
consequents.
The inference engine in the running example depicts the truth values of fuzzy rules, their updated values after solving contractions, and, finally, after applying rule weights. This is followed by the disjunctive aggregation of trust levels and the definition of the respective graphical representation.

\subsubsection{Defuzzification Module}

The output of the inference engine is a graphic representation of the aggregation of these truncated membership functions.
Several methods can be used for calculating a single defuzzified scalar from this graphic representation \citep{hellendoorn1993defuzzification}.
Two are commonly employed and selected here: \textit{mean of max} and \textit{centroid}. The first returns the average of all elements (consequent levels) with maximal membership grade. The second returns the coordinates ($x$, $y$) of the centre of gravity of the graphic representation. The defuzzified scalar is represented then by the $x$ coordinate of the centroid, as per the defuzzification module in the running example.

\subsection{Design of Non-Monotonic Reasoning Models Employing Defeasible Argumentation} \label{design:argumentation}

\begin{figure}[h]
\centering
\includegraphics[width=0.925\textwidth]{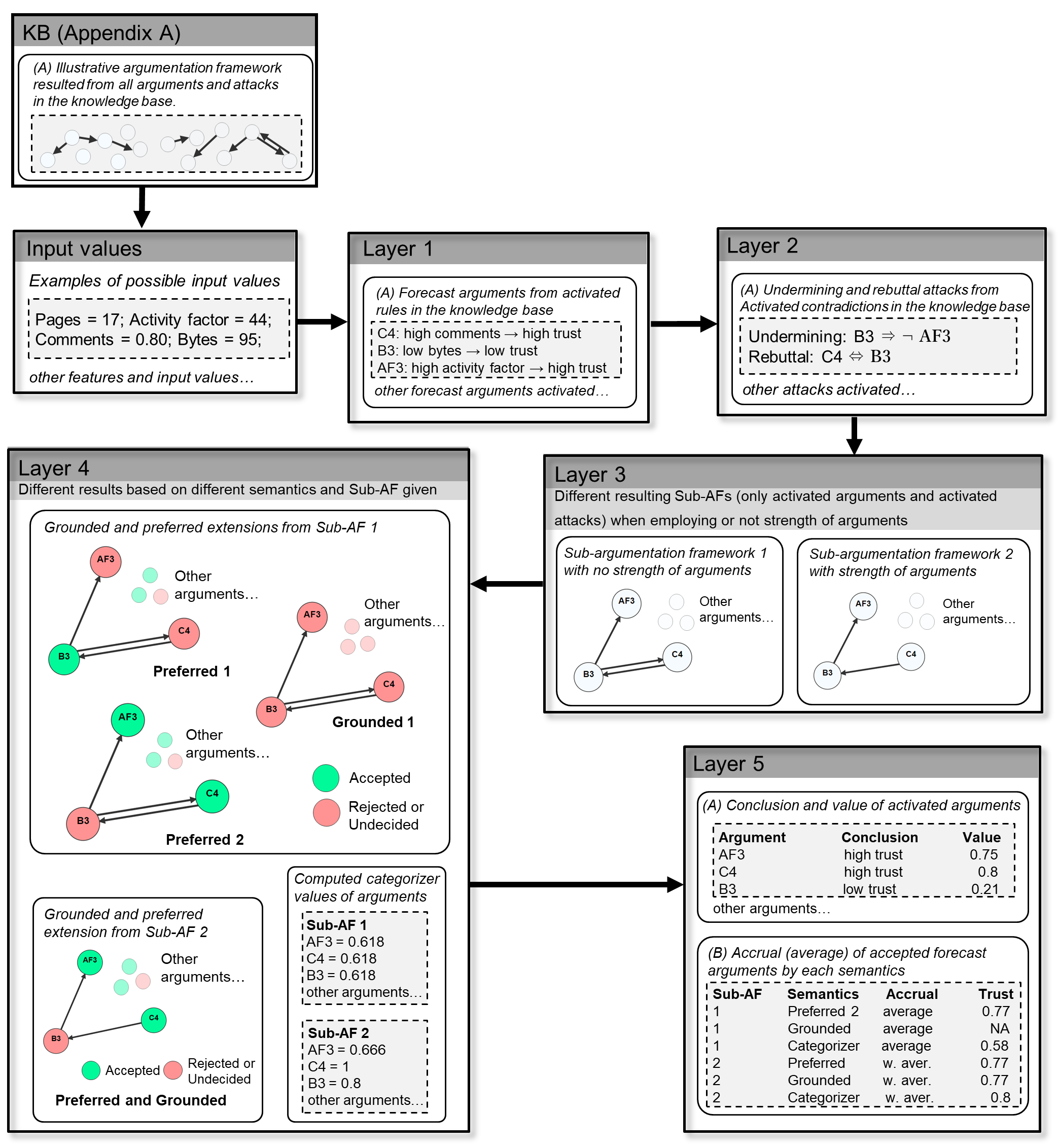}
\caption{An illustration of a reasoning process modelled by defeasible argumentation for a single Wikipedia editor.}
\label{fig:argumentation_reasoning}
\end{figure}

The definition of argument based-models follows the five-layer modelling approach proposed in \citep{longo2016argumentation} and depicted in Fig. 
\ref{fig:multilayerstructure} (p. \pageref{fig:multilayerstructure}). It starts with the definition of the internal structure of arguments, continues with 
the 
definition of their conflicts, the computation of their acceptance status, and ends with the aggregation of accepted arguments. A running example is depicted 
in Fig. \ref{fig:argumentation_reasoning} and referred throughout this subsection.

% XXX Grounded needs to have only rejected arguments or some inference with a number. rejected needs to be rejected/undecided. Layer 1 needs (A)

\subsubsection{Layer 1: Definition of the Internal Structure of Arguments}

Most commonly, an argument is composed of one or more premises and a conclusion derivable by applying an inference rule. Hence, the first step of an 
argumentation process is to define its \textit{forecast arguments}:
\begin{center}

$Forecast \; argument: premises \rightarrow conclusion$ 
\end{center}

This structure includes a set of premises (believed to influence the conclusion being inferred) and a conclusion derivable by applying an inference rule 
$\rightarrow$.
It is an uncertain implication, which is used to represent a defeasible argument. As with the rules of expert systems, premises and conclusions are strictly 
bounded in numerical ranges associated with natural language terms (for instance \textit{low} and 
\textit{high}). Forecast arguments are activated if their premises evaluate as true, according to the input data provided. Boolean logical operators 
AND and OR can be applied for the use of multiple premises, similar to the rules employed by expert system models.

\subsubsection{Layer 2: Definition of the Interactions between Arguments}

In order to evaluate inconsistencies, the notion of \textit{mitigating argument} \citep{Toni2010} is introduced.
These are arguments that attack other forecast arguments or other mitigating arguments.
Both forecast and mitigating arguments are special \textit{defeasible rules}, as defined in \citep{prakken2010abstract}. 
Informally, if their premises hold then \textit{presumably} (defeasibly) their conclusions also hold.
Different types of attacks and consequently mitigating arguments exist in the literature, as reviewed in Section \ref{AT} (p. \pageref{AT}).
For instance, in this research article, an undermining attack is represented by a forecast argument and an 
inference $\Rightarrow$ to a negated argument B (forecast or mitigating):

\begin{center}
$Undermining \; attack: forecast \; argument \Rightarrow \neg B$
\end{center}

Note that these are undermining attacks because the conflict arises from the premises in each forecast argument.
For instance, in the running example, B3 is attacking AF3 by 
stating that \texttt{activity factor} should not be \textit{high}, since \texttt{bytes} is 
\textit{low}.
An undercutting attack would be defined if for some reason the inference being 
performed by an argument was being
contested. For example, contradictions ``OnlyAge'' described in Table \ref{table:kb1contradictions}. Here, an undercutting attack is represented by a set of premises and an inference to a negated argument B (forecast or mitigating):

\begin{center}
$Undercutting \; attack: premises \Rightarrow \neg B$
\end{center}

Lastly, a rebuttal attack 
would be created if, for some reason, it was believed that the conclusion of an argument was false. For instance, some domain expert could define an attack 
targeted at C4 by saying that there 
is evidence to infer another level of trust instead of \textit{high}. 
Note that in the context of computational trust, different 
consequents (trust levels) might coexist or not according to the expert's reasoning, hence not all arguments with different conclusions should necessarily lead to rebuttal attacks. Moreover, since all arguments in the constructed knowledge bases are defeasible (or not 
strict) rebuttals would be mutual (both arguments would attack each other). Rebuttal attacks ($\Leftrightarrow$) occur in this research article when two forecast arguments support 
mutually exclusive conclusions according to a domain expert, hence, are represented as:

\begin{center}
$Rebuttal \; attack: forecast \; argument \Leftrightarrow forecast \; argument$
\end{center}

Let us point out that different types of attacks can enhance the explainability of argument-based models and aid in the process of creating knowledge bases. 
However, they do not 
impact on the computation of the acceptability status of arguments and final numerical scalar being produced by such models in the next layers. This 
computation is performed via abstract argumentation theory as proposed in \cite{Dung1995}. In this case, all attacks are seeing as a binary relation as 
described in Section \ref{AT}.

\subsubsection{Layer 3: Evaluation of the Conflicts of Arguments}\label{sec:designlayer3}

At this stage, forecast and mitigating arguments can be seen as an argumentation framework, which can be elicited with data.  
Arguments will then be activated or discarded, based on whether their premises 
evaluate as true or false. Attacks between activated arguments will be evaluated before being activated
as well. These usually have a form of a binary relation. In a binary relation, a successful (activated) attack occurs 
whenever both of its source (argument attacking) and its target (argument being attacked) are activated.
This study also makes use of the notion of strength of arguments as presented in \citep{pollock1995cognitive}.
In this case, an attack is considered successful only if the strength of its source is equal to, or greater than, the strength of its target. In the running 
example, feature weights are employed for defining the strength of 
arguments. As with the definition of rule 
weights in expert systems and fuzzy reasoning, the weight of a feature will also represent the strength of the argument employing this feature.
The running example depicts two sub-AFs from activated arguments and attacks, with and without strength or arguments. These sub-AFs 
are argumentation frameworks contained in the original ones created in Layer 2, but only considering activated arguments and attacks.

\subsubsection{Layer 4: Definition of the Acceptance Status of Arguments}

Given a sub-AF, all its arguments are considered abstract as proposed in \cite{Dung1995}. In turn, acceptability semantics \cite{Dung1995,wu2010labelling,caminada2009logical,bonzon2016comparative,besnard2001logic}  are applied to compute the acceptance status of each argument, or its acceptability.
As previously defined, acceptability semantics evaluate the overall interaction of arguments across the AF (or sub-AF in this argumentation process), in order 
to select the arguments that should ultimately be accepted. In the running example, two well-known extension-based semantics, preferred and grounded \cite{Dung1995}, and one ranking-based semantics, categoriser \cite{besnard2001logic}, are illustrated. For the sake of simplicity, their formal definitions are presented in \ref{app:semantics}.

\subsubsection{Layer 5: Accrual of Acceptable Arguments}

Eventually, in the last step of the reasoning process, a final inference must be produced. In the case of extension-base semantics if multiple extensions are 
computed, one might be preferred over the others. In this study, the cardinality of an extension (number of accepted arguments) is used as a mechanism for the 
quantification of its credibility. Intuitively, a larger extension of arguments, that by definition are also conflict-free, might be seen as more credible than 
smaller
extensions. If the computed extensions have all the same cardinality, these are all brought forward in the reasoning process. After the selection of the larger 
extension/s or best-ranked argument/s, a single scalar is produced through the accrual of the values assigned to its/their forecast arguments (arguments that 
infer some trust level). Mitigating arguments have already completed their role by contributing to the resolution of conflicting 
information (previous layer) and thus, are not considered in this layer. The values of forecast arguments follow from the same formula described in Definition 
\ref{def:rulevalue}. An assumption is made here that forecast arguments have a similar structure to the IF-THEN rules defined for 
expert systems. Premises are associated with numerical ranges and concatenated by boolean operators AND and OR, and the conclusion has a numerical range as the 
consequents of the IF-THEN rules. Having their values assigned, the accrual of forecast arguments can be made in
different ways, for instance considering measures of central tendency. Here, the average is accounted for models that use a binary relation of attacks, while the 
weighted average is accounted for models that use the notion of strengths of arguments.
Note that in the case of two or more preferred extensions with the same number of accepted forecast arguments, the outcome of the preferred semantics is the 
mean of all its extensions.

%\subsection{Participants and procedures}\label{experimentsdesign}

\subsection{Summary of Models and Comparative Metrics}\label{metricsmodels}

The list of models built with different reasoning approaches can be seen in \ref{listmodels}. Overall, 68 models were implemented, with different configuration parameters selected for evaluation, as described in the previous sections. To compare the inferences produced by them, three evaluation metrics for computational trust are employed: \textit{rank of Barnstars}, \textit{spread}, and \textit{percentage of NAs} (not assigned). Table \ref{table:wikimetricsCalculation} lists the calculation method of each metric.

\begin{table}[!h]
\footnotesize
 \setlength{\tabcolsep}{6pt}
     \renewcommand*{\arraystretch}{1.5}
   \caption{Calculation of evaluation metrics employed to assess the trust inferences performed by non-monotonic reasoning models in the Wikipedia Project.}
\label{table:wikimetricsCalculation}   
\centering
\begin{tabular}{lp{8.6cm}}
\hline
\textbf{Metric} & \textbf{Calculation}
\\
\arrayrulecolor{black}\hline
\multirow{5}{*}{Rank of Barnstars} & Remove editors with no trust value assigned from the dataset. Sort all other editors by 
their trust values in descending order. Non-Barnstars tied with Barnstars are ranked above. Sum the ranks of the Barnstar editors and normalise the result in 
the range range [0, 100] $\subset\mathbb{R}$. 0 means all Barnstars with an assigned trust value are ranked above any non-Barnstar, while 100 means they are ranked 
below any non-Barnstar.\\
\arrayrulecolor{gray!40}\hline
Spread & Standard deviation of the 
trust values assigned to Barnstars. \\
\arrayrulecolor{gray!40}\hline
%\multirow{2}{*}{Medoid difference} & Absolute 
%difference between the medoid of the trust values assigned to barnstars and the medoid of the trust values assigned to non-barnstars. \\
Percentage of NAs & Percentage of editors that had no assigned trust value.\\
\arrayrulecolor{black}\hline
\end{tabular}
\end{table}

The rationale behind the Rank of Barnstars metric, is that when sorting editors in descending order by their assigned trust values, 
it is assumed that the ranking of the best models will result in 
Barnstar editors being placed at the highest positions. Non-Barnstar editors may also be highly trustworthy. Nonetheless,
Barnstar editors still should, presumably, be ranked at the highest positions.
Moreover, since trust is not a binary concept, it is expected that the distribution of the trust values assigned by these same models to Barnstar editors should have a positive, continuous spread. Spread is measured by the standard deviation of the values assigned to Barnstar editors. Finally, models that are capable of producing a final inference in more cases are considered better, thus a higher percentage of NAs, or cases without an assigned inference, is deemed as a disadvantage.
Note that no metric was defined for computing an overall difference 
between trust values assigned to Barnstar editors and trust values assigned to non-Barnstar editors. The reason for this derives from the lack of knowledge 
about the non-Barnstar editors.

% Results section

\section{Results and Discussion}

This section presents the results of the instantiation of the designed non-monotonic reasoning models using two sets of data, each built from 
a Wikipedia XML dump, first from the Italian-language and then from the Portuguese-language edition. Defeasible argumentation and expert system models were implemented through an online framework using entirely custom code in PHP and JavaScript \cite{RizzoFramework}.
%\footnote{\url{http://www.lucasrizzo.com/framework/index.php/}}.
Differently, fuzzy reasoning models were implemented using the C++ programming language\footnote{\url{https://doi.org/10.5281/zenodo.5088015}}.
The presentation of results is structured 
around each evaluation metric, a summary, and a final discussion.

\subsection{Rank of Barnstars}

This metric is targeted at evaluating whether the investigate non-monotonic reasoning models are capable of ranking the Barnstar editors at the highest 
positions. Fig. \ref{figure:rank_KB1} and \ref{figure:rank_kb2} depicts the normalised sum of Barnstar ranks in the range $[0, 100] \subset\mathbb{R}$ for the models built with knowledge base 1 (KB1) and knowledge base 2 (KB2) respectively.
Each figure depicts first results when models are instantiated by the Italian-language dataset and then by the Portuguese-language dataset.
Instances without an assigned 
trust scalar were removed from this analysis. A baseline was computed by the average of the normalised features reported by each editor, also resulting in inferences in the range $[0, 1] \subset\mathbb{R}$. This baseline was defined only to indicate whether the non-monotonic reasoning process performed by the implemented models was effective in comparison to a non-deductive and simplified inference.

\pgfdeclarelayer{bg}    % declare background layer
\pgfsetlayers{bg,main}

\begin{figure} [!h]
\centering

\begin{tikzpicture}
\pgfplotsset{width=1.3*\textwidth}
\begin{pgfonlayer}{bg}
    \begin{axis}[
        compat=newest, %Better label placement
        width  = 1\textwidth,
        height = 4cm,
        major x tick style = transparent,
        ybar=4*\pgflinewidth,
        bar width=2mm,
        bar shift=0pt,
        ymajorgrids = true,
        xlabel = {Reasoning models for computational trust inference},
        ylabel = {Rank of Barnstars},
        label style={font=\fontsize{11}{10.5}\selectfont},
        tick label style={font=\fontsize{11}{10.5}\selectfont},
        nodes near coords,
        nodes near coords align={vertical},
        every node near coord/.append style={font=\fontsize{8}{7.5}\selectfont},
        every node near coord/.append style={rotate=90, anchor=west},
        %symbolic x coords={FL1 FC1, FL2 FC2, FL3 FC3, FL4 FC4, FL5 FC5, FL6 FC6, FL7 FC7, FL8 FC8, FL9 FC9, FL10 FC10, FL11 FC11, FL12 FC12, A1, A2, A3, A4, 
%E1, E2, E3, E4},
        %xtick={ },
        %x tick label style={font=\scriptsize,  align=left,text width=3.5cm},
        xtick={1,2,3,4,5,6,7,8,9,10,11,12,13,14,15,16,17,18,19,20,21,22,23,24,25,26,27,28,29,30,31,32,33,34},
        xticklabels={
        $\star$ $\circ$ FL3,
        $\triangleright$ $\circ$ FL9,
        $\star$ $\circ$ FL5,
        $\triangleright$ $\circ$ FL11,
        $\star$ $\circ$ FL1,
        $\triangleright$ $\circ$ FL7,
        $\star$ $\ddagger$ A2,
        $\star$ $(h_3)$ E3,
        $\star$ $\otimes$ A1,
        $\star$ $\odot$ A3,
        $\star$ $\circ$ FC3,
        $\triangleright$ $\circ$ FC9,
        $\star$ $\circ$ FC5,
        $\triangleright$ $\circ$ FC11,
        $\triangleright$ $(h_4)$ E4,
        $\triangleright$ $\otimes$ A4,
        $\triangleright$ $\odot$ A6,
        $\triangleright$ $\ddagger$ A5,
        $\star$ $\circ$ FC1,
        $\triangleright$ $\circ$ FC7,
        $\star$ $(h_1)$ E1,
        $\triangleright$ $(h_2)$ E2,
        $\star$ $\bullet$ FL6,
        $\triangleright$ $\bullet$ FL12,
        $\star$ $\bullet$ FC6,
        $\triangleright$ $\bullet$ FC12,
        $\star$ $\bullet$ FL2,
        $\star$ $\bullet$ FL4,
        $\triangleright$ $\bullet$ FL8,
        $\triangleright$ $\bullet$ FL10,
        $\star$ $\bullet$ FC2,
        $\star$ $\bullet$ FC4,
        $\triangleright$ $\bullet$ FC8,
        $\triangleright$ $\bullet$ FC10},
        x tick label style={font=\fontsize{7.5}{8.0}\selectfont, align=left, rotate=90, text width=1.20cm},
    %hide obscured x ticks=false,
        %tick align=inside,
        %xtick = data,      
        %tickwidth=4pt,
        %scaled y ticks = false,
        enlarge x limits=0.05,
        ymin=0,
        ymax=15,
        clip=false,
        %nodes near coords,
        %every node near coord/.append style={color=black, font=\tiny},
        %legend cell align=left,
        %legend style={
        %        at={(1,1.05)},
        %        anchor=north,
        %        column sep=1ex
        %},
        legend style={at={(0.5,1.30)},
        anchor=north,legend columns=-1,  
        font=\fontsize{7.5}{8.5}\selectfont},
    ]
    
     %\draw [thin, dotted, thick, draw=black] 
     %    (axis cs: -0.5, 366.152) -- (axis cs: 33.5, 366.152);
%         node[pos=0.995, right] {\fontsize{4}{4.5}\selectfont \textbf{NASA}};
    
   \draw [thin, dotted, draw=black] 
        (axis cs: -0.5, 7.54) -- (axis cs: 35.7, 7.54)
        node[pos=0.4, above] {\fontsize{7}{7.5}\selectfont Features' average (7.54)};
  
   \addplot[style={black,pattern=north east lines, pattern color=blue,mark=none, /pgf/number format/fixed,
        /pgf/number format/precision=2}] coordinates {
        (1,0.2873)
        (2,0.2873)
        (3,0.3054)
        (4,0.3054)
        (5,0.3247)
        (6,0.3247)
        (23,4.6955)
        (24,4.6955)
        (27,10.3991)
        (28,10.3991)
        (29,10.3991)
        (30,10.3991)
        };

    \addplot[style={black, pattern=crosshatch dots, pattern color=blue,mark=none, /pgf/number format/fixed,
        /pgf/number format/precision=2}] coordinates {
        (11,0.4198)
        (12,0.4198)
        (13,0.4324)
        (14,0.4324)
        (19,0.5094)
        (20,0.5094)
        (25,5.2341)
        (26,5.2341)
        (31,10.3991)
        (32,10.3991)
        (33,10.3991)
        (34,10.3991)
        };

    \addplot[style={black, pattern=grid, pattern color=green,mark=none, /pgf/number format/fixed,
        /pgf/number format/precision=2}] coordinates {
        (8,0.3746)
        (15,0.4967)
        (21,3.2071)
        (22,4.511)
        };
    
    \addplot[style={black, pattern=crosshatch, pattern color=red,mark=none, /pgf/number format/fixed,
        /pgf/number format/precision=2}] coordinates {
        (7,0.3746)
        (9,0.3762)
        (10,0.3762)
        (16,0.4973)
        (17,0.4973)
        (18,0.5005)
        };
    
    \legend{Fuzzy (linear FMF), Fuzzy (Gaussian FMF), Expert systems, Defeasible arg.}

    \end{axis}
    
    \node [text width=9cm, fill=white] (A) at (5.4,-2.75) {\textbf{(a)} Models instantiated with the Italian-language dataset.};
    
    \end{pgfonlayer}
    
    %\node [fill=none, text width=2.5cm, minimum height=4mm] (WW) at (13.2, 2)  {\tiny NASA-TLX};
    %\node [fill=none, text width=2.5cm, minimum height=13mm] (WW2) at (12.8, 0.3)  {};
    %\path[every node/.style={font=\sffamily\small}]

    %(WW) edge[<-, thick, bend left] node [left] {} (WW2);

\end{tikzpicture}

\vspace{4mm}

\begin{tikzpicture}
\pgfplotsset{width=1.3*\textwidth}
\begin{pgfonlayer}{bg}
    \begin{axis}[
         compat=newest, %Better label placement
        width  = 1\textwidth,
        height = 4cm,
        major x tick style = transparent,
        ybar=4*\pgflinewidth,
        bar width=2mm,
        bar shift=0pt,
        ymajorgrids = true,
        xlabel = {Reasoning models for computational trust inference},
        ylabel = {Rank of Barnstars},
        label style={font=\fontsize{11}{10.5}\selectfont},
        tick label style={font=\fontsize{11}{10.5}\selectfont},
        nodes near coords,
        nodes near coords align={vertical},
        every node near coord/.append style={font=\fontsize{8}{7.5}\selectfont},
        every node near coord/.append style={rotate=90, anchor=west},
        %symbolic x coords={FL1 FC1, FL2 FC2, FL3 FC3, FL4 FC4, FL5 FC5, FL6 FC6, FL7 FC7, FL8 FC8, FL9 FC9, FL10 FC10, FL11 FC11, FL12 FC12, A1, A2, A3, A4, 
%E1, E2, E3, E4},
        %xtick={ },
        %x tick label style={font=\scriptsize,  align=left,text width=3.5cm},
        xtick={1,2,3,4,5,6,7,8,9,10,11,12,13,14,15,16,17,18,19,20,21,22,23,24,25,26,27,28,29,30,31,32,33,34},
        xticklabels={
        $\star$ $\circ$ FL3,
        $\triangleright$ $\circ$ FL9,
        $\star$ $\circ$ FL5,
        $\triangleright$ $\circ$ FL11,
        $\star$ $\ddagger$ A2,
        $\star$ $(h_3)$ E3,
        $\star$ $\otimes$ A1,
        $\star$ $\odot$ A3,
        $\star$ $\circ$ FL1,
        $\triangleright$ $\circ$ FL7,
        $\star$ $\circ$ FC5,
        $\triangleright$ $\circ$ FC11,
        $\star$ $\circ$ FC3,
        $\triangleright$ $\circ$ FC9,
        $\triangleright$ $\otimes$ A4,
        $\triangleright$ $\odot$ A6,
        $\triangleright$ $(h_4)$ E4,
        $\triangleright$ $\ddagger$ A5, 
        $\star$ $\circ$ FC1,
        $\triangleright$ $\circ$ FC7,
        $\star$ $(h_1)$ E1,
        $\triangleright$ $(h_2)$ E2,
        $\star$ $\bullet$ FL6,
        $\triangleright$ $\bullet$ FL12,
        $\star$ $\bullet$ FC6,
        $\triangleright$ $\bullet$ FC12,
        $\star$ $\bullet$ FL2,
        $\star$ $\bullet$ FL4,
        $\triangleright$ $\bullet$ FL8,
        $\triangleright$ $\bullet$ FL10,
        $\star$ $\bullet$ FC2,
        $\star$ $\bullet$ FC4,
        $\triangleright$ $\bullet$ FC8,
        $\triangleright$ $\bullet$ FC10},
        x tick label style={font=\fontsize{8}{8.5}\selectfont, align=left, rotate=90, text width=1.3cm},
    %hide obscured x ticks=false,
        %tick align=inside,
        %xtick = data,      
        %tickwidth=4pt,
        %scaled y ticks = false,
        enlarge x limits=0.05,
        ymin=0,
        ymax=23,
        clip=false,
        %nodes near coords,
        %every node near coord/.append style={color=black, font=\tiny},
        %legend cell align=left,
        %legend style={
        %        at={(1,1.05)},
        %        anchor=north,
        %        column sep=1ex
        %},
        legend style={at={(0.5,1.30)},
        anchor=north,legend columns=-1,  
        font=\fontsize{7.5}{8.5}\selectfont},
    ]
    
     %\draw [thin, dotted, thick, draw=black] 
     %    (axis cs: -0.5, 366.152) -- (axis cs: 33.5, 366.152);
%         node[pos=0.995, right] {\fontsize{4}{4.5}\selectfont \textbf{NASA}};
    
   \draw [thin, dotted, draw=black] 
        (axis cs: -0.5, 7.33) -- (axis cs: 35.7, 7.33)
        node[pos=0.4, above] {\fontsize{7}{7.5}\selectfont Features' average (7.33)};
  
   \addplot[style={black,pattern=north east lines, pattern color=blue,mark=none, /pgf/number format/fixed,
        /pgf/number format/precision=2}] coordinates {
        (1,0.2479)
        (2,0.2479)
        (3,0.2648)
        (4,0.2648)
        (9,0.3422)
        (10,0.3422)
        (23,5.1866)
        (24,5.1866)
        (27,14.0546)
        (28,14.0546)
        (29,14.0546)
        (30,14.0546)};

    \addplot[style={black, pattern=crosshatch dots, pattern color=blue,mark=none, /pgf/number format/fixed,
        /pgf/number format/precision=2}] coordinates {
        (11,0.4105)
        (12,0.4105)
        (13,0.4391)
        (14,0.4391)
        (19,0.7208)
        (20,0.7208)
        (25,5.9522)
        (26,5.9522)
        (31,14.0546)
        (32,14.0546)
        (33,14.0546)
        (34,14.0546)};

    \addplot[style={black, pattern=grid, pattern color=green,mark=none, /pgf/number format/fixed,
        /pgf/number format/precision=2}] coordinates {
        (6,0.2761)
        (17,0.4584)
        (21,1.2439)
        (22,1.7756)};
    
    \addplot[style={black, pattern=crosshatch, pattern color=red,mark=none, /pgf/number format/fixed,
        /pgf/number format/precision=2}] coordinates {
        (5,0.2761)
        (7,0.2763)
        (8,0.2763)
        (15,0.4544)
        (16,0.4544)
        (18,0.4627)};
    
    \legend{Fuzzy (linear FMF), Fuzzy (Gaussian FMF), Expert systems, Defeasible arg.}

    \end{axis}
    
    \node [text width=10cm, fill=white] (A) at (5.4,-2.75) {\textbf{(b)} Models instantiated with the Portuguese-language dataset.};
    
    \end{pgfonlayer}
    
    %\node [fill=none, text width=2.5cm, minimum height=4mm] (WW) at (13.2, 2)  {\tiny NASA-TLX};
    %\node [fill=none, text width=2.5cm, minimum height=13mm] (WW2) at (12.8, 0.3)  {};
    %\path[every node/.style={font=\sffamily\small}]

    %(WW) edge[<-, thick, bend left] node [left] {} (WW2);

\end{tikzpicture}

\caption{Normalised sum of the rank of Barnstars users for models built with knowledge base 1 (\ref{app:trust}).
Inferior symbols are used to represent: centroid ($\circ$) and mean of max ($\bullet$) defuzzification approach; heuristics $h_1$ to  
$h_4$; grounded ($\odot$), preferred ($\otimes$), and categoriser semantics ($\ddagger$); and use (respectively no use) of the rule 
weights/arguments strength ($\triangleright$, respectively $\star$).}
\label{figure:rank_KB1}
\end{figure}

\begin{figure} [!h]
\centering

\begin{tikzpicture}
\pgfplotsset{width=1\textwidth}
\begin{pgfonlayer}{bg}
    \begin{axis}[
         compat=newest, %Better label placement
        width  = 1\textwidth,
        height = 4cm,
        major x tick style = transparent,
        ybar=4*\pgflinewidth,
        bar width=2mm,
        bar shift=0pt,
        ymajorgrids = true,
        xlabel = {Reasoning models for computational trust inference},
        ylabel = {Rank of Barnstars},
        label style={font=\fontsize{11}{10.5}\selectfont},
        tick label style={font=\fontsize{11}{10.5}\selectfont},
        nodes near coords,
        nodes near coords align={vertical},
        every node near coord/.append style={font=\fontsize{8}{7.5}\selectfont},
        every node near coord/.append style={rotate=90, anchor=west},
        %symbolic x coords={FL1 FC1, FL2 FC2, FL3 FC3, FL4 FC4, FL5 FC5, FL6 FC6, FL7 FC7, FL8 FC8, FL9 FC9, FL10 FC10, FL11 FC11, FL12 FC12, A1, A2, A3, A4, 
%E1, E2, E3, E4},
        %xtick={ },
        %x tick label style={font=\scriptsize,  align=left,text width=3.5cm},
        xtick={1,2,3,4,5,6,7,8,9,10,11,12,13,14,15,16,17,18,19,20,21,22,23,24,25,26,27,28,29,30,31,32,33,34},
        xticklabels={
        $\star$ $\odot$ A9,
        $\star$ $(h_1)$ E5,
        $\star$ $(h_3)$  E7,
        $\triangleright$ $(h_2)$ E6,
        $\triangleright$ $(h_4)$ E8,
        $\star$ $\circ$ FL15,
        $\star$ $\bullet$ FL16,
        $\triangleright$ $\circ$ FL21,
        $\triangleright$ $\bullet$ FL22,
        $\star$ $\bullet$ FL18,
        $\triangleright$ $\bullet$ FL24,
        $\star$ $\circ$ FL17,
        $\triangleright$ $\circ$ FL23,
        $\star$ $\bullet$ FL14,
        $\triangleright$ $\bullet$ FL20,
        $\star$ $\circ$ FL13,
        $\triangleright$ $\circ$ FL19,
        $\star$ $\circ$ FC15,
        $\triangleright$ $\circ$ FC21,
        $\star$ $\circ$ FC17,
        $\triangleright$ $\circ$ FC23,
        $\star$ $\bullet$ FC16,
        $\triangleright$ $\bullet$ FC22,
        $\star$ $\bullet$ FC18,
        $\triangleright$ $\bullet$ FC24,
        $\star$ $\bullet$ FC14,
        $\triangleright$ $\bullet$ FC20,
        $\star$ $\circ$ FC13,
        $\triangleright$ $\circ$ FC19,
        $\star$ $\otimes$ A7,
        $\star$ $\ddagger$ A8,
        $\triangleright$ $\otimes$ A10,
        $\triangleright$ $\odot$ A12,
        $\triangleright$ $\ddagger$ A11},
        x tick label style={font=\fontsize{8}{8.5}\selectfont, align=left, rotate=90, text width=1.3cm},
    %hide obscured x ticks=false,
        %tick align=inside,
        %xtick = data,      
        %tickwidth=4pt,
        %scaled y ticks = false,
        enlarge x limits=0.05,
        ymin=0,
        ymax=13,
        clip=false,
        %nodes near coords,
        %every node near coord/.append style={color=black, font=\tiny},
        %legend cell align=left,
        %legend style={
        %        at={(1,1.05)},
        %        anchor=north,
        %        column sep=1ex
        %},
        legend style={at={(0.5,1.30)},
        anchor=north,legend columns=-1,  
        font=\fontsize{7.5}{8.5}\selectfont},
    ]
    
     %\draw [thin, dotted, thick, draw=black] 
     %    (axis cs: -0.5, 366.152) -- (axis cs: 33.5, 366.152);
%         node[pos=0.995, right] {\fontsize{4}{4.5}\selectfont \textbf{NASA}};

    \draw [thin, dotted, draw=black] 
        (axis cs: -0.5, 7.54) -- (axis cs: 35.7, 7.54)
        node[pos=0.4, above] {\fontsize{7}{7.5}\selectfont Features' average (7.54)};
  
   \addplot[style={black,pattern=north east lines, pattern color=blue,mark=none, /pgf/number format/fixed,
        /pgf/number format/precision=2}] coordinates {
        (6,1.267)
        (7,1.267)
        (8,1.267)
        (9,1.267)
        (10,1.2724)
        (11,1.2724)
        (12,1.2726)
        (13,1.2726)
        (14,1.5923)
        (15,1.5923)
        (16,1.5924)
        (17,1.5924)};

    \addplot[style={black, pattern=crosshatch dots, pattern color=blue,mark=none, /pgf/number format/fixed,
        /pgf/number format/precision=2}] coordinates {
        (18,1.7347)
        (19,1.7347)
        (20,1.7437)
        (21,1.7437)
        (22,1.7919)
        (23,1.7919)
        (24,1.7954)
        (25,1.7954)
        (26,1.9569)
        (27,1.9569)
        (28,1.9574)
        (29,1.9574)};

    \addplot[style={black, pattern=grid, pattern color=green,mark=none, /pgf/number format/fixed,
        /pgf/number format/precision=2}] coordinates {
        (2,0.0008)
        (3,0.0008)
        (4,0.0009)
        (5,0.0009)};
    
    \addplot[style={black, pattern=crosshatch, pattern color=red,mark=none, /pgf/number format/fixed,
        /pgf/number format/precision=2}] coordinates {
        (1,0.0008)
        (30,2.3566)
        (31,5.2983)
        (32,8.1897)
        (33,8.1897)
        (34,9.4964)};
    
    \legend{Fuzzy (linear FMF), Fuzzy (Gaussian FMF), Expert systems, Defeasible arg.}

    \end{axis}
    
    \node [text width=9cm, fill=white] (A) at (5.4,-2.75) {\textbf{(a)} Models instantiated with the Italian-language dataset.};
    
      \end{pgfonlayer}
      
      %\node [fill=none, text width=2.5cm, minimum height=4mm] (WW) at (13.2, 2)  {\tiny NASA-TLX};
      %\node [fill=none, text width=2.5cm, minimum height=13mm] (WW2) at (12.8, 0.3)  {};
      %\path[every node/.style={font=\sffamily\small}]
    
      %(WW) edge[<-, thick, bend left] node [left] {} (WW2);
   
\end{tikzpicture}

\vspace{4mm}

\begin{tikzpicture}
\pgfplotsset{width=1\textwidth}
\begin{pgfonlayer}{bg}
    \begin{axis}[
         compat=newest, %Better label placement
        width  = 1\textwidth,
        height = 4cm,
        major x tick style = transparent,
        ybar=4*\pgflinewidth,
        bar width=2mm,
        bar shift=0pt,
        ymajorgrids = true,
        xlabel = {Reasoning models for computational trust inference},
        ylabel = {Rank of Barnstars},
        label style={font=\fontsize{11}{10.5}\selectfont},
        tick label style={font=\fontsize{11}{10.5}\selectfont},
        nodes near coords,
        nodes near coords align={vertical},
        every node near coord/.append style={font=\fontsize{8}{7.5}\selectfont},
        every node near coord/.append style={rotate=90, anchor=west},
        %symbolic x coords={FL1 FC1, FL2 FC2, FL3 FC3, FL4 FC4, FL5 FC5, FL6 FC6, FL7 FC7, FL8 FC8, FL9 FC9, FL10 FC10, FL11 FC11, FL12 FC12, A1, A2, A3, A4, 
%E1, E2, E3, E4},
        %xtick={ },
        %x tick label style={font=\scriptsize,  align=left,text width=3.5cm},
        xtick={1,2,3,4,5,6,7,8,9,10,11,12,13,14,15,16,17,18,19,20,21,22,23,24,25,26,27,28,29,30,31,32,33,34},
        xticklabels={
        $\star$ $\odot$ A9,
        $\star$ $(h_1)$ E5,
        $\star$ $(h_3)$  E7,
        $\triangleright$ $(h_2)$ E6,
        $\triangleright$ $(h_4)$ E8,
        $\star$ $\otimes$ A7,
        $\star$ $\bullet$ FL18,
        $\triangleright$ $\bullet$ FL24,
        $\star$ $\circ$ FL17,
        $\triangleright$ $\circ$ FL23,
        $\star$ $\bullet$ FL16,
        $\triangleright$ $\bullet$ FL22,
        $\star$ $\circ$ FL15,
        $\triangleright$ $\circ$ FL21,
        $\star$ $\bullet$ FL14,
        $\triangleright$ $\bullet$ FL20,
        $\star$ $\circ$ FL13,
        $\triangleright$ $\circ$ FL19,
        $\star$ $\circ$ FC17,
        $\triangleright$ $\circ$ FC23,
        $\star$ $\circ$ FC15,
        $\triangleright$ $\circ$ FC21,
        $\star$ $\bullet$ FC18,
        $\triangleright$ $\bullet$ FC24,
        $\star$ $\bullet$ FC16,
        $\triangleright$ $\bullet$ FC22,
        $\star$ $\bullet$ FC14,
        $\triangleright$ $\bullet$ FC20,
        $\star$ $\circ$ FC13,
        $\triangleright$ $\circ$ FC19,    
        $\star$ $\ddagger$ A8,
        $\triangleright$ $\otimes$ A10,
        $\triangleright$ $\odot$ A12,
        $\triangleright$ $\ddagger$ A11},
        x tick label style={font=\fontsize{8}{8.5}\selectfont, align=left, rotate=90, text width=1.3cm},
    %hide obscured x ticks=false,
        %tick align=inside,
        %xtick = data,      
        %tickwidth=4pt,
        %scaled y ticks = false,
        enlarge x limits=0.05,
        ymin=0,
        ymax=21,
        clip=false,
        %nodes near coords,
        %every node near coord/.append style={color=black, font=\tiny},
        %legend cell align=left,
        %legend style={
        %        at={(1,1.05)},
        %        anchor=north,
        %        column sep=1ex
        %},
        legend style={at={(0.5,1.30)},
        anchor=north,legend columns=-1,  
        font=\fontsize{7.5}{8.5}\selectfont},
    ]
    
     %\draw [thin, dotted, thick, draw=black] 
     %    (axis cs: -0.5, 366.152) -- (axis cs: 33.5, 366.152);
%         node[pos=0.995, right] {\fontsize{4}{4.5}\selectfont \textbf{NASA}};
 
   \draw [thin, dotted, draw=black] 
        (axis cs: -0.5, 7.33) -- (axis cs: 35.7, 7.33)
        node[pos=0.2, above] {\fontsize{7}{7.5}\selectfont Features' average (7.33)};
  
   \addplot[style={black,pattern=north east lines, pattern color=blue,mark=none, /pgf/number format/fixed,
        /pgf/number format/precision=2}] coordinates {
        (7,1.4131)
        (8,1.4131)
        (9,1.4132)
        (10,1.4132)
        (11,1.4243)
        (12,1.4243)
        (13,1.4244)
        (14,1.4244)
        (15,1.9856)
        (16,1.9856)
        (17,1.9857)
        (18,1.9857)};

    \addplot[style={black, pattern=crosshatch dots, pattern color=blue,mark=none, /pgf/number format/fixed,
        /pgf/number format/precision=2}] coordinates {
        (19,2.092)
        (20,2.092)
        (21,2.0998)
        (22,2.0998)
        (23,2.182)
        (24,2.182)
        (25,2.1884)
        (26,2.1884)
        (27,2.4119)
        (28,2.4119)
        (29,2.4719)
        (30,2.4719)};

    \addplot[style={black, pattern=grid, pattern color=green,mark=none, /pgf/number format/fixed,
        /pgf/number format/precision=2}] coordinates {
        (2,0.0006)
        (3,0.0006)
        (4,0.0007)
        (5,0.0007)
        };
    
    \addplot[style={black, pattern=crosshatch, pattern color=red,mark=none, /pgf/number format/fixed,
        /pgf/number format/precision=2}] coordinates {
        (1,0.0005)
        (6,0.6643)
        (31,7.1408)
        (32,12.1298)
        (33,12.1298)
        (34,13.3504)};
    
    \legend{Fuzzy (linear FMF), Fuzzy (Gaussian FMF), Expert systems, Defeasible arg.}

    \end{axis}
    
    \node [text width=10cm, fill=white] (A) at (5.4,-2.75) {\textbf{(b)} Models instantiated with the Portuguese-language dataset.};
    
      \end{pgfonlayer}
      
      %\node [fill=none, text width=2.5cm, minimum height=4mm] (WW) at (13.2, 2)  {\tiny NASA-TLX};
      %\node [fill=none, text width=2.5cm, minimum height=13mm] (WW2) at (12.8, 0.3)  {};
      %\path[every node/.style={font=\sffamily\small}]
    
      %(WW) edge[<-, thick, bend left] node [left] {} (WW2);
   
\end{tikzpicture}

\caption{Normalised sum of the rank of Barnstars users for models built with knowledge base 2 (\ref{app:trust}).
Inferior symbols are used to represent: centroid ($\circ$) and mean of max ($\bullet$) defuzzification approach; heuristics $h_1$ to  
$h_4$; grounded ($\odot$), preferred ($\otimes$), and categoriser semantics ($\ddagger$); and use (respectively no use) of the rule 
weights/arguments strength ($\triangleright$, respectively $\star$).}
\label{figure:rank_kb2}
\end{figure}

From both figures, it is possible to observe that the computed ranks were effective, ranging from 0 (perfect rank) to 14.05 across all models and input datasets. This 
suggests that the non-monotonic reasoning approaches were all capable of capturing, to some degree, the notions of the 
ill-defined construct of trust, even when presenting worse values compared to the features' average. Moreover, the use of different datasets does not seem to have a significant impact on the produced rank of Barnstars. It highlights the stability of models when using different sets of data. However, some differences can still be noted among the models built with different knowledge bases.
Different results were expected especially due to the contrasting topologies of these two knowledge bases. Fig. \ref{figure:kb6} and \ref{figure:kb7} (p. \pageref{figure:kb6} and p. \pageref{figure:kb7}) depict their graphical representations. Particularly, while certain parameters appear to be 
effective for models built with one knowledge base, the same parameters can be less effective when the other knowledge base is applied instead. For instance:
\begin{enumerate}
 
 \item Expert system models provided ranks between 0.28 - 4.51 when built with KB1 and perfect ranks when built with KB2. The variance when employing KB1 was observed due to the filtering of surviving rules proposed by heuristics  $h_1$ and $h_2$, which seems to diminish the quality of the ranks produced  ($E1$, $E2$ $\times$ $E3$, $E4$). Noticeably, it does not seem to affect the models built with the same heuristics but using KB2;
 
 \item Fuzzy reasoning models presented higher variance when instantiated with KB1 (ranks 
between 0.25 - 14.05) than with KB2 (ranks between 1.27 - 2.47). The use of linear or Gaussian FMF did not contribute to this variance or had a significant impact in the quality of the ranks produced. Instead, when built with KB1, fuzzy models using the centroid defuzzification approach (labelled with $\circ$) seem to be much more effective than those using the max defuzzification approach (labelled with $\bullet$). When built with KB2, no particular parameter seems to provide significantly better ranks, possibly due to the difficulty of these models in dealing with a higher number of contradictions contained in this knowledge base; 

 \item Argument-based models demonstrated great stability when instantiated with KB1 (ranks between 0.28 - 0.5), but higher variance when instantiated with KB2 (ranks between 0 - 13.35). It shows that acceptability semantics and the use (or not) of 
strength or arguments did not impact results when using KB1. Contrarily, for the argument-based models built with KB2, the use of grounded and preferred 
semantics (labelled with $\odot$ and $\otimes$) resulted in better ranks. 
%It suggests that, when using KB2, no single argument resulting from the categoriser semantics is as strong as a set of interacting reasons (extensions). In other words, Dung's extensions provided more appealing trust scalars to rank Barnstar editors.
Lastly, models built with KB2 and no strength of arguments (labelled with $\star$) also performed better than their counterparts (labelled with $\triangleright$), suggesting that the strengths of arguments defined by the author did not improve the rank of Barnstar editors.

\end{enumerate}

In summary, the first knowledge base presents a simple topology. It seems to work more effectively with argument-based models, a subset of expert system models, and a subset of
fuzzy reasoning models. Contrarily, the second knowledge base is built with many more attacks, resulting in a more complex topology.
Despite such complexity, certain models did achieve a perfect rank of Barnstar editors. However, as it will be evaluated in the next subsection,
a higher topological complexity might also hamper the capacity of these same models of producing inferences.
Thus, the next subsection evaluates the percentage of NAs, or the percentage of instances without a trust value assigned by each model. It further 
investigates a probable trade-off between these two knowledge bases: while one is more simplified and allows inferences to be produced for all instances, the other is more complex and precise, but prevents models from reaching a conclusion in a number of cases.

\subsection{Percentage of NAs}

The capacity for assigning trust values under conflicting information is assumed to be a favourable property. Thus, it is evaluated through the percentage of 
NAs. It is known that certain acceptability semantics employed by argument-based models, such as the categoriser, will always return a 
non-empty 
extension. In addition, the preferred semantics is less likely to return an empty extension compared to the grounded semantics. In fact, due to the topology 
of the knowledge bases employed in this research article, the preferred semantics always returns a non-empty extension. However, the models implemented with other semantics and other 
reasoning approaches might not reach a final inference for certain cases. Hence, it is important to evaluate the extent to 
which this can impact the quality of the designed reasoning models. Fig. \ref{figure:nasKB6} depicts the percentage of NAs for models instantiated with KB2. Due 
to the simplified topology of the KB1, NAs were not reported for any reasoning model built with it. Similarly to the previous evaluation metric, the instantiation of models with different datasets did not result in different trends.

\begin{figure} [!h]
\centering

\begin{tikzpicture}
\pgfplotsset{width=1\textwidth}
\begin{pgfonlayer}{bg}
    \begin{axis}[
        compat=newest, %Better label placement
        width  = 1\textwidth,
        height = 4cm,
        major x tick style = transparent,
        ybar=4*\pgflinewidth,
        bar width=2mm,
        bar shift=0pt,
        ymajorgrids = true,
        xlabel = {Reasoning models for computational trust inference},
        ylabel = {Percentage of NAs},
        label style={font=\fontsize{11}{10.5}\selectfont},
        tick label style={font=\fontsize{11}{10.5}\selectfont},
        nodes near coords,
        nodes near coords align={vertical},
        every node near coord/.append style={font=\fontsize{8}{7.5}\selectfont},
        every node near coord/.append style={rotate=90, anchor=west},
        %symbolic x coords={FL1 FC1, FL2 FC2, FL3 FC3, FL4 FC4, FL5 FC5, FL6 FC6, FL7 FC7, FL8 FC8, FL9 FC9, FL10 FC10, FL11 FC11, FL12 FC12, A1, A2, A3, A4, 
%E1, E2, E3, E4},
        %xtick={ },
        %x tick label style={font=\scriptsize,  align=left,text width=3.5cm},
        xtick={1,2,3,4,5,6,7,8,9,10,11,12,13,14,15,16,17,18,19,20,21,22,23,24,25,26,27,28,29,30,31,32,33,34},
        xticklabels={
        $\star$ $\otimes$ A7,
        $\star$ $\ddagger$ A8,
        $\triangleright$ $\otimes$ A10,
        $\triangleright$ $\odot$ A12,
        $\triangleright$ $\ddagger$ A11,
        $\star$ $\bullet$ FC16,
        $\triangleright$ $\bullet$ FC22,
        $\star$ $\bullet$ FC18,
        $\triangleright$ $\bullet$ FC24,
        $\star$ $\circ$ FC15,
        $\triangleright$ $\circ$ FC21,
        $\star$ $\circ$ FC17,
        $\triangleright$ $\circ$ FC23,
        $\star$ $\bullet$ FC14,
        $\triangleright$ $\bullet$ FC20,
        $\star$ $\circ$ FC13,
        $\triangleright$ $\circ$ FC19,
        $\star$ $\circ$ FL15,
        $\star$ $\bullet$ FL16,
        $\triangleright$ $\circ$ FL21,
        $\triangleright$ $\bullet$ FL22,
        $\star$ $\bullet$ FL18,
        $\triangleright$ $\bullet$ FL24,
        $\star$ $\circ$ FL17,
        $\triangleright$ $\circ$ FL23,
        $\star$ $\bullet$ FL14,
        $\triangleright$ $\bullet$ FL20,
        $\star$ $\circ$ FL13,
        $\triangleright$ $\circ$ FL19,
        $\star$ $\odot$ A9,
        $\star$ $(h_1)$ E5,
        $\star$ $(h_3)$ E7,
        $\triangleright$ $(h_2)$ E6,
        $\triangleright$ $(h_4)$ E8},
        x tick label style={font=\fontsize{8}{8.5}\selectfont, align=left, rotate=90, text width=1.3cm},
    %hide obscured x ticks=false,
        %tick align=inside,
        %xtick = data,      
        %tickwidth=4pt,
        %scaled y ticks = false,
        enlarge x limits=0.05,
        ymin=0,
        ymax=90,
        clip=false,
        %nodes near coords,
        %every node near coord/.append style={color=black, font=\tiny},
        %legend cell align=left,
        %legend style={
        %        at={(1,1.05)},
        %        anchor=north,
        %        column sep=1ex
        %},
        legend style={at={(0.5,1.30)},
        anchor=north,legend columns=-1,  
        font=\fontsize{7.5}{8.5}\selectfont},
    ]
    
     %\draw [thin, dotted, thick, draw=black] 
     %    (axis cs: -0.5, 366.152) -- (axis cs: 33.5, 366.152);
%         node[pos=0.995, right] {\fontsize{4}{4.5}\selectfont \textbf{NASA}};

   \addplot[style={black,pattern=north east lines, pattern color=blue,mark=none, /pgf/number format/fixed,
        /pgf/number format/precision=3}] coordinates {
        (18,3.5916)
            (19,3.5916)
            (20,3.5916)
            (21,3.5916)
            (22,3.5916)
            (23,3.5916)
            (24,3.5916)
            (25,3.5916)
            (26,3.5925)
            (27,3.5925)
            (28,3.5925)
            (29,3.5925)
            };

    \addplot[style={black, pattern=crosshatch dots, pattern color=blue,mark=none, /pgf/number format/fixed,
        /pgf/number format/precision=3}] coordinates {
        (6,3.3817)
        (7,3.3817)
        (8,3.3817)
        (9,3.3817)
        (10,3.4271)
        (11,3.4271)
        (12,3.4271)
        (13,3.4271)
        (14,3.5152)
        (15,3.5152)
        (16,3.5629)
        (17,3.5629)
        };

    \addplot[style={black, pattern=grid, pattern color=green,mark=none, /pgf/number format/fixed,
        /pgf/number format/precision=3}] coordinates {
        (31,51.3056)
        (32,51.3056)
        (33,51.3056)
        (34,51.3056)
        };
    
    \addplot[style={black, pattern=crosshatch, pattern color=red,mark=none, /pgf/number format/fixed,
        /pgf/number format/precision=3}] coordinates {
        (1,0)
        (2,0)
        (3,0)
        (4,0)
        (5,0)
        (30,51.3056)
        };
    
    \legend{Fuzzy (linear FMF), Fuzzy (Gaussian FMF), Expert systems, Defeasible arg.}

    \end{axis}
    
    \node [text width=9cm, fill=white] (A) at (6.0,-2.75) {(a) Input data from the Wikipedia Italian edition};
    
      \end{pgfonlayer}
      
      %\node [fill=none, text width=2.5cm, minimum height=4mm] (WW) at (13.2, 2)  {\tiny NASA-TLX};
      %\node [fill=none, text width=2.5cm, minimum height=13mm] (WW2) at (12.8, 0.3)  {};
      %\path[every node/.style={font=\sffamily\small}]
    
      %(WW) edge[<-, thick, bend left] node [left] {} (WW2);
   
\end{tikzpicture}

\vspace{4mm}

\begin{tikzpicture}
\pgfplotsset{width=1\textwidth}
\begin{pgfonlayer}{bg}
    \begin{axis}[
        compat=newest, %Better label placement
        width  = 1\textwidth,
        height = 4cm,
        major x tick style = transparent,
        ybar=4*\pgflinewidth,
        bar width=2mm,
        bar shift=0pt,
        ymajorgrids = true,
        xlabel = {Reasoning models for computational trust inference},
        ylabel = {Percentage of NAs},
        label style={font=\fontsize{11}{10.5}\selectfont},
        tick label style={font=\fontsize{11}{10.5}\selectfont},
        nodes near coords,
        nodes near coords align={vertical},
        every node near coord/.append style={font=\fontsize{8}{7.5}\selectfont},
        every node near coord/.append style={rotate=90, anchor=west},
        %symbolic x coords={FL1 FC1, FL2 FC2, FL3 FC3, FL4 FC4, FL5 FC5, FL6 FC6, FL7 FC7, FL8 FC8, FL9 FC9, FL10 FC10, FL11 FC11, FL12 FC12, A1, A2, A3, A4, 
%E1, E2, E3, E4},
        %xtick={ },
        %x tick label style={font=\scriptsize,  align=left,text width=3.5cm},
        xtick={1,2,3,4,5,6,7,8,9,10,11,12,13,14,15,16,17,18,19,20,21,22,23,24,25,26,27,28,29,30,31,32,33,34},
        xticklabels={
        $\star$ $\otimes$ A7,
        $\star$ $\ddagger$ A8,
        $\triangleright$ $\otimes$ A10,
        $\triangleright$ $\ddagger$ A11,
        $\triangleright$ $\odot$ A12,
        $\star$ $\bullet$ FC18,
        $\triangleright$ $\bullet$ FC24,
        $\star$ $\bullet$ FC16,
        $\triangleright$ $\bullet$ FC22,
        $\star$ $\circ$ FC17,
        $\triangleright$ $\circ$ FC23,
        $\star$ $\circ$ FC15,
        $\triangleright$ $\circ$ FC21,
        $\star$ $\bullet$ FC14,
        $\triangleright$ $\bullet$ FC20,
        $\star$ $\circ$ FC13,
        $\triangleright$ $\circ$ FC19,
        $\star$ $\bullet$ FL18,
        $\triangleright$ $\bullet$ FL24,
        $\star$ $\circ$ FL17,
        $\triangleright$ $\circ$ FL23,
        $\star$ $\bullet$ FL16,
        $\triangleright$ $\bullet$ FL22,
        $\star$ $\circ$ FL15,
        $\triangleright$ $\circ$ FL21,
        $\star$ $\bullet$ FL14,
        $\triangleright$ $\bullet$ FL20,
        $\star$ $\circ$ FL13,
        $\triangleright$ $\circ$ FL19,
        $\star$ $\odot$ A9,
        $\star$ $(h_1)$ E5,
        $\star$ $(h_3)$ E7,
        $\triangleright$ $(h_2)$ E6,
        $\triangleright$ $(h_4)$ E8},
        x tick label style={font=\fontsize{8}{8.5}\selectfont, align=left, rotate=90, text width=1.3cm},
    %hide obscured x ticks=false,
        %tick align=inside,
        %xtick = data,      
        %tickwidth=4pt,
        %scaled y ticks = false,
        enlarge x limits=0.05,
        ymin=0,
        ymax=80,
        clip=false,
        %nodes near coords,
        %every node near coord/.append style={color=black, font=\tiny},
        %legend cell align=left,
        %legend style={
        %        at={(1,1.05)},
        %        anchor=north,
        %        column sep=1ex
        %},
        legend style={at={(0.5,1.30)},
        anchor=north,legend columns=-1,  
        font=\fontsize{7.5}{8.5}\selectfont},
    ]
    
     %\draw [thin, dotted, thick, draw=black] 
     %    (axis cs: -0.5, 366.152) -- (axis cs: 33.5, 366.152);
%         node[pos=0.995, right] {\fontsize{4}{4.5}\selectfont \textbf{NASA}};

   \addplot[style={black,pattern=north east lines, pattern color=blue,mark=none, /pgf/number format/fixed,
        /pgf/number format/precision=3}] coordinates {
        (18,5.3126)
        (19,5.3126)
        (20,5.3126)
        (21,5.3126)
        (22,5.3126)
        (23,5.3126)
        (24,5.3126)
        (25,5.3126)
        (26,5.3152)
        (27,5.3152)
        (28,5.3152)
        (29,5.3152)
        };

    \addplot[style={black, pattern=crosshatch dots, pattern color=blue,mark=none, /pgf/number format/fixed,
        /pgf/number format/precision=3}] coordinates {
        (6,4.9446)
        (7,4.9446)
        (8,4.9446)
        (9,4.9446)
        (10,5.0053)
        (11,5.0053)
        (12,5.0054)
        (13,5.0054)
        (14,5.1042)
        (15,5.1042)
        (16,5.1693)
        (17,5.1693)
        };

    \addplot[style={black, pattern=grid, pattern color=green,mark=none, /pgf/number format/fixed,
        /pgf/number format/precision=3}] coordinates {
        (31,50.4301)
        (32,50.4301)
        (33,50.4301)
        (34,50.4301)
        };
    
    \addplot[style={black, pattern=crosshatch, pattern color=red,mark=none, /pgf/number format/fixed,
        /pgf/number format/precision=3}] coordinates {
        (1,0)
        (2,0)
        (3,0)
        (4,0)
        (5,0)
        (30,50.4301)
        };
    
    \legend{Fuzzy (linear FMF), Fuzzy (Gaussian FMF), Expert systems, Defeasible arg.}

    \end{axis}
    
     \node [text width=9cm, fill=white] (A) at (6.0,-2.75) {(b) Input data from the Wikipedia Portuguese edition};
    
      \end{pgfonlayer}
      
      %\node [fill=none, text width=2.5cm, minimum height=4mm] (WW) at (13.2, 2)  {\tiny NASA-TLX};
      %\node [fill=none, text width=2.5cm, minimum height=13mm] (WW2) at (12.8, 0.3)  {};
      %\path[every node/.style={font=\sffamily\small}]
    
      %(WW) edge[<-, thick, bend left] node [left] {} (WW2);
   
\end{tikzpicture}

\caption{Percentage of instances without a computational trust scalar assigned 
by models built with knowledge base 2 (\ref{app:trust}).
Inferior symbols are used to represent: centroid ($\circ$) and mean of max ($\bullet$) defuzzification approach; heuristics $h_1$ to  
$h_4$; grounded ($\odot$), preferred ($\otimes$), and categoriser ($\ddagger$) semantics; and use (respectively no use) of the rule 
weights/arguments strength ($\triangleright$, respectively $\star$).}
\label{figure:nasKB6}
\end{figure}

As for expert system models, it seems clear that the simplistic conflict resolution strategy employed by them might work well when applied to knowledge bases of 
simplified topology (as per Fig. \ref{figure:rank_kb2}). However, once a higher number of conflicts is presented several instances might not have an assigned 
inference. When instantiated with KB2, 51.3\% (Italian-language dataset) and 50.43\% (Portuguese-language dataset) NAs were reported for all expert system models. Thus, limiting the applicability of the reasoning approach with KB2.

With respect to fuzzy reasoning models, similar percentages of NAs were observed, indicating an equivalent capability of resolving the conflicts provided in 
KB2. Thus, note that the use of possibility theory for the resolution of large amounts of conflicts does not appear to be impacted by other 
configuration parameters of fuzzy reasoning models in this knowledge base.

Finally, in relation to argument-based models, it is possible to observe that $A12$
and $A9$, both built with the grounded semantics, presented very different results. While $A12$ makes use of strength of arguments and presented 0\% of NAs, 
$A9$ makes no use of strength of arguments and presented 51.3\% (Italian-language dataset) and 50.43\% (Portuguese-language dataset) of NAs. Therefore, the use of strength of arguments can assist in the issue of empty 
extensions when employing the grounded semantics. However, the quality of the inferences is not maintained when using such strengths.
$A9$ had a perfect rank, while $A12$ had a rank value of 12.13 as depicted in 
Fig. \ref{figure:rank_kb2}. This is an indication that the strengths defined by the author helped to 
solve the excessive amount of conflicts in KB2, but did not seem to enhance the rank of Barnstar editors. This is 
likely due to the way rule weights and strength of arguments were defined, feature by feature. 
Weights/strengths could have been defined for each rule and arguments directly, in a more time-consuming manner and requiring more domain knowledge, but likely better capable of quantifying their importance.
The only argument-based models that could achieve a strong rank of Barnstars while not reporting NAs were 
$A7$ (preferred semantics) and $A8$ (categoriser semantics), which were built with no strength of arguments. It reinforces the suitability of the preferred semantics and the categoriser semantics (despite worse rank of Barnstars editors)
for the inference of computational trust.

\subsection{Spread}

Another metric selected for evaluating the quality of the inferences produced by the non-monotonic reasoning models was the spread of the 
trust scalars assigned to Barnstar editors. This was measured through the standard deviation ($\sigma$) of these scalars. As previously mentioned, trust is not 
a binary concept. Thus, if we assume that Barnstar editors are trustworthy, we 
should also expect that they will have different trust levels. Fig. \ref{figure:kb1_spread} and \ref{figure:kb2_spread} depicts the results for the models built with KB1 and KB2. The baseline instrument is depicted again. It produces trust scalars through the average of the normalised values of the features reported by each editor. It is employed as an attempt to show whether the non-monotonic reasoning processes implemented could outperform, to some degree, a non-deductive and simplified inference.

\begin{figure} [!h]
\centering

\begin{tikzpicture}
\pgfplotsset{width=1\textwidth}
\begin{pgfonlayer}{bg}
    \begin{axis}[
        compat=newest, %Better label placement
        width  = 1\textwidth,
        height = 4cm,
        major x tick style = transparent,
        ybar=4*\pgflinewidth,
        bar width=2mm,
        bar shift=0pt,
        ymajorgrids = true,
        xlabel = {Reasoning models for computational trust inference},
        ylabel = {Standard deviation},
        label style={font=\fontsize{11}{10.5}\selectfont},
        tick label style={font=\fontsize{11}{10.5}\selectfont},
        nodes near coords,
        nodes near coords align={vertical},
        every node near coord/.append style={font=\fontsize{8}{7.5}\selectfont},
        every node near coord/.append style={rotate=90, anchor=west},
        %symbolic x coords={FL1 FC1, FL2 FC2, FL3 FC3, FL4 FC4, FL5 FC5, FL6 FC6, FL7 FC7, FL8 FC8, FL9 FC9, FL10 FC10, FL11 FC11, FL12 FC12, A1, A2, A3, A4, 
%E1, E2, E3, E4},
        %xtick={ },
        %x tick label style={font=\scriptsize,  align=left,text width=3.5cm},
        xtick={1,2,3,4,5,6,7,8,9,10,11,12,13,14,15,16,17,18,19,20,21,22,23,24,25,26,27,28,29,30,31,32,33,34},
        xticklabels={
        $\star$ $\circ$ FL3,
        $\triangleright$ $\circ$ FL9,
        $\star$ $\circ$ FL5,
        $\triangleright$ $\circ$ FL11,
        $\star$ $\circ$ FL1,
        $\triangleright$ $\circ$ FL7,
        $\star$ $\ddagger$ A2,
        $\star$ $(h_3)$ E3,
        $\star$ $\otimes$ A1,
        $\star$ $\odot$ A3,
        $\star$ $\circ$ FC3,
        $\triangleright$ $\circ$ FC9,
        $\star$ $\circ$ FC5,
        $\triangleright$ $\circ$ FC11,
        $\triangleright$ $(h_4)$ E4,
        $\triangleright$ $\otimes$ A4,
        $\triangleright$ $\odot$ A6,
        $\triangleright$ $\ddagger$ A5,
        $\star$ $\circ$ FC1,
        $\triangleright$ $\circ$ FC7,
        $\star$ $(h_1)$ E1,
        $\triangleright$ $(h_2)$ E2,
        $\star$ $\bullet$ FL6,
        $\triangleright$ $\bullet$ FL12,
        $\star$ $\bullet$ FC6,
        $\triangleright$ $\bullet$ FC12,
        $\star$ $\bullet$ FL2,
        $\star$ $\bullet$ FL4,
        $\triangleright$ $\bullet$ FL8,
        $\triangleright$ $\bullet$ FL10,
        $\star$ $\bullet$ FC2,
        $\star$ $\bullet$ FC4,
        $\triangleright$ $\bullet$ FC8,
        $\triangleright$ $\bullet$ FC10
        },
        x tick label style={font=\fontsize{8}{8.5}\selectfont, align=left, rotate=90, text width=1.3cm},
    %hide obscured x ticks=false,
        %tick align=inside,
        %xtick = data,      
        %tickwidth=4pt,
        %scaled y ticks = false,
        enlarge x limits=0.05,
        ymin=0,
        ymax=0.34,
        clip=false,
        %nodes near coords,
        %every node near coord/.append style={color=black, font=\tiny},
        %legend cell align=left,
        %legend style={
        %        at={(1,1.05)},
        %        anchor=north,
        %        column sep=1ex
        %},
        legend style={at={(0.5,1.30)},
        anchor=north,legend columns=-1,  
        font=\fontsize{7.5}{8.5}\selectfont},
    ]
    
     %\draw [thin, dotted, thick, draw=black] 
     %    (axis cs: -0.5, 366.152) -- (axis cs: 33.5, 366.152);
%         node[pos=0.995, right] {\fontsize{4}{4.5}\selectfont \textbf{NASA}};
    
   \draw [thin, dotted, draw=black] 
        (axis cs: -0.5, 0.07) -- (axis cs: 35.7, 0.07);
        %node[pos=0.85, above] {\fontsize{5}{5.5}\selectfont \textbf{Features' average (0.07)}}; 

   \addplot[style={black,pattern=north east lines, pattern color=blue,mark=none, /pgf/number format/fixed,
        /pgf/number format/precision=3}] coordinates {
        (1,0.1893)
        (2,0.1767)
        (3,0.1175)
        (4,0.1175)
        (5,0.1175)
        (6,0.1175)
        (23,0.0697)
        (24,0.0697)
        (27,0)
        (28,0)
        (29,0)
        (30,0)
        };

    \addplot[style={black, pattern=crosshatch dots, pattern color=blue,mark=none, /pgf/number format/fixed,
        /pgf/number format/precision=3}] coordinates {
        (11,0.0837)
        (12,0.0837)
        (13,0.082)
        (14,0.082)
        (19,0.0788)
        (20,0.0788)
        (25,0.0395)
        (26,0.0395)
        (31,0)
        (32,0)
        (33,0)
        (34,0)
        };

    \addplot[style={black, pattern=grid, pattern color=green,mark=none, /pgf/number format/fixed,
        /pgf/number format/precision=3}] coordinates {
        (8,0.1143)
        (15,0.0815)
        (21,0.0744)
        (22,0.0744)
        };
    
    \addplot[style={black, pattern=crosshatch, pattern color=red,mark=none, /pgf/number format/fixed,
        /pgf/number format/precision=3}] coordinates {
        (7,0.1143)
        (9,0.1143)
        (10,0.1143)
        (16,0.0815)
        (17,0.0794)
        (18,0.0794)
        };
    
    \legend{Fuzzy (linear FMF), Fuzzy (Gaussian FMF), Expert systems, Defeasible arg.}

    \end{axis}
    
    \node [text width=9cm, fill=white] (A) at (5.4,-2.75) {\textbf{(a)} Models instantiated with the Italian-language dataset.};
    
      \end{pgfonlayer}
      
      %node[pos=0.85, above] {\fontsize{5}{5.5}\selectfont \textbf{Features' average (0.07)}}; 
      \node [fill=none, text width=3cm, minimum height=4mm] (WW) at (8.7, 1.5)  {\fontsize{7}{7.5}\selectfont Features' average (0.07)};
      \node [fill=none, text width=2.5cm, minimum height=13mm] (WW2) at (8.3, -0.14)  {};
      \path[every node/.style={font=\sffamily\small}]
    
      (WW) edge[<-, thick, bend left] node [left] {} (WW2);
      
      %\node [fill=none, text width=2.5cm, minimum height=4mm] (WW) at (13.2, 2)  {\tiny NASA-TLX};
      %\node [fill=none, text width=2.5cm, minimum height=13mm] (WW2) at (12.8, 0.3)  {};
      %\path[every node/.style={font=\sffamily\small}]
    
      %(WW) edge[<-, thick, bend left] node [left] {} (WW2);
   
\end{tikzpicture}

\vspace{4mm}

\begin{tikzpicture}
\pgfplotsset{width=1\textwidth}
\begin{pgfonlayer}{bg}
    \begin{axis}[
        compat=newest, %Better label placement
        width  = 1\textwidth,
        height = 4cm,
        major x tick style = transparent,
        ybar=4*\pgflinewidth,
        bar width=2mm,
        bar shift=0pt,
        ymajorgrids = true,
        xlabel = {Reasoning models for computational trust inference},
        ylabel = {Standard deviation},
        label style={font=\fontsize{11}{10.5}\selectfont},
        tick label style={font=\fontsize{11}{10.5}\selectfont},
        nodes near coords,
        nodes near coords align={vertical},
        every node near coord/.append style={font=\fontsize{8}{7.5}\selectfont},
        every node near coord/.append style={rotate=90, anchor=west},
        %symbolic x coords={FL1 FC1, FL2 FC2, FL3 FC3, FL4 FC4, FL5 FC5, FL6 FC6, FL7 FC7, FL8 FC8, FL9 FC9, FL10 FC10, FL11 FC11, FL12 FC12, A1, A2, A3, A4, 
%E1, E2, E3, E4},
        %xtick={ },
        %x tick label style={font=\scriptsize,  align=left,text width=3.5cm},
        xtick={1,2,3,4,5,6,7,8,9,10,11,12,13,14,15,16,17,18,19,20,21,22,23,24,25,26,27,28,29,30,31,32,33,34},
        xticklabels={
        $\triangleright$ $(h_2)$ E2,
        $\star$ $(h_1)$ E1,
        $\triangleright$ $\otimes$ A4,
        $\triangleright$ $\odot$ A6,
        $\triangleright$ $(h_4)$ E4,
        $\triangleright$ $\ddagger$ A5,
        $\star$ $\otimes$ A1,
        $\star$ $\odot$ A3,
        $\star$ $\ddagger$ A2,
        $\star$ $(h_3)$ E3,
        $\star$ $\circ$ FC5,
        $\triangleright$ $\circ$ FC11,
        $\star$ $\circ$ FC3,
        $\triangleright$ $\circ$ FC9,
        $\star$ $\circ$ FC1,
        $\triangleright$ $\circ$ FC7,
        $\star$ $\circ$ FL5,
        $\triangleright$ $\circ$ FL11,
        $\star$ $\circ$ FL3,
        $\triangleright$ $\circ$ FL9,
        $\star$ $\circ$ FL1,
        $\triangleright$ $\circ$ FL7,
        $\star$ $\bullet$ FC6,
        $\triangleright$ $\bullet$ FC12,
        $\star$ $\bullet$ FL6,
        $\triangleright$ $\bullet$ FL12,
        $\star$ $\bullet$ FL2,
        $\star$ $\bullet$ FL4,
        $\triangleright$ $\bullet$ FL8,
        $\triangleright$ $\bullet$ FL10,
        $\star$ $\bullet$ FC2,
        $\star$ $\bullet$ FC4,
        $\triangleright$ $\bullet$ FC8,
        $\triangleright$ $\bullet$ FC10
        },
        x tick label style={font=\fontsize{8}{8.5}\selectfont, align=left, rotate=90, text width=1.3cm},
    %hide obscured x ticks=false,
        %tick align=inside,
        %xtick = data,      
        %tickwidth=4pt,
        %scaled y ticks = false,
        enlarge x limits=0.05,
        ymin=0,
        ymax=0.34,
        clip=false,
        %nodes near coords,
        %every node near coord/.append style={color=black, font=\tiny},
        %legend cell align=left,
        %legend style={
        %        at={(1,1.05)},
        %        anchor=north,
        %        column sep=1ex
        %},
        legend style={at={(0.5,1.30)},
        anchor=north,legend columns=-1,  
        font=\fontsize{7.5}{8.5}\selectfont},
    ]
    
     %\draw [thin, dotted, thick, draw=black] 
     %    (axis cs: -0.5, 366.152) -- (axis cs: 33.5, 366.152);
%         node[pos=0.995, right] {\fontsize{4}{4.5}\selectfont \textbf{NASA}};

   \draw [thin, dotted, draw=black] 
        (axis cs: -0.5, 0.08) -- (axis cs: 35.7, 0.08);
   
   \addplot[style={black,pattern=north east lines, pattern color=blue,mark=none, /pgf/number format/fixed,
        /pgf/number format/precision=3}] coordinates {
        (17,0.0742)
        (18,0.0742)
        (19,0.0728)
        (20,0.0728)
        (21,0.0698)
        (22,0.0698)
        (25,0.0229)
        (26,0.0229)
        (27,0)
        (28,0)
        (29,0)
        (30,0)};

    \addplot[style={black, pattern=crosshatch dots, pattern color=blue,mark=none, /pgf/number format/fixed,
        /pgf/number format/precision=3}] coordinates {
        (11,0.0777)
        (12,0.0777)
        (13,0.0769)
        (14,0.0769)
        (15,0.0758)
        (16,0.0758)
        (23,0.0418)
        (24,0.0418)
        (31,0)
        (32,0)
        (33,0)
        (34,0)};

    \addplot[style={black, pattern=grid, pattern color=green,mark=none, /pgf/number format/fixed,
        /pgf/number format/precision=3}] coordinates {
        (1,0.1379)
        (2,0.1308)
        (5,0.111)
        (10,0.1028)};
    
    \addplot[style={black, pattern=crosshatch, pattern color=red,mark=none, /pgf/number format/fixed,
        /pgf/number format/precision=3}] coordinates {
        (3,0.1113)
        (4,0.1113)
        (6,0.111)
        (7,0.1035)
        (8,0.1035)
        (9,0.1028)};
    
    \legend{Fuzzy (linear FMF), Fuzzy (Gaussian FMF), Expert systems, Defeasible arg.}

    \end{axis}
    
    \node [text width=10cm, fill=white] (A) at (5.4,-2.75) {\textbf{(b)} Models instantiated with the Portuguese-language dataset.};
    
      \end{pgfonlayer}
      
      \node [fill=none, text width=3cm, minimum height=4mm] (WW) at (8.7, 1.5)  {\fontsize{7}{7.5}\selectfont Features' average (0.08)};
      \node [fill=none, text width=2.5cm, minimum height=13mm] (WW2) at (8.3, -0.09)  {};
      \path[every node/.style={font=\sffamily\small}]
    
      (WW) edge[<-, thick, bend left] node [left] {} (WW2); 
      
      %\node [fill=none, text width=2.5cm, minimum height=4mm] (WW) at (13.2, 2)  {\tiny NASA-TLX};
      %\node [fill=none, text width=2.5cm, minimum height=13mm] (WW2) at (12.8, 0.3)  {};
      %\path[every node/.style={font=\sffamily\small}]
    
      %(WW) edge[<-, thick, bend left] node [left] {} (WW2);
   
\end{tikzpicture}

\caption{Spread (calculated with the standard deviation) of computational trust scalars inferred to Barnstar users with models built with knowledge base 1 (\ref{app:trust}). Inferior symbols are used to represent: centroid ($\circ$) and mean of max defuzzification approach ($\bullet$); heuristics $h_1$ to  
$h_4$; grounded ($\odot$), preferred ($\otimes$), and categoriser ($\ddagger$) semantics; and use (respectively no use) of the rule 
weights/arguments strength ($\triangleright$, respectively $\star$).}
\label{figure:kb1_spread}
\end{figure}

\begin{figure} [!h]
\centering

\begin{tikzpicture}
\pgfplotsset{width=1\textwidth}
\begin{pgfonlayer}{bg}
    \begin{axis}[
        compat=newest, %Better label placement
        width  = 1\textwidth,
        height = 4cm,
        major x tick style = transparent,
        ybar=4*\pgflinewidth,
        bar width=2mm,
        bar shift=0pt,
        ymajorgrids = true,
        xlabel = {Reasoning models for computational trust inference},
        ylabel = {Standard deviation},
        label style={font=\fontsize{11}{10.5}\selectfont},
        tick label style={font=\fontsize{11}{10.5}\selectfont},
        nodes near coords,
        nodes near coords align={vertical},
        every node near coord/.append style={font=\fontsize{8}{7.5}\selectfont},
        every node near coord/.append style={rotate=90, anchor=west},
        %symbolic x coords={FL1 FC1, FL2 FC2, FL3 FC3, FL4 FC4, FL5 FC5, FL6 FC6, FL7 FC7, FL8 FC8, FL9 FC9, FL10 FC10, FL11 FC11, FL12 FC12, A1, A2, A3, A4, 
%E1, E2, E3, E4},
        %xtick={ },
        %x tick label style={font=\scriptsize,  align=left,text width=3.5cm},
        xtick={1,2,3,4,5,6,7,8,9,10,11,12,13,14,15,16,17,18,19,20,21,22,23,24,25,26,27,28,29,30,31,32,33,34},
        xticklabels={
        $\star$ $\ddagger$ A8,
        $\star$ $\otimes$ A7,
        $\triangleright$ $\bullet$ FL18,
        $\triangleright$ $\bullet$ FL24,
        $\star$ $\bullet$ FL16,
        $\triangleright$ $\bullet$ FL22,
        $\star$ $\bullet$ FC18,
        $\triangleright$ $\bullet$ FC24,
        $\star$ $\bullet$ FL14,
        $\triangleright$ $\bullet$ FL20,
        $\star$ $\bullet$ FC16,
        $\triangleright$ $\bullet$ FC22,
        $\triangleright$ $\ddagger$ A11,
        $\star$ $\bullet$ FC14,
        $\triangleright$ $\bullet$ FC20,
        $\triangleright$ $\otimes$ A10,
        $\triangleright$ $\odot$ A12,
        $\star$ $\circ$ FC17,
        $\triangleright$ $\circ$ FC23,
        $\triangleright$ $(h_2)$ E6,
        $\triangleright$ $(h_4)$ E8,
        $\star$ $\circ$ FC15,
        $\triangleright$ $\circ$ FC21,
        $\star$ $\circ$ FC13,
        $\triangleright$ $\circ$ FC19,
        $\star$ $\odot$ A9,
        $\star$ $(h_1)$ E5,
        $\star$ $(h_3)$ E7,
        $\star$ $\circ$ FL17,
        $\triangleright$ $\circ$ FL23,
        $\star$ $\circ$ FL15,
        $\triangleright$ $\circ$ FL21,
        $\star$ $\circ$ FL13,
        $\triangleright$ $\circ$ FL19
        },
        x tick label style={font=\fontsize{8}{8.5}\selectfont, align=left, rotate=90, text width=1.3cm},
    %hide obscured x ticks=false,
        %tick align=inside,
        %xtick = data,      
        %tickwidth=4pt,
        %scaled y ticks = false,
        enlarge x limits=0.05,
        ymin=0,
        ymax=0.4,
        clip=false,
        %nodes near coords,
        %every node near coord/.append style={color=black, font=\tiny},
        %legend cell align=left,
        %legend style={
        %        at={(1,1.05)},
        %        anchor=north,
        %        column sep=1ex
        %},
        legend style={at={(0.5,1.30)},
        anchor=north,legend columns=-1,  
        font=\fontsize{7.5}{8.5}\selectfont},
    ]
    
     %\draw [thin, dotted, thick, draw=black] 
     %    (axis cs: -0.5, 366.152) -- (axis cs: 33.5, 366.152);
%         node[pos=0.995, right] {\fontsize{4}{4.5}\selectfont \textbf{NASA}};

   \draw [thin, dotted, draw=black] 
        (axis cs: -0.5, 0.07) -- (axis cs: 35.7, 0.07);
        %node[pos=0.8, above] {\fontsize{5}{5.5}\selectfont \textbf{Features' average (0.07)}};  
  
   \addplot[style={black,pattern=north east lines, pattern color=blue,mark=none, /pgf/number format/fixed,
        /pgf/number format/precision=3}] coordinates {
        (3,0.0548)
        (4,0.0548)
        (5,0.0464)
        (6,0.0464)
        (9,0.0365)
        (10,0.0365)
        (29,0.0174)
        (30,0.0174)
        (31,0.0162)
        (32,0.0162)
        (33,0.0143)
        (34,0.0143)
        };

    \addplot[style={black, pattern=crosshatch dots, pattern color=blue,mark=none, /pgf/number format/fixed,
        /pgf/number format/precision=3}] coordinates {
        (7,0.0423)
        (8,0.0423)
        (11,0.0364)
        (12,0.0364)
        (14,0.0335)
        (15,0.0335)
        (18,0.0232)
        (19,0.0232)
        (22,0.0224)
        (23,0.0224)
        (24,0.0223)
        (25,0.0223)
        };

    \addplot[style={black, pattern=grid, pattern color=green,mark=none, /pgf/number format/fixed,
        /pgf/number format/precision=3}] coordinates {
        (20,0.0225)
        (21,0.0225)
        (27,0.0194)
        (28,0.0194)
        };
    
    \addplot[style={black, pattern=crosshatch, pattern color=red,mark=none, /pgf/number format/fixed,
        /pgf/number format/precision=3}] coordinates {
        (1,0.1981)
        (2,0.1526)
        (13,0.0355)
        (16,0.0265)
        (17,0.0265)
        (26,0.0194)
        };
        
    \legend{Fuzzy (linear FMF), Fuzzy (Gaussian FMF), Expert systems, Defeasible arg.}

    \end{axis}
    
        \node [text width=9cm, fill=white] (A) at (5.4,-2.75) {\textbf{(a)} Models instantiated with the Italian-language dataset.};
    
      \end{pgfonlayer}

      \node [fill=none, text width=3cm, minimum height=4mm] (WW) at (8.7, 1.5)  {\fontsize{7}{7.5}\selectfont Features' average (0.07)};
      \node [fill=none, text width=2.5cm, minimum height=13mm] (WW2) at (10.6, -0.2)  {};
      \path[every node/.style={font=\sffamily\small}]
    
      (WW) edge[<-, thick, bend left] node [left] {} (WW2);
      
      %\node [fill=none, text width=2.5cm, minimum height=4mm] (WW) at (13.2, 2)  {\tiny NASA-TLX};
      %\node [fill=none, text width=2.5cm, minimum height=13mm] (WW2) at (12.8, 0.3)  {};
      %\path[every node/.style={font=\sffamily\small}]
    
      %(WW) edge[<-, thick, bend left] node [left] {} (WW2);
   
\end{tikzpicture}

\vspace{4mm}

\begin{tikzpicture}
\pgfplotsset{width=1\textwidth}
\begin{pgfonlayer}{bg}
    \begin{axis}[
        compat=newest, %Better label placement
        width  = 1\textwidth,
        height = 4cm,
        major x tick style = transparent,
        ybar=4*\pgflinewidth,
        bar width=2mm,
        bar shift=0pt,
        ymajorgrids = true,
        xlabel = {Reasoning models for computational trust inference},
        ylabel = {Standard deviation},
        label style={font=\fontsize{11}{10.5}\selectfont},
        tick label style={font=\fontsize{11}{10.5}\selectfont},
        nodes near coords,
        nodes near coords align={vertical},
        every node near coord/.append style={font=\fontsize{8}{7.5}\selectfont},
        every node near coord/.append style={rotate=90, anchor=west},
        %symbolic x coords={FL1 FC1, FL2 FC2, FL3 FC3, FL4 FC4, FL5 FC5, FL6 FC6, FL7 FC7, FL8 FC8, FL9 FC9, FL10 FC10, FL11 FC11, FL12 FC12, A1, A2, A3, A4, 
%E1, E2, E3, E4},
        %xtick={ },
        %x tick label style={font=\scriptsize,  align=left,text width=3.5cm},
        xtick={1,2,3,4,5,6,7,8,9,10,11,12,13,14,15,16,17,18,19,20,21,22,23,24,25,26,27,28,29,30,31,32,33,34},
        xticklabels={
        $\star$ $\ddagger$ A8,
        $\triangleright$ $(h_2)$ E6,
        $\triangleright$ $(h_4)$ E8,
        $\star$ $\otimes$ A7,
        $\star$ $(h_1)$ E5,
        $\star$ $(h_3)$ E7,
        $\triangleright$ $\bullet$ FL18,
        $\triangleright$ $\bullet$ FL24,
        $\star$ $\bullet$ FL16,
        $\triangleright$ $\bullet$ FL22,
        $\star$ $\bullet$ FC18,
        $\triangleright$ $\bullet$ FC24,
        $\triangleright$ $\ddagger$ A11,
        $\star$ $\bullet$ FC16,
        $\triangleright$ $\bullet$ FC22,
        $\star$ $\bullet$ FC14,
        $\triangleright$ $\bullet$ FC20,
        $\star$ $\bullet$ FL14,
        $\triangleright$ $\bullet$ FL20,
        $\star$ $\odot$ A9,
        $\triangleright$ $\otimes$ A10,
        $\triangleright$ $\odot$ A12,
        $\star$ $\circ$ FC17,
        $\triangleright$ $\circ$ FC23,
        $\star$ $\circ$ FC15,
        $\triangleright$ $\circ$ FC21,
        $\star$ $\circ$ FC13,
        $\triangleright$ $\circ$ FC19,
        $\star$ $\circ$ FL17,
        $\triangleright$ $\circ$ FL23,
        $\star$ $\circ$ FL15,
        $\triangleright$ $\circ$ FL21,
        $\star$ $\circ$ FL13,
        $\triangleright$ $\circ$ FL19},
        x tick label style={font=\fontsize{8}{8.5}\selectfont, align=left, rotate=90, text width=1.3cm},
    %hide obscured x ticks=false,
        %tick align=inside,
        %xtick = data,      
        %tickwidth=4pt,
        %scaled y ticks = false,
        enlarge x limits=0.05,
        ymin=0,
        ymax=0.4,
        clip=false,
        %nodes near coords,
        %every node near coord/.append style={color=black, font=\tiny},
        %legend cell align=left,
        %legend style={
        %        at={(1,1.05)},
        %        anchor=north,
        %        column sep=1ex
        %},
        legend style={at={(0.5,1.30)},
        anchor=north,legend columns=-1,  
        font=\fontsize{7.5}{8.5}\selectfont},
    ]
    
     %\draw [thin, dotted, thick, draw=black] 
     %    (axis cs: -0.5, 366.152) -- (axis cs: 33.5, 366.152);
%         node[pos=0.995, right] {\fontsize{4}{4.5}\selectfont \textbf{NASA}};
 
    \draw [thin, dotted, draw=black] 
        (axis cs: -0.5, 0.08) -- (axis cs: 35.7, 0.08);
  
   \addplot[style={black,pattern=north east lines, pattern color=blue,mark=none, /pgf/number format/fixed,
        /pgf/number format/precision=3}] coordinates {
        (7,0.0536)
        (8,0.0536)
        (9,0.0429)
        (10,0.0429)
        (18,0.0308)
        (19,0.0308)
        (29,0.0171)
        (30,0.0171)
        (31,0.0155)
        (32,0.0155)
        (33,0.0127)
        (34,0.0127)
        };

    \addplot[style={black, pattern=crosshatch dots, pattern color=blue,mark=none, /pgf/number format/fixed,
        /pgf/number format/precision=3}] coordinates {
        (11,0.0412)
        (12,0.0412)
        (14,0.0351)
        (15,0.0351)
        (16,0.0333)
        (17,0.0333)
        (23,0.023)
        (24,0.023)
        (25,0.0217)
        (26,0.0217)
        (27,0.0216)
        (28,0.0216)};

    \addplot[style={black, pattern=grid, pattern color=green,mark=none, /pgf/number format/fixed,
        /pgf/number format/precision=3}] coordinates {
        (2,0.0828)
        (3,0.0828)
        (5,0.07)
        (6,0.07)};
    
    \addplot[style={black, pattern=crosshatch, pattern color=red,mark=none, /pgf/number format/fixed,
        /pgf/number format/precision=3}] coordinates {
        (1,0.2272)
        (4,0.0797)
        (13,0.0382)
        (20,0.0245)
        (21,0.0231)
        (22,0.0231)};
        
    \legend{Fuzzy (linear FMF), Fuzzy (Gaussian FMF), Expert systems, Defeasible arg.}

    \end{axis}
    
    \node [text width=10cm, fill=white] (A) at (5.4,-2.75) {\textbf{(b)} Models instantiated with the Portuguese-language dataset.};
    
      \end{pgfonlayer}
      
      \node [fill=none, text width=3cm, minimum height=4mm] (WW) at (8.7, 1.5)  {\fontsize{7}{7.5}\selectfont Features' average (0.08)};
      \node [fill=none, text width=2.5cm, minimum height=13mm] (WW2) at (10.6, -0.17)  {};
      \path[every node/.style={font=\sffamily\small}]
    
      (WW) edge[<-, thick, bend left] node [left] {} (WW2);
      
      %\node [fill=none, text width=2.5cm, minimum height=4mm] (WW) at (13.2, 2)  {\tiny NASA-TLX};
      %\node [fill=none, text width=2.5cm, minimum height=13mm] (WW2) at (12.8, 0.3)  {};
      %\path[every node/.style={font=\sffamily\small}]
    
      %(WW) edge[<-, thick, bend left] node [left] {} (WW2);
   
\end{tikzpicture}

\caption{Spread (calculated with the standard deviation) of computational trust scalars inferred to Barnstar users with models built with knowledge base 2 (\ref{app:trust}). Inferior symbols are used to represent: centroid ($\circ$) and mean of max defuzzification approach ($\bullet$); heuristics $h_1$ to  
$h_4$; grounded ($\odot$), preferred ($\otimes$), and categoriser ($\ddagger$) semantics; and use (respectively no use) of the rule 
weights/arguments strength ($\triangleright$, respectively $\star$).}
\label{figure:kb2_spread}
\end{figure}

As depicted in Fig. \ref{figure:kb1_spread}, most models built with KB1 achieved a $\sigma$ between 0.07 - 0.18, close to or higher than the results reported by the features' average. Hence, the capacity to differentiate trust levels by most designed models is similar and can be considered appealing.   
Low $\sigma$ values were expected due to the required difference in trust scalars assigned to Barnstars and non-Barnstar editors. In other words, the higher the $\sigma$ the higher the chance of overlapping trust values assigned to Barnstar and non-Barnstar 
editors and, consequently, the worse the rank of Barnstar editors.
% 
% . For instance, suppose a mean of 1 and $\sigma$ of 0.3 for the trust scalars assigned to Barnstar editors by a certain model. By the calculation of 
% standard deviation\footnote{$\sqrt[2]{\sum_{i=1}^{n} \frac{(x_{i} - \mu)}{n}}$, where $n$ is the number of inferences produced by a certain model, $x_{i}$ is 
% an 
% individual inference, and $\mu$ is the mean of the set of inferences.}, 
% 2.1\% of the inferences would be between 0.1 - 0.4, hence, likely overlapping with trust scalars assigned to non-Barnstar editors.

In addition, contrary to the previous analysed evaluation metrics, the models' $\sigma$ varies greatly when instantiated with the Italian-language and the Portuguese-language dataset. This result indicates that no single reasoning approach is better suited to design inference models able to achieve higher values of $\sigma$. An exception was noted for fuzzy reasoning models built with KB1 and employing the mean max defuzzification approach (labelled with $\bullet$). These reported the lowest $\sigma$ (between 0 - 0.07), regardless of the dataset employed and below the features' average. One possibility is that the lower number of inputs required by the mean max approach (only the maximum membership grades) resulted in inferences of less variability when compared to the centroid defuzzification approach, which considers the whole aggregated fuzzy set given by the models' fuzzification module.

As for results depicted in Fig. \ref{figure:kb2_spread}, note that most models built with KB2 achieved even lower $\sigma$ (between 0.01 - 0.08), and were below the features' average.
Therefore, their capacity to differentiate trust levels is similar, but lower when compared to the counterpart models built with KB1.
No single expert system or fuzzy reasoning model reported a high $\sigma$ when built with KB2.
This is an indication that expert system and fuzzy reasoning models are likely not capable of inferring robust trust scalars when built with knowledge bases containing a large number of conflicting information. Particularly, 
they are presumably not able to follow the assumption of trust not being a binary concept.
Some few exceptions were observed for argument-based models.
In particular, models $A7$ and $A8$ achieved 0.19 and 0.15 $\sigma$ (Italian-language dataset), and 0.22 and 0.08 $\sigma$ (Portuguese-language dataset).
These seem to be the only models
built with KB2 able to achieve a favourable spread, while maintaining 0\% of NAs and a robust rank of Barnstar editors (between 0.66 - 2.36 for $A7$, and between 5.3 - 7.14 for $A8$).
This balance illustrates the strong capacity for modelling non-monotonic reasoning by argument-based models defined with the preferred and categoriser semantics while built with KB2. This capacity was not observed in any other investigated reasoning approach. 

\subsection{Summary and Discussion}

The inferential capacity of the employed non-monotonic reasoning approaches 
is examined 
in terms of them being able to produce models whose inferences could be considered valid in the domain of computational trust.
Reasons for the inferences of a model being considered invalid include spread ($\sigma$) lower than the baseline (features' average)
or very high percentage of NAs. In terms of rank of Barnstar all models were considered effective.
Table \ref{table:statusmodels} summarises the results. 

\renewcommand\theadalign{bc}
\begin{table}[!h]
%%\vspace{-3mm}
\renewcommand*{\arraystretch}{1.0}
\footnotesize
\centering
\small
   \caption{Status of the reasoning approaches with respect to the successfully modelling of 
computational trust by models implemented with them. In case of mixed results, it is reported the number of models built with the reasoning approach that were successful over all the ones implemented. Reasons for failing are detailed in parenthesis.}
\setlength{\tabcolsep}{0.1pt}

\setlength{\leftmargini}{0mm}
   \label{table:statusmodels}
    \begin{tabular}{p{3.0cm}P{3.0cm}P{3.5cm}P{2.5cm}} % Column formatting, @{} suppresses leading/trailing space
    \hline
    \multirow{2}{*}{\textbf{Dataset / KB}} &
    \multirow{2}{*}{\begin{tabular}{@{}{c}}\textbf{Expert} \\ \textbf{Systems}\end{tabular}} & 
    \multirow{2}{*}{\begin{tabular}{@{}{c}}\textbf{Fuzzy} \\ \textbf{Reasoning}\end{tabular}} &
    
    \multirow{2}{*}{\begin{tabular}{@{}{c}}\textbf{Defeasible} \\ \textbf{Argumentation}\end{tabular}}  \\
    \\
    \hline
    Italian / KB1 & \textbf{All} & 14/24 (low $\sigma$) & \textbf{All} \\
     \arrayrulecolor{gray!40}\hline
    \multirow{2}{*}{Italian / KB2} & None \, (high \% of NAs and low $\sigma$) & \multirow{2}{*}{None (low $\sigma$)}  & \textbf{2/6} (high \% of NAs and low $\sigma$)\\
    \arrayrulecolor{gray!40}\hline
    \multirow{2}{*}{Portuguese / KB1} & \multirow{2}{*}{\textbf{All}} & 12/24 successful with borderline $\sigma$ & \multirow{2}{*}{\textbf{All}}\\
     \arrayrulecolor{gray!40}\hline
    \multirow{2}{*}{Portuguese / KB2} & \begin{tabular}{@{}{c}}None \\ (high \% of NAs)\end{tabular} & \multirow{2}{*}{None (low $\sigma$)} & \textbf{2/6} (high \% of NAs and low $\sigma$)\\
    \arrayrulecolor{black}\hline
\end{tabular}
\end{table}

As it can be observed, expert systems models presented appealing results when built with KB1. 
This demonstrates that the reasoning approach can be used effectively when instantiated with knowledge bases with a lower number of conflicts, supporting the vast use of expert systems in an ample range of domains present in the literature. Thus, it is important to highlight strengths such as clear reasoning process, capacity to 
keep the language of the domain, and capacity to add and retract rules.
By contrast, when built with KB2, all expert system models were considered unsuccessful when modelling reasoning applied to computational trust. This was mostly due to the high percentage of NAs regardless of the dataset employed. As anticipated, limitations were expected in line with the performed 
literature review. 
%Mainly, the required knowledge acquisition and the not built-in non-monotonicity layer
%are likely the dominant factors limiting the achievement of better inferences.
The lack of a built-in non-monotonicity layer or options for implementing it reinforce the disadvantage of expert systems in dealing with large amounts of uncertain, vague and contradictory information.

Fuzzy reasoning models presented the most divergent reasoning process, when compared to expert system and defeasible argumentation models. While expert systems 
and defeasible argumentation share similarities such as quantification of rules/arguments and their aggregation through measures of central tendency, fuzzy 
reasoning adopts the notions of fuzzy sets and FMF, thus providing a disparate inferential process. This difference offers advantages such as higher precision 
for the modelling of natural language terms and capacity to handle fuzzy concepts. Certainly, such advantages are of great importance when non-monotonic 
reasoning is being performed. However, this higher precision comes with a number of disadvantages, including the definition of membership functions that differ 
from the way in which humans reason. Moreover, in order to manipulate these functions and define an inferential process with them, the definition of a number of 
configuration parameters is necessary. For example, a fuzzy logic operator, a fuzzy inference method, and a defuzzification approach are all needed.
Some mathematical reasoning is required to select the most appropriate parameters,
limiting the applicability of the reasoning approach by domain experts who are not familiar with fuzzy parameters and their interpretations.
In addition, the available options for implementation of non-monotonicity are not well developed. In this research article, the use of possibility theory was selected for being 
intuitively the only approach that allowed the retraction of rules in a fuzzy sense. In other words, partial retractions were allowed, according to the truth 
value of the propositions being evaluated. However, this approach was limited by its order of application, not being commutative. The requirement of all these configurations, and the variability in the inferences produced, seem 
to place fuzzy reasoning in between expert systems and defeasible argumentation, in terms of inferential capacity when performing non-monotonic reasoning. 

Lastly, defeasible argumentation presented the most robust results, being able to produce successful models despite the knowledge base and dataset employed. Similar to expert systems, all models were considered successful when built with KB1, in addition to producing better ranks of Barnstar. When built with KB2, argument-based models were deemed successful when employing the preferred and categoriser semantics without strength of arguments. This shows that some knowledge for the selection of configuration parameters is also required when domain experts decide to use defeasible argumentation.
However, it is argued here that the amount of knowledge is lesser than that required by fuzzy reasoning. 
Moreover, the notion of scepticism behind the behaviour of semantics has been widely discussed in the literature of defeasible argumentation,
making its adoption by experts in other fields more accessible. Lastly, when built with KB2, only argument-based models were able to produce instances in all cases while reporting appealing spread and rank of Barnstars.

To sum up, defeasible
argumentation proved to be the most 
balanced reasoning approach, with models capable of maintaining strong results, despite the complexity of the knowledge base.
Nonetheless, it is important to mention other limitations 
that could be noted when employing it. For instance, as with the other reasoning approaches, knowledge acquisition was a boundary when developing inferential models. This 
limitation could be observed, for example, in the accrual of arguments performed in the last stage of the reasoning process. The accrual based on the cardinality of extensions and measures of 
central tendency, such as average and weighted average of accepted arguments, are simplistic approaches. They could be seen as a surrogate for conflicts not known 
or not established between arguments in the knowledge base. Ideally, if the goal is to reach a single conclusion, another iteration of knowledge acquisition 
should be performed, instead of choosing extensions of higher cardinality (when necessary) and averaging accepted arguments of distinct consequents. This could likely improve the percentage of NAs and increase the number of models deemed successful.
Still, defeasible argumentation was seemingly the most suitable to model
non-monotonic reasoning applied in the domain of computational trust, allowing the creation of models that were constantly among the top-performing ones.
Hence, in conclusion, the acceptance 
status of the hypothesis formulated in Section \ref{designmethodology} is described below.

\begin{itemize}
 \item [-] \textbf{Hypothesis}:  If computational trust is modelled with defeasible argumentation, 
then the inferential capacity of its models will be superior than that achieved by the selected baselines, non-monotonic fuzzy reasoning and expert systems models,
according to a predefined set of 
evaluation metrics from the domain of application. 
\item [-] \textbf{Acceptance status}: \textit{accepted}. 
Reasoning models built with defeasible argumentation were superior for maintaining a balance between the metrics of evaluation, while being able to provide inferences for all instances in 
the datasets. Expert 
systems and fuzzy reasoning models achieved appropriated inferences with less-complex knowledge bases, but did not guarantee the production of inferences in all cases
or appealing models regardless of the knowledge base.
\end{itemize}

% Conclusion section

\section{Conclusion and Future Work}

This research article focused on reviewing possible implementations of non-monotonic reasoning by knowledge-based approaches and conducting
an empirical comparison between them.
Non-monotonic reasoning allows the retraction of previous conclusions in light of new information. It is a compelling approach to model reasoning applied to domains of uncertain information.
The implementation of non-monotonic reasoning required the use of credible domain knowledge bases, which in this study were designed by the author. Their process of creation was exemplified, resulting in two variations: one simplified, with less conflicts; and another with more conflicts and higher topological complexity. Both contained sets of rules, contradictions, natural language terms, and fuzzy membership functions as reported in \ref{app:trust}.  
At last, these knowledge bases were exploited for the design of inference models 
built upon three knowledge-based, reasoning approaches able to perform non-monotonic reasoning: expert systems, fuzzy reasoning, and defeasible argumentation.
Eventually, these same models were instantiated with real-world, quantitative datasets extracted from two Wikipedia dumps of different language editions.
Two dumps were selected to reinforce
findings and extend the generalisability of the study in terms of datasets employed.
After being instantiated, the designed models were used to infer a trust scalar in the [0, 1] $\subset\mathbb{R}$ to each editor of the encyclopaedia, where 1 means complete trust, and 0 means complete absence of trust should be assigned.
Still, the proposed comparison was \textit{not} aimed at enhancing the assessment of computational trust. Instead, it
attempted to situate defeasible 
argumentation, between two other well-known reasoning techniques: expert systems and fuzzy reasoning.
In particular, the inferential capacity of reasoning models built with such approaches were examined.
Three evaluation metrics were selected for this analysis: 1) the sum of the ranks by assigned scalars to recognised trustworthy individuals; 2) the spread of assigned scalars to these same trustworthy editors; and 3) the percentage of instances without an assignment.

This research included a number of limitations.
These are important and can inform future work for extending research on further domains of
application and for considering additional non-monotonic reasoning approaches.
In terms of its design, the domain-specific
metrics selected for the evaluation of reasoning models increased the complexity
of the proposed comparison due to the lack of ground truths. The scope of the design was also
bounded by an empirical comparison between knowledge-based, non-monotonic reasoning
approaches using quantitative data in a real-world context. Alternatively stated, the method employed (empirical
study) to perform the envisioned comparison, was restricted.
The employed
models of inference were designed upon different existing building blocks in the literature of the reasoning approaches they were grounded on.
This approach was necessary in order to achieve working models in the domain of computational trust with the available data and knowledge bases. 
Therefore, the understanding
of the differences between such reasoning approaches in theoretical scenarios, or scenarios unlikely to happen in
the context of modelling reasoning applied to computational trust, is also limited. In addition, the knowledge bases employed by the non-monotonic reasoning models
were limited by their process of formation. This process was highly dependent on domain
experts and on the time required for manual acquisition of information and
formation/validation of structured knowledge. Moreover, each knowledge base was the result
of the information extracted from a single human reasoner and not from multiple reasoners.
The selection of multiple reasoners might require additional financial resources not available
in this research work. 

Despite such limitations, findings indicated how all the employed reasoning approaches allowed for the construction of models that were 
capable of assigning satisfactory trust scalars in certain scenarios. 
In the scenario of the knowledge base with less conflicts, all expert systems and defeasible argumentation models were deemed successful
despite the dataset selected. Contrarily, fuzzy reasoning models led to mixed results, due to its higher number of configuration parameters for design of models. These parameters provided greater flexibility but their complexity and abundance likely limited the applicability and use of fuzzy reasoning by domain experts. When faced with the knowledge base of higher topological complexity, models built with expert systems 
reported a significant number of instances that remained unsolved, suggesting that a lower applicability in this kind of scenario is possible despite the dataset chosen. Fuzzy reasoning models reported less unsolved instances, but the spread of assigned scalars to trustworthy editors was too low for any model to be considered effective, also regardless of the dataset chosen. In contrast, some argument-based models were able to solve all the instances while demonstrating a better capacity of inferring robust trust scalars, or scalars able to show an effective balance among the selected evaluation metrics.

These results could lead to a possible 
interpretation of defeasible argumentation being better suited to capture the underlying reasoning of the knowledge bases employed in this study. 
Still, some differences could be observed by argument-based models employing different configuration parameters. 
For instance, models built with a more credulous rationale (preferred semantics), achieved better inferences. 
The reason for such performance may be due to a high
uncertainty of the domain of computational trust.
This higher 
uncertainty could have originated from: the fact that no other experts were consulted for validation or collaboration during the process of creation of knowledge bases; or by the fact that computational trust is not a well defined construct. Therefore, knowledge bases could likely be further improved in 
order to be used by sceptical argument-based models.

In conclusion, the originality of this research lies in the extensive comparison performed among defeasible argumentation and two other approaches capable of performing non-monotonic reasoning in the domain of computational trust. Previous works \cite{rizzo2020thesis} have attempted to employ defeasible argumentation and to perform similar comparisons also in other domains, such as mental workload modelling \cite{rizzo2020empirical,LONGO2021106514} and mortality occurrence modelling \cite{rizzo2018investigation,RizzoML18}. In
this research article different sets of data and/or models of inference were employed, extending the use of defeasible argumentation in real-world, quantitative contexts. Hence, a broad study has been performed, empirically enhancing the generalisability of defeasible argumentation as a possible approach to reason with quantitative data and conflicting/uncertain knowledge. 
In addition, a review of the investigated reasoning approaches was carried out, including their options for adding a non-monotonic layer.
The practical use of such approaches coupled with a modular design that facilitates similar experiments was exemplified
and their respective implementations made public.
Moreover, the addition and use of a non-monotonic layer in the inferential processes
of models built with expert systems and fuzzy reasoning was also exemplified. Such use is seldom reported in the field
of non-monotonic reasoning.
It might be a useful aid to scholars familiar with these
reasoning approaches and also interested in performing non-monotonic reasoning
activities. 
Overall, this study attempts to serve as a beginning point for other scholars, who could use it to replicate and/or improve the proposed approach in other domains of application. Consequently, this could contribute to the long-term goal of demonstrating the applicability and generalisability of defeasible argumentation with quantitative data in real-world contexts.

Different avenues can be pursued for future work. For example, a comparison of broader scope can be performed by 
adopting different 
structures and configurations of reasoning models.
Another improvement to make the applicability of defeasible argumentation more generalisable
could come from the analysis of other argumentation systems, such as fuzzy argumentation \citep{dondio2017propagating,janssen2008fuzzy}, possibilistic abstract dialectical frameworks \cite{heyninck2021relation},
probabilistic argumentation \citep{li2011probabilistic}, or bipolar 
argumentation \citep{cayrol2009bipolar,cayrol2005acceptability}.
Hybrid reasoning techniques, such as neuro-fuzzy 
systems \citep{nauck1997foundations}, genetic fuzzy systems \citep{cordon2004ten} and fuzzy argumentation \citep{dondio2017propagating} are also recommended.
An investigation of different knowledge bases, both in the domain of computational trust and in new ones, will result 
in new findings. For example, in other contexts of digital collaborative environments, such as blogs, forums and social 
networks.
Another interesting technique for the construction of knowledge bases could be the use of multiple reasoners for knowledge acquisition. Contradictions are generally hard to be 
formalised, and more reasoners might argue among themselves, leading to the creation of 
more 
conflicting rules/arguments. A less time-consuming method to produce new knowledge bases could also be attempted through the development of human-in-the-loop solutions, partially automating the construction of arguments and attacks.
For instance, several works have proposed different techniques for rule extraction from machine learning models \citep{Augasta2012,BARAKAT2010178,Chen2012fuzzyrule,ViloneRL20}.
Finally, a higher explanatory capacity might lead to higher levels 
 of adoption and deployment of non-monotonic reasoning. Explainability is a multifaceted concept \citep{VILONE202189,abdul2018trends,Preece2018why,lipton2016mythos,Miller2019explanation}. Thus, an in-depth investigation in this aspect is 
also suggested.

\bibliography{references.bib}

\appendix

\section{Knowledge Bases} \label{app:trust}

In this section, two knowledge bases are defined for the inference of computational trust. Their features were extract from the files provided by 
Wikipedia dumps (Fig. \ref{fig:xmlstructure}, p. \pageref{fig:xmlstructure}). Nine quantitative features were selected and are detailed next. Rules and contradictions are defined by the 
author, 
who is qualified in computer science and has appropriate experience in a multitude of digital collaborative environments.
Both knowledge bases are consisted of:

\begin{itemize}
 \setlength\itemsep{0.1em} 
 \item A set of features employed for the modelling and assessment of computational trust (Table \ref{table:featurestransf}).
 \item A set of natural language terms associated with numerical ranges used for reasoning with such features, for instance \textit{low} and \textit{high} 
(Table 
\ref{table:featurestransf}).
 \item A set of inferential rules in the form:
 \begin{itemize}
  \item [-] IF B feature A THEN C trust. Where B is a level of feature A and C is a trust level. For 
instance ``\textbf{IF}  \textit{high} \texttt{bytes} \textbf{THEN} \textit{high} \texttt{trust}''. Boolean operators AND/OR might also be used to add other 
premises.
 \end{itemize}
 \item A set of contradictions or meta-rules in the form:
 \begin{itemize}
  \setlength\itemsep{0.1em}
  \item [-] IF B feature A THEN not Rule B
  \item [-] IF Rule A THEN not Rule B
 \end{itemize}
 \setlength\itemsep{0.1em}
 \item A graphical representation of rules and contradictions.
 \item A set of fuzzy membership functions associated with the natural language terms.
\end{itemize}

\normalsize \selectfont

\subsection*{Features and Natural Language Terms of Knowledge Bases 1 and 2}

\fontsize{7}{7.5}\selectfont

\begin{longtable}{p{1.3cm}P{1.6cm}P{2.4cm}P{0.5cm}p{4.1cm}}
\caption{List of features employed by the author for reasoning and inference of trust (as described in Table \ref{table:wikifeatures}), 
followed by transformations applied to each of them, possible values found in the dataset and natural language terms with respective numerical range 
associated. Weights were also defined by the author through a pairwise comparison process.}
\label{table:featurestransf}
\\
\hline
\textbf{Feature} & \textbf{Transformation} & \textbf{Values} & \textbf{Weight} & \textbf{$\quad \quad$Natural lang. terms}\\
\hline
\endfirsthead
\caption[]{List of features employed by the author for reasoning and inference of trust (as described in Table \ref{table:wikifeatures}), 
followed by transformations applied to each of them, possible values found in the dataset and natural language terms with respective numerical range 
associated. Weights were also defined by the author through a pairwise comparison process.}
\\
\hline
\textbf{Feature} & \textbf{Transformation} & \textbf{Values} & \textbf{Weight} & \textbf{$\quad \quad$Natural lang. terms}\\
\hline
\endhead

Pages & None & [1, 1,576,621] $\in\mathbb{N}$ & 1 & 
\begin{minipage}[c]{5cm}
    \begin{enumerate}[label={(\arabic*)},topsep=10pt,itemsep=-0.5ex,partopsep=0ex,parsep=0ex]
    \vspace{1mm}
    \item[-] \textit{low $\in [0, 5]$}
    \item[-] \textit{medium high $\in [10, 19]$}
    \item[-] \textit{high $\in [20, 20+]$}
    \end{enumerate}
  \end{minipage} \\ 
\hline
Activity Factor  & None & [1, 2,850,913] $\in\mathbb{N}$ & 3 &
\begin{minipage}[c]{5cm}
    \begin{enumerate}[label={(\arabic*)},topsep=10pt,itemsep=-0.5ex,partopsep=0ex,parsep=0ex]
    \vspace{1mm}
    \item[-] \textit{low $\in [0, 5]$}
    \item[-] \textit{medium high $\in [10, 19]$}
    \item[-] \textit{high $\in [20, 20+]$}
    \end{enumerate}
  \end{minipage} \\ 
  \hline
Anonymous & None & 1 or 0 & 8 &
\begin{minipage}[c]{5cm}
    \begin{enumerate}[label={(\arabic*)},topsep=10pt,itemsep=-0.5ex,partopsep=0ex,parsep=0ex]
    \vspace{1mm}
    \item[-] \textit{yes $ = 1$}
    \item[-] \textit{no $ = 0$}
    \end{enumerate}
  \end{minipage} \\ 
  \hline
Not minor & $\frac{\text{Not minor}}{\text{Activity fac.}}$ & [0, 1] $\in\mathbb{R}$ & 7 &
\begin{minipage}[c]{5cm}
    \begin{enumerate}[label={(\arabic*)},topsep=10pt,itemsep=-0.5ex,partopsep=0ex,parsep=0ex]
    \vspace{1mm}
    \item[-] \textit{very low $\in [0, 0.05]$}
    \item[-] \textit{medium to high $\in [0.25, 1)$}
    \end{enumerate}
  \end{minipage} \\ 
  \hline
Comments &  $\frac{\text{Comments }}{\text{Activity fac.}}$ & [0, 1] $\in\mathbb{R}$ & 5 & 
\begin{minipage}[c]{5cm}
    \begin{enumerate}[label={(\arabic*)},topsep=10pt,itemsep=-0.5ex,partopsep=0ex,parsep=0ex]
    \vspace{1mm}
    \item[-] \textit{low $\in [0, 0.25)$}
    \item[-] \textit{medium low $\in [0.25, 0.5)$}
    \item[-] \textit{medium high $\in [0.5, 0.75)$}
    \item[-] \textit{high $\in [0.75, 1]$}
    \end{enumerate}
  \end{minipage} \\ 
  \hline
Presence factor & $\frac{\text{Pres. factor}}{\text{Wiki life time}}$ & [0, 1] $\in\mathbb{R}$ & 3 & 
\begin{minipage}[c]{5cm}
    \begin{enumerate}[label={(\arabic*)},topsep=10pt,itemsep=-0.5ex,partopsep=0ex,parsep=0ex]
    \vspace{1mm}
    \item[-] \textit{low $\in [0, 0.25)$}
    \item[-] \textit{medium low $\in [0.25, 0.5)$}
    \item[-] \textit{medium high $\in [0.5, 0.75)$}
    \item[-] \textit{high $\in [0.75, 1]$}
    \end{enumerate}
  \end{minipage} \\ 
  \hline 
Frequency factor & Capped at 1 & [0, 1] $\in\mathbb{R}$ & 5 &
\begin{minipage}[c]{5cm}
    \begin{enumerate}[label={(\arabic*)},topsep=10pt,itemsep=-0.5ex,partopsep=0ex,parsep=0ex]
    \vspace{1mm}
    \item[-] \textit{low $\in [0, 0.25)$}
    \item[-] \textit{medium low $\in [0.25, 0.5)$}
    \item[-] \textit{medium high $\in [0.5, 0.75)$}
    \item[-] \textit{high $\in [0.75, 1]$}
    \end{enumerate}
  \end{minipage} 
\\ 
\hline
Regularity factor & Capped at 1 & [0, 1] $\in\mathbb{R}$ & 3 &
\begin{minipage}[c]{5cm}
    \begin{enumerate}[label={(\arabic*)},topsep=10pt,itemsep=-0.5ex,partopsep=0ex,parsep=0ex]
    \vspace{1mm}
    \item[-] \textit{low $\in [0, 0.25)$}
    \item[-] \textit{medium low $\in [0.25, 0.5)$}
    \item[-] \textit{medium high $\in [0.5, 0.75)$}
    \item[-] \textit{high $\in [0.75, 1]$}
    \end{enumerate}
  \end{minipage} \\ 
  \hline
Bytes & None & [-1 $\times 10^8$, 8 $\times 10^8$] $\in\mathbb{N}$ & 1 &
\begin{minipage}[c]{5cm}
    \begin{enumerate}[label={(\arabic*)},topsep=10pt,itemsep=-0.5ex,partopsep=0ex,parsep=0ex]
    \vspace{1mm}
    \item[-] \textit{low $\in [0, 110]$}
    \item[-] \textit{medium high $\in [512, 2387]$}
    \item[-] \textit{high $\in [2388, 2388+]$}
    \end{enumerate}
  \end{minipage}\\ 
\hline
\end{longtable}
%\end{tabular}
%\end{table}

\begin{figure}[!h]

\lstset{
  language=xml,
  tabsize=3,
  %frame=lines,
  %caption=Xml file structure of Wikipedia,
  %lstlistingname
  label=code:sample,
  %frame=shadowbox,
  rulesepcolor=\color{gray},
  xleftmargin=20pt,
  framexleftmargin=15pt,
  keywordstyle=\color{blue}\bf,
  commentstyle=\color{OliveGreen},
  stringstyle=\color{black},
  numbers=left,
  numberstyle=\tiny,
  numbersep=5pt,
  breaklines=true,
  showstringspaces=false,
  basicstyle=\fontsize{7.5}{8}\selectfont,
  emph={food,name,price},emphstyle={\color{magenta}}}
\begin{lstlisting}
...
<page>
  <title>AccessibleComputing</title>
  <ns>0</ns>
  <id>10</id>
  <redirect title="Computer accessibility" />
  <revision>
    <id>233192</id>
    <timestamp>2001-01-21T02:12:21Z</timestamp>
    <contributor>
      <username>RoseParks</username>
      <id>99</id>
    </contributor>
    <comment>*</comment>
    <model>wikitext</model>
    <format>text/x-wiki</format>
    <text id="233192" bytes="124" />
    <sha1>8kul9tlwjm9oxgvqzbwuegt9b2830vw</sha1>
  </revision>
  <revision>
    ...
  </revision>
</page>
...
\end{lstlisting}

\caption{XML file structure of Wikipedia.}
\label{fig:xmlstructure}
\end{figure}

\newpage

\normalsize

\subsection*{IF-THEN Rules Employed by Knowledge Bases 1 and 2}

\fontsize{7}{7.5}\selectfont
\begin{longtable}{lL{10cm}}
\caption{(fuzzy) IF-THEN rules employed in knowledge bases 1 and 2 designed by the author for inference of computational trust of Wikipedia editors.}
\label{table:kb5ifthenrules}
\\
\hline
\textbf{Label} & \textbf{Internal structure}\\
\hline
\endfirsthead
\caption[]{(fuzzy) IF-THEN rules employed in knowledge bases 1 and 2 designed by the author for inference of computational trust of Wikipedia editors.}
\\
\hline
\textbf{Label} & \textbf{Internal structure}\\
\hline
\endhead
B1 & \textbf{IF}  \textit{medium high} \texttt{bytes} \textbf{THEN} \textit{medium high} \texttt{trust} \\
B2 & \textbf{IF}  \textit{high} \texttt{bytes} \textbf{THEN} \textit{high} \texttt{trust} \\
B3 & \textbf{IF}  \textit{low} \texttt{bytes} \textbf{THEN} \textit{low} \texttt{trust} \\
AF1 & \textbf{IF}  \textit{low} \texttt{activity factor} \textbf{THEN} \textit{low} \texttt{trust} \\
AF2 & \textbf{IF}  \textit{medium high} \texttt{activity factor} \textbf{THEN} \textit{medium high} \texttt{trust} \\
AF3 & \textbf{IF}  \textit{high} \texttt{activity factor} \textbf{THEN} \textit{high} \texttt{trust} \\
AN1 & \textbf{IF}  \textit{no} \texttt{anonymous} \textbf{THEN} \textit{high} \texttt{trust} \\
AN2 & \textbf{IF}  \textit{yes} \texttt{anonymous} \textbf{THEN} \textit{low} \texttt{trust} \\
U1 & \textbf{IF}  \textit{low} \texttt{pages} \textbf{THEN} \textit{low} \texttt{trust} \\
U2 & \textbf{IF}  \textit{medium high} \texttt{pages} \textbf{THEN} \textit{medium low} \texttt{trust} \\
U3 & \textbf{IF}  \textit{high} \texttt{pages} \textbf{THEN} \textit{medium high} \texttt{trust} \\
C1 & \textbf{IF}  \textit{low} \texttt{comments} \textbf{THEN} \textit{low} \texttt{trust} \\
C2 & \textbf{IF}  \textit{medium low} \texttt{comments} \textbf{THEN} \textit{medium low} \texttt{trust} \\
C3 & \textbf{IF}  \textit{medium high} \texttt{comments} \textbf{THEN} \textit{medium high} \texttt{trust} \\
C4 & \textbf{IF}  \textit{high} \texttt{comments} \textbf{THEN} \textit{high} \texttt{trust} \\
P1 & \textbf{IF}  \textit{low} \texttt{presence factor} \textbf{THEN} \textit{low} \texttt{trust} \\
P2 & \textbf{IF}  \textit{medium low} \texttt{presence factor} \textbf{THEN} \textit{medium low} \texttt{trust} \\
P3 & \textbf{IF}  \textit{medium high} \texttt{presence factor} \textbf{THEN} \textit{medium high} \texttt{trust} \\
P4 & \textbf{IF}  \textit{high} \texttt{presence factor} \textbf{THEN} \textit{high} \texttt{trust} \\
F1 & \textbf{IF}  \textit{low} \texttt{frequency factor} \textbf{THEN} \textit{low} \texttt{trust} \\
F2 & \textbf{IF}  \textit{medium low} \texttt{frequency factor} \textbf{THEN} \textit{medium low} \texttt{trust} \\
F3 & \textbf{IF}  \textit{medium high} \texttt{frequency factor} \textbf{THEN} \textit{medium high} \texttt{trust} \\
F4 & \textbf{IF}  \textit{high} \texttt{frequency factor} \textbf{THEN} \textit{high} \texttt{trust} \\
R1 & \textbf{IF}  \textit{low} \texttt{regularity factor} \textbf{THEN} \textit{low} \texttt{trust} \\
R2 & \textbf{IF}  \textit{medium low} \texttt{regularity factor} \textbf{THEN} \textit{medium low} \texttt{trust} \\
R3 & \textbf{IF}  \textit{medium high} \texttt{regularity factor} \textbf{THEN} \textit{medium high} \texttt{trust} \\
R4 & \textbf{IF}  \textit{high} \texttt{regularity factor} \textbf{THEN} \textit{high} \texttt{trust} \\
NM1 & \textbf{IF}  \textit{very low} \texttt{not minor} \textbf{THEN} \textit{low} \texttt{trust} \\
NM2 & \textbf{IF}  \textit{medium to high} \texttt{not minor} \textbf{THEN} \textit{high} \texttt{trust} \\
\hline 
\end{longtable}

\normalsize \selectfont

\subsection*{Contradictions Employed by Knowledge Bases 1 and Graphical Representation}

\fontsize{7}{7.5}\selectfont

\begin{longtable}{lL{10cm}}
\caption{(fuzzy) Contradictions for knowledge base 1 designed by the author for inference of computational trust of Wikipedia editors.}
\label{table:kb1contradictions}
\\
\hline
\textbf{Label } & \textbf{Internal structure}\\
\hline
\endfirsthead
\caption[]{(fuzzy) Contradictions for knowledge base 1 designed by the author for inference of computational trust of Wikipedia editors.}
\\
\hline
\textbf{Label } & \textbf{Internal structure}\\
\hline
\endhead
CC1 & \textbf{IF} NM1 \textbf{THEN not} B1 \\
CC2 & \textbf{IF} NM1 \textbf{THEN not} B2 \\
CC3 & \textbf{IF} NM2 \textbf{THEN not} OnlyAge \\
CC4 & \textbf{IF} P1 \textbf{THEN not} R4 \\
CC5 & \textbf{IF} AF1 \textbf{THEN not} R4 \\
CC6 & \textbf{IF} AF1 \textbf{THEN not} F4 \\
CC7 & \textbf{IF} R1 \textbf{THEN not} P4 \\
CC8 & \textbf{IF} F1 \textbf{THEN not} P4 \\
CC9 & \textbf{IF} NM1 \textbf{THEN not} AF2 \\
CC10 & \textbf{IF} NM1 \textbf{THEN not} AF3 \\
CC11 & \textbf{IF} NM2 \textbf{THEN not} U1 \\
CC12 & \textbf{IF} NM2 \textbf{THEN not} C1 \\
CC13 & \textbf{IF} NM2 \textbf{THEN not} P1 \\
CC14 & \textbf{IF} AN2 \textbf{THEN not} U2 \\
CC15 & \textbf{IF} AN2 \textbf{THEN not} U3 \\
CC16 & \textbf{IF} AN2 \textbf{THEN not} C3 \\
CC17 & \textbf{IF} AN2 \textbf{THEN not} C4 \\
CC18 & \textbf{IF} AN2 \textbf{THEN not} AF2 \\ 
CC19 & \textbf{IF} AN2 \textbf{THEN not} AF3 \\
CC20 & \textbf{IF} AN2 \textbf{THEN not} R4 \\
CC21 & \textbf{IF} AN2 \textbf{THEN not} F4 \\
CC22 & \textbf{IF} AN2 \textbf{THEN not} F3 \\
CC23 & \textbf{IF} AN2 \textbf{THEN not} R3 \\
CC24 & \textbf{IF} AN2 \textbf{THEN not} P3 \\ 
CC25 & \textbf{IF} AN2 \textbf{THEN not} P4 \\
CC26 & \textbf{IF} AN2 \textbf{THEN not} B2 \\
CC27 & \textbf{IF} AN2 \textbf{THEN not} B1 \\
CC28 & \textbf{IF} AN2 \textbf{THEN not} NM2 \\
Bot.a & \textbf{IF}  \textit{yes} \texttt{anonymous} \textit{AND} \textit{low} \texttt{comments} \textit{AND} (\textit{medium high} \texttt{bytes} \textsc{OR} 
\textit{high} \texttt{bytes}) \textit{AND} 
\textit{very low} \texttt{not minor} \textit{AND} (\textit{high} \texttt{unique pages} \textsc{OR} \textit{medium high} \texttt{unique pages})  \textbf{THEN not} U4 \\
Bot.b & \textbf{IF}  \textit{yes} \texttt{anonymous} \textit{AND} \textit{low} \texttt{comments} \textit{AND} (\textit{medium high} \texttt{bytes} \textsc{OR} 
\textit{high} \texttt{bytes}) \textit{AND} 
\textit{very low} \texttt{not minor} \textit{AND} (\textit{high} \texttt{unique pages} \textsc{OR} \textit{medium high} \texttt{unique pages})  \textbf{THEN not} U3 \\
Bot.c & \textbf{IF}  \textit{yes} \texttt{anonymous} \textit{AND} \textit{low} \texttt{comments} \textit{AND} (\textit{medium high} \texttt{bytes} \textsc{OR} 
\textit{high} \texttt{bytes}) \textit{AND} 
\textit{very low} \texttt{not minor} \textit{AND} (\textit{high} \texttt{unique pages} \textsc{OR} \textit{medium high} \texttt{unique pages})  \textbf{THEN not} U2 \\
Bot.d & \textbf{IF}  \textit{yes} \texttt{anonymous} \textit{AND} \textit{low} \texttt{comments} \textit{AND} (\textit{medium high} \texttt{bytes} \textsc{OR} 
\textit{high} \texttt{bytes}) \textit{AND} 
\textit{very low} \texttt{not minor} \textit{AND} (\textit{high} \texttt{unique pages} \textsc{OR} \textit{medium high} \texttt{unique pages})  \textbf{THEN not} C1 \\
Bot.e & \textbf{IF}  \textit{yes} \texttt{anonymous} \textit{AND} \textit{low} \texttt{comments} \textit{AND} (\textit{medium high} \texttt{bytes} \textsc{OR} 
\textit{high} \texttt{bytes}) \textit{AND} 
\textit{very low} \texttt{not minor} \textit{AND} (\textit{high} \texttt{unique pages} \textsc{OR} \textit{medium high} \texttt{unique pages})  \textbf{THEN not} B2 \\
Bot.f & \textbf{IF}  \textit{yes} \texttt{anonymous} \textit{AND} \textit{low} \texttt{comments} \textit{AND} (\textit{medium high} \texttt{bytes} \textsc{OR} 
\textit{high} \texttt{bytes}) \textit{AND} 
\textit{very low} \texttt{not minor} \textit{AND} (\textit{high} \texttt{unique pages} \textsc{OR} \textit{medium high} \texttt{unique pages})  \textbf{THEN not} B1 \\
Bot.g & \textbf{IF}  \textit{yes} \texttt{anonymous} \textit{AND} \textit{low} \texttt{comments} \textit{AND} (\textit{medium high} \texttt{bytes} \textsc{OR} 
\textit{high} \texttt{bytes}) \textit{AND} 
\textit{very low} \texttt{not minor} \textit{AND} (\textit{high} \texttt{unique pages} \textsc{OR} \textit{medium high} \texttt{unique pages})  \textbf{THEN not} AF2 \\
Bot.h & \textbf{IF}  \textit{yes} \texttt{anonymous} \textit{AND} \textit{low} \texttt{comments} \textit{AND} (\textit{medium high} \texttt{bytes} \textsc{OR} 
\textit{high} \texttt{bytes}) \textit{AND} 
\textit{very low} \texttt{not minor} \textit{AND} (\textit{high} \texttt{unique pages} \textsc{OR} \textit{medium high} \texttt{unique pages})  \textbf{THEN not} AF3 \\
Vandal.a & \textbf{IF}  (\textit{low} \texttt{presence factor} \textsc{OR} \textit{medium low} \texttt{presence factor}) \textit{AND} \textit{low} 
\texttt{regularity factor} \textit{AND} \textit{low} 
\texttt{comments} \textit{AND} \textit{low} \texttt{unique pages} \textbf{THEN not} AF2 \\
Vandal.b & \textbf{IF}  (\textit{low} \texttt{presence factor} \textsc{OR} \textit{medium low} \texttt{presence factor}) \textit{AND} \textit{low} 
\texttt{regularity factor} \textit{AND} \textit{low} 
\texttt{comments} \textit{AND} \textit{low} \texttt{unique pages}  \textbf{THEN not} AF3 \\
Vandal.c & \textbf{IF}  (\textit{low} \texttt{presence factor} \textsc{OR} \textit{medium low} \texttt{presence factor}) \textit{AND} \textit{low} 
\texttt{regularity factor} \textit{AND} \textit{low} 
\texttt{comments} \textit{AND} \textit{low} \texttt{unique pages}  \textbf{THEN not} B1 \\
Vandal.d & \textbf{IF}  (\textit{low} \texttt{presence factor} \textsc{OR} \textit{medium low} \texttt{presence factor}) \textit{AND} \textit{low} 
\texttt{regularity factor} \textit{AND} \textit{low} 
\texttt{comments} \textit{AND} \textit{low} \texttt{unique pages}  \textbf{THEN not} B2 \\
OnlyAge.a & \textbf{IF}  \textit{low} \texttt{frequency factor} \textit{AND} \textit{low} \texttt{regularity factor} \textit{AND} \textit{low} \texttt{activity 
factor} \textbf{THEN not} P4 \\
OnlyAge.b & \textbf{IF}  \textit{low} \texttt{frequency factor} \textit{AND} \textit{low} \texttt{regularity factor} \textit{AND} \textit{low} \texttt{activity 
factor} \textbf{THEN not} P3 \\
OnlyAge.c & \textbf{IF}  \textit{low} \texttt{frequency factor} \textit{AND} \textit{low} \texttt{regularity factor} \textit{AND} \textit{low} \texttt{activity 
factor} \textbf{THEN not} P2 \\
\hline 
\end{longtable}

\normalsize \selectfont

\begin{figure}[H]
\centering
    \includegraphics[scale=0.30]{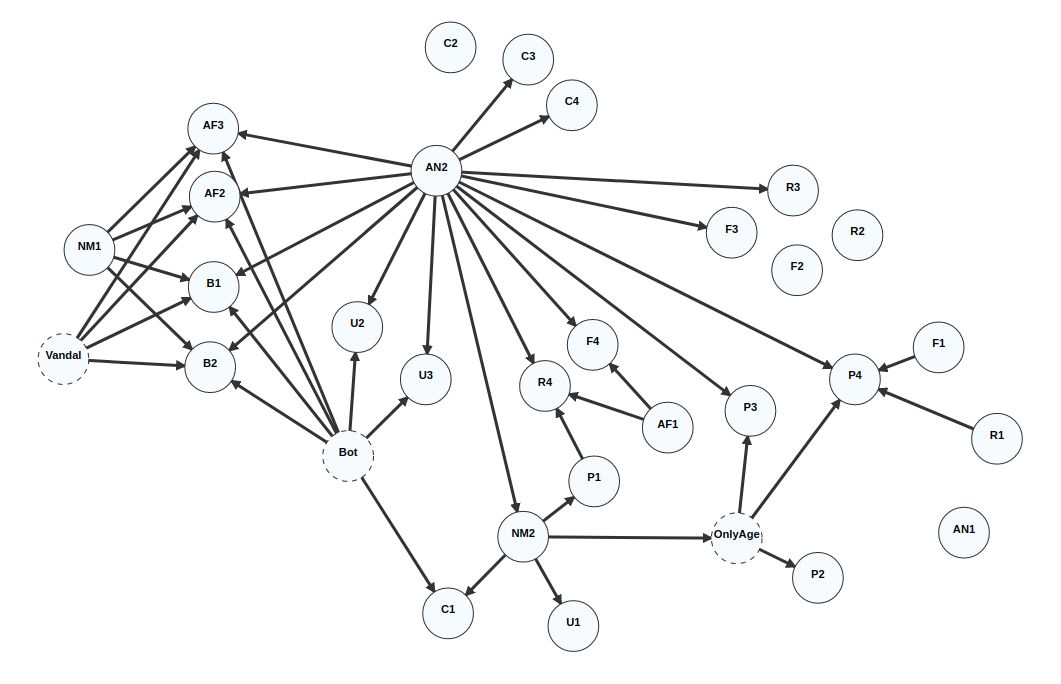}
    \caption{Graphical representation of knowledge base 1. Nodes might represent (fuzzy) IF-THEN rules (continuous circles) or premises of contradictions (dashed circles). Arrows represent 
contradictions between two rules.}
\label{figure:kb6}
\end{figure}

\subsection*{Contradictions Employed by Knowledge Bases 2 and Graphical Representation}

\newlength{\NOTskip} 
\def\NOT#1{\settowidth{\NOTskip}{\ensuremath{#1}}%
            \hspace{0.5\NOTskip}\mathclap{\not}\hspace{-0.5\NOTskip}#1}

\fontsize{7}{7.5}\selectfont
\begingroup
\renewcommand\arraystretch{1.14}
\begin{longtable}{llll}
\caption{(fuzzy) Contradictions for knowledge base 2 designed by the author for inference of computational trust of Wikipedia editors. ``Rule A $\nrightarrow$ 
Rule B'' reads as ``If Rule A then not Rule B''. ``Rule A $\nleftrightarrow$ Rule B'' reads as two rules: ``If Rule A 
then not Rule B'', and ``If Rule B then not Rule A''.}
\label{table:kb1contradictions}
\\
\hline
\textbf{Contradic. Rules} & \textbf{Contradict. Rules} & \textbf{Contradict. Rules} & \textbf{Contradict. Rules}\\
\hline
\endfirsthead
\caption[]{(fuzzy) Contradictions for knowledge base 2 designed by the author for inference of computational trust of Wikipedia editors. ``Rule A 
$\nrightarrow$ 
Rule B'' reads as ``If Rule A then not Rule B''. ``Rule A $\nleftrightarrow$ Rule B'' reads as two rules: ``If Rule A 
then not Rule B'', and ``If Rule B then not Rule A''.}
\\
\hline
\textbf{Contradict. Rules} & \textbf{Contradict. Rules} & \textbf{Contradict. Rules} & \textbf{Contradict. Rules}\\
\hline
\endhead
AN2 $\nrightarrow$ AF3 & AN2 $\nrightarrow$ AF2 & C1 $\nrightarrow$ AF3 & C1 $\nrightarrow$ AF2 \\
F1 $\nrightarrow$ AF3 & F1 $\nrightarrow$ AF2 & NM1 $\nrightarrow$ AF3 & NM1 $\nrightarrow$ AF2 \\
R1 $\nrightarrow$ AF2 & R1 $\nrightarrow$ AF3 & P1 $\nrightarrow$ AF3 & P1 $\nrightarrow$ AF2 \\
U1 $\nrightarrow$ AF2 & U1 $\nrightarrow$ AF3 & AN2 $\nrightarrow$ B1 & AN2 $\nrightarrow$ B2 \\
AN2 $\nrightarrow$ C3 & AN2 $\nrightarrow$ C4 & AN2 $\nrightarrow$ F4 & AN2 $\nrightarrow$ F3 \\
AN2 $\nrightarrow$ NM2 & AN2 $\nrightarrow$ R4 & AN2 $\nrightarrow$ R3 & AN2 $\nrightarrow$ P3 \\
AN2 $\nrightarrow$ P4 & AN2 $\nrightarrow$ U2 & AN2 $\nrightarrow$ U3 & AF1 $\nrightarrow$ B2 \\
AF1 $\nrightarrow$ B1 & B3 $\nrightarrow$ AF2 & B3 $\nrightarrow$ AF3 & NM1 $\nrightarrow$ B1 \\
NM1 $\nrightarrow$ B2 & R3 $\nrightarrow$ C1 & R4 $\nrightarrow$ C1 & AF1 $\nrightarrow$ F4 \\
AF1 $\nrightarrow$ F3 & F1 $\nrightarrow$ R3 & F1 $\nrightarrow$ R4 & R1 $\nrightarrow$ F4 \\
R1 $\nrightarrow$ F3 & F1 $\nrightarrow$ P4 & F1 $\nrightarrow$ P3 & P1 $\nrightarrow$ F4 \\
P1 $\nrightarrow$ F3 & C4 $\nrightarrow$ NM1 & C3 $\nrightarrow$ NM1 & AF1 $\nrightarrow$ R3 \\
AF1 $\nrightarrow$ R4 & R1 $\nrightarrow$ P4 & R1 $\nrightarrow$ P3 & P1 $\nrightarrow$ R3 \\
P1 $\nrightarrow$ R4 & AF1 $\nrightarrow$ P4 & AF1 $\nrightarrow$ P3 & AF2 $\nleftrightarrow$ AF1 \\
AF3 $\nleftrightarrow$ AF1 & AN1 $\nleftrightarrow$ AF1 & C2 $\nleftrightarrow$ AF1 & C3 $\nleftrightarrow$ AF1 \\
C4 $\nleftrightarrow$ AF1 & F2 $\nleftrightarrow$ AF1 & NM2 $\nleftrightarrow$ AF1 & R2 $\nleftrightarrow$ AF1 \\
P2 $\nleftrightarrow$ AF1 & U2 $\nleftrightarrow$ AF1 & U3 $\nleftrightarrow$ AF1 & AF3 $\nleftrightarrow$ AF2 \\
AN1 $\nleftrightarrow$ AF2 & B2 $\nleftrightarrow$ AF2 & C2 $\nleftrightarrow$ AF2 & C4 $\nleftrightarrow$ AF2 \\
F2 $\nleftrightarrow$ AF2 & F4 $\nleftrightarrow$ AF2 & NM2 $\nleftrightarrow$ AF2 & R2 $\nleftrightarrow$ AF2 \\ 
R4 $\nleftrightarrow$ AF2 & P2 $\nleftrightarrow$ AF2 & P4 $\nleftrightarrow$ AF2 & U3 $\nleftrightarrow$ AF2 \\ 
B1 $\nleftrightarrow$ AF3 & C2 $\nleftrightarrow$ AF3 & C3 $\nleftrightarrow$ AF3 & F2 $\nleftrightarrow$ AF3 \\
F3 $\nleftrightarrow$ AF3 & R2 $\nleftrightarrow$ AF3 & R3 $\nleftrightarrow$ AF3 & P2 $\nleftrightarrow$ AF3 \\
P3 $\nleftrightarrow$ AF3 & U2 $\nleftrightarrow$ AF3 & AN2 $\nleftrightarrow$ AN1 & B1 $\nleftrightarrow$ AN1 \\
C1 $\nleftrightarrow$ AN1 & C2 $\nleftrightarrow$ AN1 & C3 $\nleftrightarrow$ AN1 & F1 $\nleftrightarrow$ AN1 \\
F2 $\nleftrightarrow$ AN1 & F3 $\nleftrightarrow$ AN1 & NM1 $\nleftrightarrow$ AN1 & R1 $\nleftrightarrow$ AN1 \\
R2 $\nleftrightarrow$ AN1 & R3 $\nleftrightarrow$ AN1 & P1 $\nleftrightarrow$ AN1 & P2 $\nleftrightarrow$ AN1 \\
P3 $\nleftrightarrow$ AN1 & U1 $\nleftrightarrow$ AN1 & U2 $\nleftrightarrow$ AN1 & B3 $\nleftrightarrow$ AN1 \\
C2 $\nleftrightarrow$ AN2 & F2 $\nleftrightarrow$ AN2 & R2 $\nleftrightarrow$ AN2 & P2 $\nleftrightarrow$ AN2 \\
B2 $\nleftrightarrow$ B1 & C1 $\nleftrightarrow$ B1 & C2 $\nleftrightarrow$ B1 & C4 $\nleftrightarrow$ B1 \\
F1 $\nleftrightarrow$ B1 & F2 $\nleftrightarrow$ B1 & F4 $\nleftrightarrow$ B1 & NM2 $\nleftrightarrow$ B1 \\
R1 $\nleftrightarrow$ B1 & R2 $\nleftrightarrow$ B1 & R4 $\nleftrightarrow$ B1 & P1 $\nleftrightarrow$ B1 \\
P2 $\nleftrightarrow$ B1 & P4 $\nleftrightarrow$ B1 & U1 $\nleftrightarrow$ B1 & U3 $\nleftrightarrow$ B1 \\
B3 $\nleftrightarrow$ B1 & C1 $\nleftrightarrow$ B2 & C2 $\nleftrightarrow$ B2 & C3 $\nleftrightarrow$ B2 \\
F1 $\nleftrightarrow$ B2 & F2 $\nleftrightarrow$ B2 & F3 $\nleftrightarrow$ B2 & R1 $\nleftrightarrow$ B2 \\
R2 $\nleftrightarrow$ B2 & R3 $\nleftrightarrow$ B2 & P1 $\nleftrightarrow$ B2 & P2 $\nleftrightarrow$ B2 \\
P3 $\nleftrightarrow$ B2 & U1 $\nleftrightarrow$ B2 & U2 $\nleftrightarrow$ B2 & B3 $\nleftrightarrow$ B2 \\
C2 $\nleftrightarrow$ C1 & C3 $\nleftrightarrow$ C1 & C4 $\nleftrightarrow$ C1 & F2 $\nleftrightarrow$ C1 \\
F3 $\nleftrightarrow$ C1 & F4 $\nleftrightarrow$ C1 & NM2 $\nleftrightarrow$ C1 & R2 $\nleftrightarrow$ C1 \\
P2 $\nleftrightarrow$ C1 & P3 $\nleftrightarrow$ C1 & P4 $\nleftrightarrow$ C1 & U2 $\nleftrightarrow$ C1 \\
U3 $\nleftrightarrow$ C1 & C3 $\nleftrightarrow$ C2 & C4 $\nleftrightarrow$ C2 & F1 $\nleftrightarrow$ C2 \\
F3 $\nleftrightarrow$ C2 & F4 $\nleftrightarrow$ C2 & NM1 $\nleftrightarrow$ C2 & NM2 $\nleftrightarrow$ C2 \\
R1 $\nleftrightarrow$ C2 & R3 $\nleftrightarrow$ C2 & R4 $\nleftrightarrow$ C2 & P1 $\nleftrightarrow$ C2 \\
P3 $\nleftrightarrow$ C2 & P4 $\nleftrightarrow$ C2 & U1 $\nleftrightarrow$ C2 & U2 $\nleftrightarrow$ C2 \\
U3 $\nleftrightarrow$ C2 & B3 $\nleftrightarrow$ C2 & C4 $\nleftrightarrow$ C3 & F1 $\nleftrightarrow$ C3 \\
F2 $\nleftrightarrow$ C3 & F4 $\nleftrightarrow$ C3 & NM2 $\nleftrightarrow$ C3 & R1 $\nleftrightarrow$ C3 \\
R2 $\nleftrightarrow$ C3 & R4 $\nleftrightarrow$ C3 & P1 $\nleftrightarrow$ C3 & P2 $\nleftrightarrow$ C3 \\
P4 $\nleftrightarrow$ C3 & U1 $\nleftrightarrow$ C3 & U3 $\nleftrightarrow$ C3 & B3 $\nleftrightarrow$ C3 \\
F1 $\nleftrightarrow$ C4 & F2 $\nleftrightarrow$ C4 & F3 $\nleftrightarrow$ C4 & R1 $\nleftrightarrow$ C4 \\
R2 $\nleftrightarrow$ C4 & R3 $\nleftrightarrow$ C4 & P1 $\nleftrightarrow$ C4 & P2 $\nleftrightarrow$ C4 \\
P3 $\nleftrightarrow$ C4 & U1 $\nleftrightarrow$ C4 & U2 $\nleftrightarrow$ C4 & B3 $\nleftrightarrow$ C4 \\
F2 $\nleftrightarrow$ F1 & F3 $\nleftrightarrow$ F1 & F4 $\nleftrightarrow$ F1 & NM2 $\nleftrightarrow$ F1 \\
R2 $\nleftrightarrow$ F1 & P2 $\nleftrightarrow$ F1 & U2 $\nleftrightarrow$ F1 & U3 $\nleftrightarrow$ F1 \\
F3 $\nleftrightarrow$ F2 & F4 $\nleftrightarrow$ F2 & NM1 $\nleftrightarrow$ F2 & NM2 $\nleftrightarrow$ F2 \\
R1 $\nleftrightarrow$ F2 & R3 $\nleftrightarrow$ F2 & R4 $\nleftrightarrow$ F2 & P1 $\nleftrightarrow$ F2 \\
P3 $\nleftrightarrow$ F2 & P4 $\nleftrightarrow$ F2 & U1 $\nleftrightarrow$ F2 & U2 $\nleftrightarrow$ F2 \\
U3 $\nleftrightarrow$ F2 & B3 $\nleftrightarrow$ F2 & F4 $\nleftrightarrow$ F3 & NM1 $\nleftrightarrow$ F3 \\
NM2 $\nleftrightarrow$ F3 & R2 $\nleftrightarrow$ F3 & R4 $\nleftrightarrow$ F3 & P2 $\nleftrightarrow$ F3 \\
P4 $\nleftrightarrow$ F3 & U1 $\nleftrightarrow$ F3 & U3 $\nleftrightarrow$ F3 & B3 $\nleftrightarrow$ F3 \\
NM1 $\nleftrightarrow$ F4 & R2 $\nleftrightarrow$ F4 & R3 $\nleftrightarrow$ F4 & P2 $\nleftrightarrow$ F4 \\
P3 $\nleftrightarrow$ F4 & U1 $\nleftrightarrow$ F4 & U2 $\nleftrightarrow$ F4 & B3 $\nleftrightarrow$ F4 \\
NM2 $\nleftrightarrow$ NM1 & R2 $\nleftrightarrow$ NM1 & R3 $\nleftrightarrow$ NM1 & R4 $\nleftrightarrow$ NM1 \\
P2 $\nleftrightarrow$ NM1 & P3 $\nleftrightarrow$ NM1 & P4 $\nleftrightarrow$ NM1 & U2 $\nleftrightarrow$ NM1 \\
U3 $\nleftrightarrow$ NM1 & R1 $\nleftrightarrow$ NM2 & R2 $\nleftrightarrow$ NM2 & R3 $\nleftrightarrow$ NM2 \\
P1 $\nleftrightarrow$ NM2 & P2 $\nleftrightarrow$ NM2 & P3 $\nleftrightarrow$ NM2 & U1 $\nleftrightarrow$ NM2 \\
U2 $\nleftrightarrow$ NM2 & B3 $\nleftrightarrow$ NM2 & R2 $\nleftrightarrow$ R1 & R3 $\nleftrightarrow$ R1 \\
R4 $\nleftrightarrow$ R1 & P2 $\nleftrightarrow$ R1 & U2 $\nleftrightarrow$ R1 & U3 $\nleftrightarrow$ R1 \\
R3 $\nleftrightarrow$ R2 & R4 $\nleftrightarrow$ R2 & P1 $\nleftrightarrow$ R2 & P3 $\nleftrightarrow$ R2 \\
P4 $\nleftrightarrow$ R2 & U1 $\nleftrightarrow$ R2 & U2 $\nleftrightarrow$ R2 & U3 $\nleftrightarrow$ R2 \\
B3 $\nleftrightarrow$ R2 & R4 $\nleftrightarrow$ R3 & P2 $\nleftrightarrow$ R3 & P4 $\nleftrightarrow$ R3 \\
U1 $\nleftrightarrow$ R3 & U3 $\nleftrightarrow$ R3 & B3 $\nleftrightarrow$ R3 & P2 $\nleftrightarrow$ R4 \\
P3 $\nleftrightarrow$ R4 & U1 $\nleftrightarrow$ R4 & U2 $\nleftrightarrow$ R4 & B3 $\nleftrightarrow$ R4 \\
P2 $\nleftrightarrow$ P1 & P3 $\nleftrightarrow$ P1 & P4 $\nleftrightarrow$ P1 & U2 $\nleftrightarrow$ P1 \\
U3 $\nleftrightarrow$ P1 & P3 $\nleftrightarrow$ P2 & P4 $\nleftrightarrow$ P2 & U1 $\nleftrightarrow$ P2 \\
U2 $\nleftrightarrow$ P2 & U3 $\nleftrightarrow$ P2 & B3 $\nleftrightarrow$ P2 & P4 $\nleftrightarrow$ P3 \\
U1 $\nleftrightarrow$ P3 & U3 $\nleftrightarrow$ P3 & B3 $\nleftrightarrow$ P3 & U1 $\nleftrightarrow$ P4 \\
U2 $\nleftrightarrow$ P4 & B3 $\nleftrightarrow$ P4 & U2 $\nleftrightarrow$ U1 & U3 $\nleftrightarrow$ U1 \\
U3 $\nleftrightarrow$ U2 & B3 $\nleftrightarrow$ U2 & B3 $\nleftrightarrow$ U3 & \\
\hline 
\end{longtable}
\endgroup
\normalsize \selectfont

\begin{figure}[H]
\centering
    \includegraphics[scale=0.32]{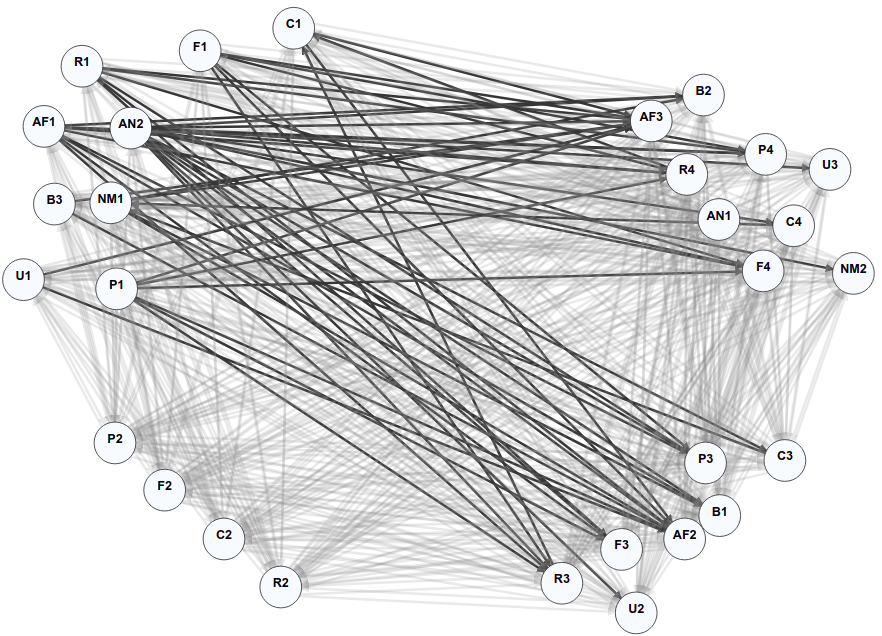}
    \caption{Graphical representation of knowledge base 2. Nodes represent (fuzzy) IF-THEN rules. Dark arrows represent 
contradictions between two rules due to some believed inconsistency by the author grounded in his domain knowledge. Light arrows represent 
contradictions between rules due to different consequents being inferred by each rule.}
\label{figure:kb7}
\end{figure}

\subsection*{Fuzzy Membership Functions}

\normalsize \selectfont

Fig. \ref{fig:fuzzyTrust} depicts the possible fuzzy membership functions employed for modelling the natural language terms listed in 
Table \ref{table:featurestransf}. Some terms present only triangular membership functions because they were modelled with absolute values extracted from the 
Wikimedia Foundation's Analytics.  

\begin{figure}[H]
  \centering

  \subfloat [Activity factor levels]{\includegraphics[width=114mm, height=34mm]{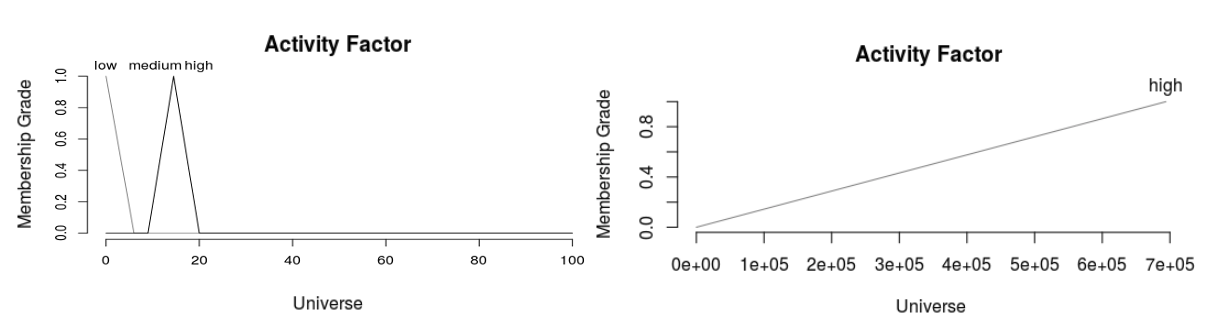}}
  \\
    \subfloat[Bytes levels]{\includegraphics[width=114mm, height=34mm]{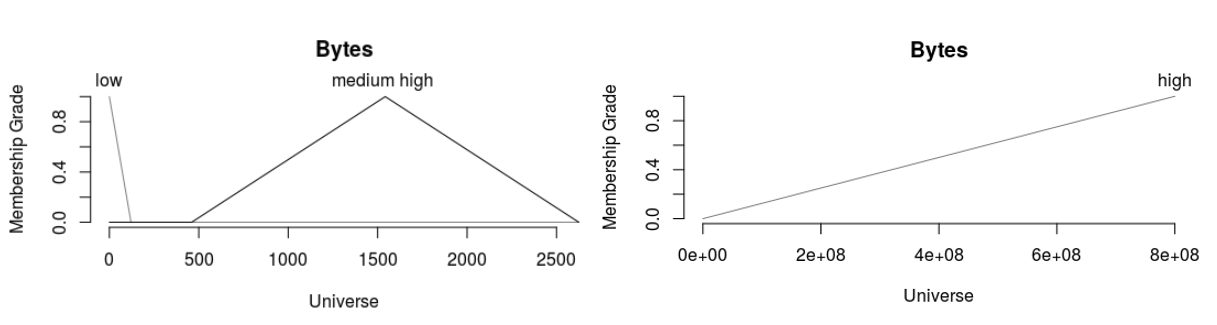}}
  \\
  \subfloat[Pages levels]{\includegraphics[width=114mm, height=34mm]{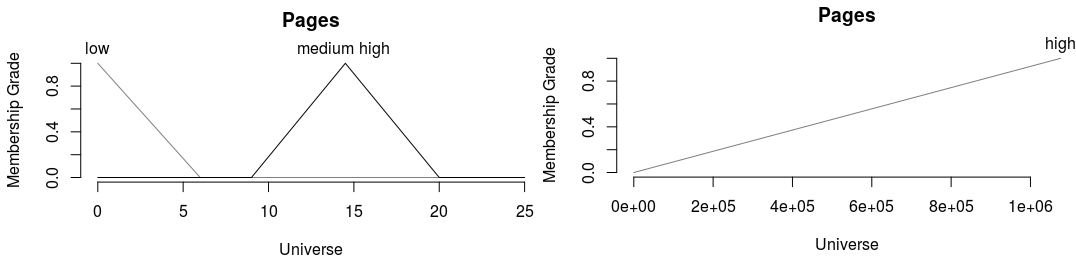}}
  \\
  \subfloat[Not minor levels]{\includegraphics[scale=0.41]{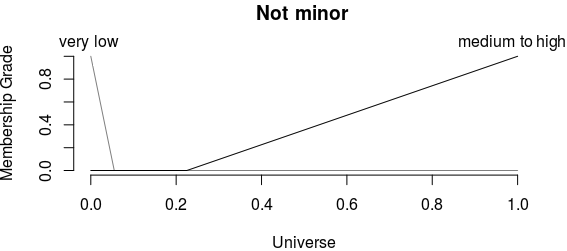}}
  \\
  
  \caption{Employed fuzzy membership functions for different levels related to computational trust itself and its selected features.}
    \label{fig:fuzzyTrust}

\end{figure}

\begin{figure}[H]
\ContinuedFloat
  \centering

  \subfloat[Triangular functions for comments, frequency factor, regularity factor, presence 
factor, and computational trust]{\includegraphics[scale=0.4]{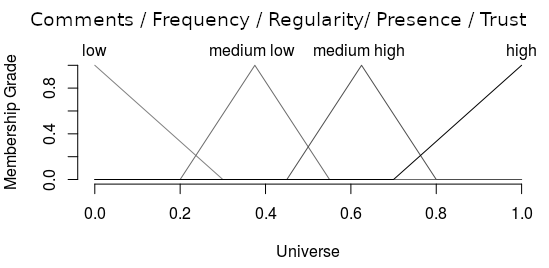}}
\\
\subfloat[Gaussian functions for comments, frequency factor, regularity factor, and presence 
factor, and computational trust]{\includegraphics[scale=0.4]{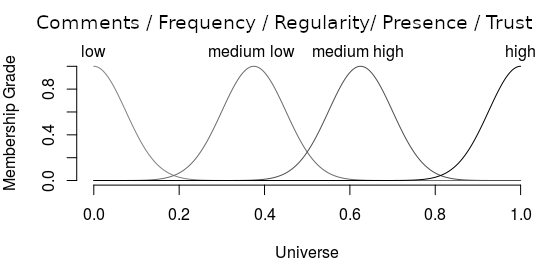}}
  \\
  \caption{Employed fuzzy membership functions for different levels related to computational trust itself and its selected features.}
    \label{fig:fuzzyTrust}

\end{figure}

\section{List of Designed Non-monotonic Reasoning Models}\label{listmodels}

\footnotesize

\begin{table}[!h]
\footnotesize
\setlength{\tabcolsep}{6pt}
    \renewcommand*{\arraystretch}{1.10}
   \caption{Non-monotonic models built upon expert systems.}
\label{table:modelsExpert}   
\centering
\begin{tabular}{cP{5cm}c}
\hline
\multirow{2}{*}{\textbf{Model}} & \multirow{2}{*}{\textbf{Knowledge base} (\ref{app:trust})} & 
\multirow{2}{*}{\begin{tabular}{@{}{c}}\textbf{Heuristic} \\ (p. \pageref{list:heuristics}) \end{tabular}} \\
%& \multirow{2}{*}{\begin{tabular}{@{}{c}}\textbf{Experiment} \\ (Table \ref{table:datasets})\end{tabular}}\\
\\
 \hline
E1 (E5) & KB1 (KB2) & $h_1$ \\ % & $E_a$\\
E2 (E6) & KB1 (KB2) & $h_2$ \\ % & $E_a$\\
E3 (E7) & KB1 (KB2) & $h_3$ \\ % & $E_a$\\
E4 (E8) & KB1 (KB2) & $h_4$ \\ % & $E_a$\\
\hline
\end{tabular}
\end{table}

\begingroup
\renewcommand\arraystretch{0.8}
\begin{longtable}{P{1.9cm}P{0.6cm}cP{0.8cm}P{4.6cm}}
\caption{Non-monotonic models built upon fuzzy reasoning. Fuzzy operators selected are Zadeh (Z), Product (P) and \L{}ukasiewicz (\L{}). Knowledge bases (KB) 
and fuzzy membership functions (FMF) are detailed in \ref{app:trust}.}
\label{table:modelsFuzzy}
\\
%\renewcommand*{\arraystretch}{0.95}
%\footnotesize
%\setlength{\tabcolsep}{3pt}
%\scriptsize
%\centering
\hline
\multirow{2}{*}{\textbf{Model}} & \multirow{2}{*}{\textbf{Oper.}} & 
\multirow{2}{*}{\begin{tabular}{@{}{c}}\textbf{Def.} \\ \textbf{method} 
\end{tabular}} & \multirow{2}{*}{\begin{tabular}{@{}{c}}\textbf{Rule} \\ \textbf{weight}\end{tabular}} &
\multirow{2}{*}{\textbf{KB} + \textbf{FMF} (\ref{app:trust})} 
\\
\\
\hline
\endhead
$FL1$ ($FL13$) & Z  & Centroid & no & KB1 (KB2) + Triangular \\ % & $E_a$\\
$FL2$ ($FL14$) & Z  & Mean of max & no & KB1 (KB2)+ Triangular \\ % & $E_a$\\
$FL3$ ($FL15$) & P  & Centroid & no & KB1 (KB2) + Triangular \\ % & $E_a$\\
$FL4$ ($FL16$) & P  & Mean of max & no & KB1 (KB2) + Triangular \\ % & $E_a$\\
$FL5$ ($FL17$) & \L{}  & Centroid & no & KB1 (KB2) + Triangular \\ % & $E_a$\\
$FL6$ ($FL18$) & \L{}  & Mean of max & no & KB1 (KB2) + Triangular \\ % & $E_a$\\
$FL7$ ($FL19$) & Z  & Centroid & yes & KB1 (KB2) + Triangular \\ % & $E_a$\\
$FL8$ ($FL20$) & Z  & Mean of max & yes & KB1 (KB2) + Triangular \\ % & $E_a$\\
$FL9$ ($FL21$) & P  & Centroid & yes & KB1 (KB2) + Triangular \\ % & $E_a$\\
$FL10$ ($FL22$) & P  & Mean of max & yes & KB1 (KB2) + Triangular \\ % & $E_a$\\
$FL11$ ($FL23$) & \L{}  & Centroid & yes & KB1 (KB2) + Triangular \\ % & $E_a$\\
$FL12$ ($FL24$) & \L{}  & Mean of max & yes & KB1 (KB2) + Triangular \\ % & $E_a$\\
\hline
$FC1$ ($FC13$) & Z  & Centroid & no & KB1 (KB2) + Gaussian \\ % & $E_a$\\
$FC2$ ($FC14$) & Z  & Mean of max & no & KB1 (KB2) + Gaussian \\ % & $E_a$\\
$FC3$ ($FC15$) & P  & Centroid & no & KB1 (KB2) + Gaussian \\ % & $E_a$\\
$FC4$ ($FC16$) & P  & Mean of max & no & KB1 (KB2) + Gaussian \\ % & $E_a$\\
$FC5$ ($FC17$) & \L{}  & Centroid & no & KB1 (KB2) + Gaussian \\ % & $E_a$\\
$FC6$ ($FC18$) & \L{}  & Mean of max & no & KB1 (KB2) + Gaussian \\ % & $E_a$\\
$FC7$ ($FC19$) & Z  & Centroid & yes & KB1 (KB2) + Gaussian \\ % & $E_a$\\
$FC8$ ($FC20$) & Z  & Mean of max & yes & KB1 (KB2) + Gaussian \\ % & $E_a$\\
$FC9$ ($FC21$) & P  & Centroid & yes & KB1 (KB2) + Gaussian \\ % & $E_a$\\
$FC10$ ($FC22$) & P  & Mean of max & yes & KB1 (KB2) + Gaussian \\ % & $E_a$\\
$FC11$ ($FC23$) & \L{}  & Centroid & yes & KB1 (KB2) + Gaussian \\ % & $E_a$\\
$FC12$ ($FC24$) & \L{}  & Mean of max & yes & KB1 (KB2) + Gaussian \\ % & $E_a$\\
\hline
\end{longtable}
\endgroup

\footnotesize

\begingroup
\renewcommand\arraystretch{1.0}
\begin{longtable}{cP{2.5cm}P{2.2cm}cc}
\caption{Non-monotonic models built upon defeasible argumentation. Knowledge bases employed in Layer 1 and 2 are detailed in \ref{app:trust}.}
\label{table:modelsArgumentation}
\\
\hline
%\renewcommand*{\arraystretch}{0.95}
%\footnotesize
\setlength{\tabcolsep}{1pt}
%\scriptsize
%\centering
%\hline
\multirow{4}{*}{\textbf{Model}} & 
%\multirow{2}{*}{\begin{tabular}{@{}{c}}\textbf{Exp.} \\ (Table \ref{table:datasets}) \end{tabular}} & 
\textbf{Layer 1 and Layer 2} & \multirow{2}{*}{\textbf{Layer 3}} & \multirow{2}{*}{\textbf{Layer 4}} & \multirow{2}{*}{\textbf{Layer 5}}\\
\cline{2-5}
& \textbf{Arguments and Conflicts} & \textbf{Attack relation} & \multirow{2}{*}{\textbf{Semantics}} & \multirow{2}{*}{\textbf{Accrual}}
\\
\hline
\endhead
A1 (A7)  & KB1 (KB2) & Binary & Preferred & card. + average \\
A2 (A8) & KB1 (KB2) & Binary & Categorizer & average \\
A3 (A9) & KB1 (KB2) & Binary & Grounded & average \\
A4 (A10) & KB1 (KB2) & Strength of arg. & Preferred & card. + w. average \\
A5 (A11) & KB1 (KB2) & Strength of arg. & Categorizer & w. average \\
A6 (A12) & KB1 (KB2) & Strength of arg. & Grounded & w. average \\
\hline
\end{longtable}
\endgroup

\section{List of Results by Model}

\begingroup
\renewcommand\arraystretch{0.80}
\begin{longtable}{cccccccc}
%P{1.2cm}P{0.6cm}cP{0.8cm}P{4.6cm}}
\caption{List or results achieved by each model for each evaluated metric when instantiated by the Wikipedia Italian edition and by the Wikipedia Portuguese edition. The full list of models and their configurations are given in \ref{listmodels}. The sum of the absolute difference between results achieved for each dataset is given in the last column. The closer to 0 this difference gets, the more stable a model behaved when instantiated with two different datasets.}
\label{table:allresults}
\\
\hline
\multirow{2}{*}{\textbf{Model}} & 
%\multirow{2}{*}{\begin{tabular}{@{}{c}}\textbf{Exp.} \\ (Table \ref{table:datasets}) \end{tabular}} & 
\multicolumn{3}{c}{\textbf{Italian dataset}} & \multicolumn{3}{c}{\textbf{Portuguese dataset}} & \multirow{2}{*}{\textbf{Sum abs. diff.}}\\
\cline{2-7}
& \textbf{Rank} & \textbf{Spread} & \textbf{NAs} & \textbf{Rank} & \textbf{Spread} & \textbf{NAs} \\
\hline
\endhead
$A01$ & 0.38 & 0.11 & 0 & 0.28 & 0.1 & 0 & 0.11\\
$A02$ & 0.37 & 0.11 & 0 & 0.28 & 0.1 & 0 & 0.11\\
$A03$ & 0.38 & 0.11 & 0 & 0.28 & 0.1 & 0 & 0.11\\
$A04$ & 0.42 & 0.08 & 0 & 0.45 & 0.11 & 0 & 0.06\\
$A05$ & 0.29 & 0.08 & 0 & 0.46 & 0.11 & 0 & 0.21\\
$A06$ & 0.29 & 0.08 & 0 & 0.45 & 0.11 & 0 & 0.2\\
$A07$ & 2.36 & 0.15 & 0 & 0.66 & 0.08 & 0 & 1.77\\
$A08$ & 5.3 & 0.2 & 0 & 7.14 & 0.23 & 0 & 1.87\\
$A09$ & 0 & 0.02 & 51.31 & 0 & 0.02 & 50.43 & 0.88\\
$A10$ & 8.19 & 0.03 & 0 & 12.13 & 0.02 & 0 & 3.94\\
$A11$ & 9.5 & 0.04 & 0 & 13.35 & 0.04 & 0 & 3.86\\
$A12$ & 8.19 & 0.03 & 0 & 12.13 & 0.02 & 0 & 3.94\\
\hline
$E01$ & 0.32 & 0.07 & 0 & 1.24 & 0.13 & 0 & 0.98\\
$E02$ & 0.32 & 0.07 & 0 & 1.78 & 0.14 & 0 & 1.51\\
$E03$ & 0.37 & 0.11 & 0 & 0.28 & 0.1 & 0 & 0.11\\
$E04$ & 0.42 & 0.08 & 0 & 0.46 & 0.11 & 0 & 0.07\\
$E05$ & 0 & 0.02 & 51.31 & 0 & 0.07 & 50.43 & 0.93\\
$E06$ & 0 & 0.02 & 51.31 & 0 & 0.08 & 50.43 & 0.94\\
$E07$ & 0 & 0.02 & 51.31 & 0 & 0.07 & 50.43 & 0.93\\
$E08$ & 0 & 0.02 & 51.31 & 0 & 0.08 & 50.43 & 0.94\\
\hline
$FC01$ & 0.51 & 0.08 & 0 & 0.72 & 0.08 & 0 & 0.21\\
$FC02$ & 10.4 & 0 & 0 & 14.05 & 0 & 0 & 3.66\\
$FC03$ & 0.43 & 0.08 & 0 & 0.44 & 0.08 & 0 & 0.01\\
$FC04$ & 10.4 & 0 & 0 & 14.05 & 0 & 0 & 3.66\\
$FC05$ & 0.31 & 0.08 & 0 & 0.41 & 0.08 & 0 & 0.11\\
$FC06$ & 4.7 & 0.04 & 0 & 5.95 & 0.04 & 0 & 1.26\\
$FC07$ & 0.51 & 0.08 & 0 & 0.72 & 0.08 & 0 & 0.21\\
$FC08$ & 10.4 & 0 & 0 & 14.05 & 0 & 0 & 3.66\\
$FC09$ & 0.43 & 0.08 & 0 & 0.44 & 0.08 & 0 & 0.01\\
$FC10$ & 10.4 & 0 & 0 & 14.05 & 0 & 0 & 3.66\\
$FC11$ & 0.31 & 0.08 & 0 & 0.41 & 0.08 & 0 & 0.11\\
$FC12$ & 4.7 & 0.04 & 0 & 5.95 & 0.04 & 0 & 1.26\\
$FC13$ & 1.96 & 0.02 & 3.56 & 2.47 & 0.02 & 5.17 & 2.12\\
$FC14$ & 1.96 & 0.03 & 3.52 & 2.41 & 0.03 & 5.1 & 2.04\\
$FC15$ & 1.73 & 0.02 & 3.43 & 2.1 & 0.02 & 5.01 & 1.94\\
$FC16$ & 1.79 & 0.04 & 3.38 & 2.19 & 0.04 & 4.94 & 1.96\\
$FC17$ & 1.74 & 0.02 & 3.43 & 2.09 & 0.02 & 5.01 & 1.93\\
$FC18$ & 1.8 & 0.04 & 3.38 & 2.18 & 0.04 & 4.94 & 1.95\\
$FC19$ & 1.96 & 0.02 & 3.56 & 2.47 & 0.02 & 5.17 & 2.12\\
$FC20$ & 1.96 & 0.03 & 3.52 & 2.41 & 0.03 & 5.1 & 2.04\\
$FC21$ & 1.73 & 0.02 & 3.43 & 2.1 & 0.02 & 5.01 & 1.94\\
$FC22$ & 1.79 & 0.04 & 3.38 & 2.19 & 0.04 & 4.94 & 1.96\\
$FC23$ & 1.74 & 0.02 & 3.43 & 2.09 & 0.02 & 5.01 & 1.93\\
$FC24$ & 1.8 & 0.04 & 3.38 & 2.18 & 0.04 & 4.94 & 1.95\\
\hline
$FL01$ & 0.5 & 0.12 & 0 & 0.34 & 0.07 & 0 & 0.2\\
$FL02$ & 10.4 & 0 & 0 & 14.05 & 0 & 0 & 3.66\\
$FL03$ & 4.51 & 0.19 & 0 & 0.25 & 0.07 & 0 & 4.38\\
$FL04$ & 10.4 & 0 & 0 & 14.05 & 0 & 0 & 3.66\\
$FL05$ & 0.5 & 0.12 & 0 & 0.26 & 0.07 & 0 & 0.28\\
$FL06$ & 5.23 & 0.07 & 0 & 5.19 & 0.02 & 0 & 0.09\\
$FL07$ & 0.5 & 0.12 & 0 & 0.34 & 0.07 & 0 & 0.21\\
$FL08$ & 10.4 & 0 & 0 & 14.05 & 0 & 0 & 3.66\\
$FL09$ & 3.21 & 0.18 & 0 & 0.25 & 0.07 & 0 & 3.06\\
$FL10$ & 10.4 & 0 & 0 & 14.05 & 0 & 0 & 3.66\\
$FL11$ & 0.5 & 0.12 & 0 & 0.26 & 0.07 & 0 & 0.28\\
$FL12$ & 5.23 & 0.07 & 0 & 5.19 & 0.02 & 0 & 0.09\\
$FL13$ & 1.59 & 0.01 & 3.59 & 1.99 & 0.01 & 5.32 & 2.12\\
$FL14$ & 1.59 & 0.04 & 3.59 & 1.99 & 0.03 & 5.32 & 2.12\\
$FL15$ & 1.27 & 0.02 & 3.59 & 1.42 & 0.02 & 5.31 & 1.88\\
$FL16$ & 1.27 & 0.05 & 3.59 & 1.42 & 0.04 & 5.31 & 1.88\\
$FL17$ & 1.27 & 0.02 & 3.59 & 1.41 & 0.02 & 5.31 & 1.86\\
$FL18$ & 1.27 & 0.05 & 3.59 & 1.41 & 0.05 & 5.31 & 1.86\\
$FL19$ & 1.59 & 0.01 & 3.59 & 1.99 & 0.01 & 5.32 & 2.12\\
$FL20$ & 1.59 & 0.04 & 3.59 & 1.99 & 0.03 & 5.32 & 2.12\\
$FL21$ & 1.27 & 0.02 & 3.59 & 1.42 & 0.02 & 5.31 & 1.88\\
$FL22$ & 1.27 & 0.05 & 3.59 & 1.42 & 0.04 & 5.31 & 1.88\\
$FL23$ & 1.27 & 0.02 & 3.59 & 1.41 & 0.02 & 5.31 & 1.86\\
$FL24$ & 1.27 & 0.05 & 3.59 & 1.41 & 0.05 & 5.31 & 1.86\\
\hline
\end{longtable}
\endgroup

\section{Definitions of Acceptability Semantics}\label{app:semantics}

In this research article, it is important to define the most common Dung semantics \cite{Dung1995}, such as grounded, preferred, and stable, as well as other important notions such as reinstatement and conflict-freeness.

\begin{definition}[Shorthand notations \cite{caminada2009logical}]
Let $\langle Ar , att\rangle$ be an Argumentation Framework (AF), $A, B \in Ar$ and $Args \subseteq Ar$.

\begin{itemize}
 \item $A^+$ as $\{B\,|\,(A,B) \in att\}$.
 \item $Args^+$ as $\{B\,|\,(A,B) \in att$ for some $A \in Args\}$.
 \item $A^-$ as $\{B\,|\,(B,A) \in att\}$.
 \item $Args^-$ as $\{B\,|\,(B,A) \in att$ for some $A \in Args\}$.
\end{itemize}
\end{definition}

$A^+$ indicates the arguments attacked by $A$, while $A^-$ indicates the arguments attacking $A$. $Args^+$ indicates the set of arguments attacked by $Args^+$, while $Args^-$ indicates the set of arguments attacking $Args^-$.

\begin{definition}[Conflict-free \cite{caminada2009logical}]
Let $\langle Ar , att\rangle$ be an AF and $Args \subseteq Ar$. $Args$ is \textit{conflict-free} iff $Args \cap Args^+ = \emptyset$.
\end{definition}

\begin{definition}[Defence \cite{caminada2009logical}]
Let $\langle Ar , att\rangle$ be an AF, $A \in Ar$ and $Args \subseteq Ar$. $Args$ defends an argument $A$ iff $A^- \subseteq Args^+$.
\end{definition}

\begin{definition}[Reinstatement labelling \citep{caminada2006issue}]
Let $\langle Arg, att \rangle$ be an AF and $Lab : Arg \rightarrow \{in, out,$ $undec\}$ be a labelling function. $Lab$ is a 
\textit{reinstatement labelling} iff it satisfies:

\begin{itemize}
 \item $\forall A \in Ar : (Lab(A) = out \equiv \exists B \in Ar : (B$ \textit{defends} $A \wedge Lab(B) = in))$ and
 \item $\forall A \in Ar : (Lab(A) = in \equiv \forall B \in Ar : (B$ \textit{defends} $A \supset Lab(B) = out))$ 
\end{itemize}
\end{definition}

\begin{definition}[Dung's acceptability semantics \citep{Dung1995}, as defined in \citep{caminada2009logical}]\label{def:dungssemantics}
Let $Args$ be a conflict-free set of arguments, $F : 2^{Args} \rightarrow 2^{Args}$ a function such that $F(Args) = \{A\,|\,A$ is defended by $Args\}$ and
$Lab : Args \rightarrow \{in, out, undec\}$ a reinstatement labelling function. Also consider $in(Lab)$ short for $\{A \in Args\,|\,Lab(A) = in\}$, 
$out(Lab)$ short for $\{A \in Args\,|\,Lab(A) = out\}$ and $undec(Lab)$ short for $\{A \in Args\,|\,Lab(A) = undec\}$.

\begin{itemize}
 \item $Args$ is \textit{admissible} if $Args \subseteq F(Args)$.
 \item $Args$ is a \textit{complete} extension if $Args = F(Args)$.
 \item $in(Lab)$ is a \textit{grounded} extension if $undec(Lab)$ is maximal, or $in(Lab)$ is minimal, or $out(Lab)$ is minimal.
 \item $in(Lab)$ is a \textit{preferred} extension if $in(Lab)$ is maximal or $out(Lab)$ is maximal.
 \item $in(Lab)$ is a \textit{stable} extension if $undec(Lab) = \emptyset$.
\end{itemize}
\end{definition}

The categoriser ranking-based semantics \cite{besnard2001logic} is also employed in this research article. It assigns a value to each argument based on its number of attackers. To do so, a categoriser 
function and a categoriser semantics are defined as follows:

\begin{definition}[Categoriser function \cite{besnard2001logic}]
Let $\langle Args, att \rangle$ be an argumentation framework. Then, 
$Cat : Args \rightarrow (0, 1]$ is the categoriser function defined as:
\[
    Cat(a)= 
\begin{cases}{}
    1 & \text{if } a^- = \emptyset \\
    \frac{1}{1 + \sum\limits_{c \in a^-}Cat(c)} & \text{otherwise}
\end{cases}
\]
\end{definition}

\begin{definition}[Categoriser semantics \cite{besnard2001logic}]
Given an argumentation framework $\langle Args, att \rangle$ and a categoriser function $Cat : Args \rightarrow (0, 1]$, a ranking-based categoriser semantics 
associates a ranking $\succeq^{Cat}_{AF}$ on $Args$ such that $\forall a, b \in Args, a \succeq^{Cat}_{AF} b$ iff $Cat(a) \ge Cat(b)$.
\end{definition}

\end{document}